\documentclass[journal]{IEEEtran}

\usepackage{graphicx}
\usepackage{cite}
\usepackage{amsthm,amsmath,amssymb}
\usepackage{tabularx}
\usepackage{mathrsfs}
\usepackage{bm}
\usepackage{subfigure}
\usepackage{multirow}

\usepackage[ruled]{algorithm2e}
\usepackage{makecell}
\usepackage{stfloats}
\usepackage{float}
\usepackage{url}
\usepackage[hidelinks]{hyperref}

\newcommand{\bc}[1]{\textcolor{blue}{#1}}
\newcommand{\rc}[1]{\textcolor{red}{#1}}

\hypersetup{
    colorlinks=true
}

\begin{document}


\title{Unsupervised Foggy Scene Understanding via Self Spatial-Temporal Label Diffusion}


\author{Liang~Liao,~\IEEEmembership{Member,~IEEE,}
        Wenyi~Chen,
        Jing~Xiao,~\IEEEmembership{Member,~IEEE,}
        Zheng~Wang,~\IEEEmembership{Member,~IEEE,} \\
        Chia-Wen~Lin,~\IEEEmembership{Fellow,~IEEE,}
        and~Shin'ichi~Satoh,~\IEEEmembership{Member,~IEEE}
\thanks{L. Liao and S. Satoh are with the Digital Content and Media Sciences Research Division, National Institute of Informatics, Tokyo 101-8430, Japan (e-mail:\{liang, satoh\}@nii.ac.jp).}
\thanks{J. Xiao, W. Chen and Z. Wang are with the School of Computer Science, Wuhan University, Wuhan 430072, China (e-mail:\{jing, wenyichen, wangzwhu\}@whu.edu.cn).}
\thanks{C.-W. Lin is with the Department of Electrical Engineering and the Institute of Communications Engineering, National Tsing Hua University, Hsinchu 30013, Taiwan (e-mail: cwlin@ee.nthu.edu.tw).}
}

\markboth{IEEE Transactions on Image Processing}%
{Liao \MakeLowercase{\textit{et al.}}: Unsupervised Foggy Scene Understanding via Self Spatial-Temporal Label Diffusion}

\maketitle

\begin{abstract}

Understanding foggy image sequence in driving scene is critical for autonomous driving, but it remains a challenging task due to the difficulty in collecting and annotating real-world images of adverse weather. Recently, self-training strategy has been considered as a powerful solution for unsupervised domain adaptation, 
which iteratively adapts the model from the source domain to the target domain by generating target pseudo labels and re-training the model. However, the selection of confident pseudo labels inevitably suffers from the conflict between sparsity and accuracy, both of which will lead to suboptimal models. To tackle this problem, we exploit the characteristics of the foggy image sequence of driving scenes to densify the confident pseudo labels. Specifically, based on the two discoveries of local spatial similarity and adjacent temporal correspondence of the sequential image data, we propose a novel Target-Domain driven pseudo label Diffusion 
(TDo-Dif) scheme. It employs superpixels and optical flows to identify the spatial similarity and temporal correspondence, respectively, and then diffuses the confident but sparse pseudo labels within a superpixel or a temporal corresponding pair linked by the flow. Moreover, to ensure the feature similarity of the diffused pixels, we introduce local spatial similarity loss and temporal contrastive loss in the model re-training stage. 
Experimental results show that our TDo-Dif scheme helps the adaptive model achieve 51.92\% and 53.84\% mean intersection-over-union (mIoU) on two publicly available natural foggy datasets (Foggy Zurich and Foggy Driving), which exceeds the state-of-the-art unsupervised domain adaptive semantic segmentation methods. 
The proposed method can also be applied to non-sequential images in the target domain by considering only spatial similarity. Models and data can be found at https://github.com/velor2012/TDo-Dif.
\end{abstract}

\begin{IEEEkeywords}
Natural foggy scene, semantic segmentation, unsupervised domain adaptation, self-training, label diffusion
\end{IEEEkeywords}

\IEEEpeerreviewmaketitle

\section{Introduction}

\IEEEPARstart{S}{emantic} segmentation refers to the task of assigning semantic labels to each pixel of an input image~\cite{journals/pami/ChenPKMY18,journals/tip/ZhouWCYBX21,seg_survey}. 
This task has been active in the field of computer vision for decades due to its broad range of downstream applications, such as autonomous driving~\cite{conf/mm/WuJYZC19,journals/tits/ZhouBWN20}, medical image analysis~\cite{journals/tmi/RaviFCY17,journals/air/TaghanakiACCH21}, and image restoration~and image restoration~\cite{conf/cvpr/WangYDL18,Liao2020ECCV}.
Recently, convolutional neural networks (CNNs) have achieved great success in semantic scene understanding and high accuracy on standard vision benchmarks \cite{EveringhamEGWWZ15,conf/eccv/LinMBHPRDZ14,cordts2016cityscapes}. However, most learning-based algorithms are developed under clear visibility and often encounter challenges in real-life scenarios, especially in outdoor applications under ``bad" weather conditions~\cite{journals/vtm/ZangDSTRK19,Jiang2022TIP,Xiao2020ijcai,Xu2021TOMM}. In this paper, we will focus on the understanding of foggy weather scenarios. 

\begin{figure}[tb]
     \centering
     \includegraphics[width=0.95\columnwidth]{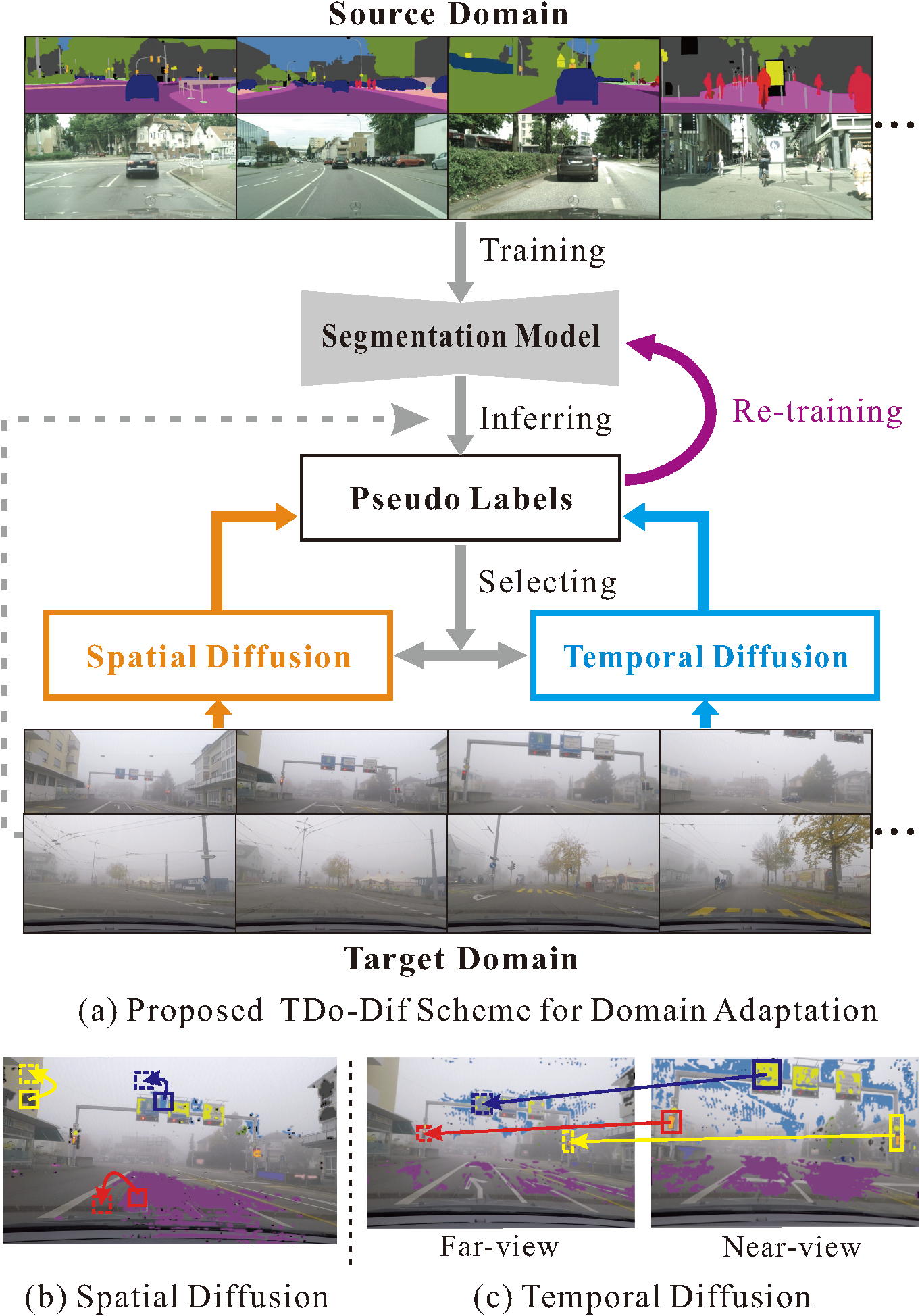}
     \caption{(a) The unsupervised self-training framework for domain adaptation using the proposed TDo-Dif scheme. The spatial and temporal pseudo label diffusion modules are proposed by exploiting the spatial similarity and temporal correspondences in the \textbf{target domain data}. In addition to the initial pseudo labels generated by knowledge transfer from the source domain, labels are diffused to the unknown pixels by exploiting the knowledge from the target domain. All pseudo labels are then used to re-train the segmentation model, helping the model better adapt to the distribution of the target domain. In (b) and (c), the initial pseudo labels are overlaid on top of the input foggy images, and the arrows indicate the  directions of diffusion.}
     \label{motivation}
\end{figure}

One of the issues that hinder the performance of learning-based algorithms under foggy weather is the difficulty in collecting and annotating real-world images of adverse weather. The reason is that the manual annotation is hardly scalable to so many scenarios, and it is much harder to provide precise manual annotations due to poor visibility. An alternative way is to simulate real foggy scenarios using synthetic datasets, which renders the available clear-weather images with some physical models (\emph{e.g.}, atmospheric scattering model) \cite{sakaridis2018semantic,SakaridisDHG18,dai2020curriculum} or by generative adversarial networks (GANs) based image-to-image translation \cite{gong2020analogical}. 
Nevertheless, it is difficult to approximate fog in complex natural environments because existing physical models for fog synthesis tend to assume that the fog is homogeneous. In addition, the subtle differences in texture, color and illumination conditions between clear-weather and foggy-weather images pose a challenge for performing image translation. As a result, there remains a domain gap between the synthetic images and the real foggy images, resulting in a significant performance degradation when applying the model trained on synthetic data to real scenes.


Deep self-training approaches emerge as powerful solutions for unsupervised domain adaptation and have been commonly used in cross-domain semantic segmentation \cite{ZouYKW18,LianDLG19,conf/iccv/ZouYLKW19,journals/pami/ZhangDFG20,conf/eccv/MeiZZZ20,conf/eccv/ShinWPK20}. They typically adopt an iterative two-step pipeline: 1) predicting the pseudo labels of the target domain; and 2) re-training the segmentation network by using the source labels and the target pseudo labels. However, the pseudo labels predicted by models trained on the source domain (\emph{i.e.}, clear-weather images) usually suffer from inaccurate predictions for the target domain (\emph{i.e.}, foggy-weather images). Although some works set confidence thresholds to neglect the low-confidence predictions, this often leads to very sparse pseudo labels, which may further result in sub-optimal models \cite{ZouYKW18,zou2019confidence}. The problem of sparse labels is exacerbated in the foggy scenarios, as both the scene and weather conditions change.

In this paper, we address domain adaptive semantic segmentation for foggy driving scenarios based on self-training from a novel perspective, \emph{i.e.}, by casting it as a pseudo label diffusion problem from the target domain. In the driving scenario, sequence data is collected from the forward-looking dash-board camera of a car (examples of image sequence shown in Fig.~\ref{motivation}), and the object in the later image is closer to the camera than the previous image. By exploiting the sequential data in the foggy target domain, we have made two discoveries. 
\textbf{Discovery 1 (Spatial Similarity):} As shown in Fig.~\ref{motivation}(b), only a small portion of a category is assigned confident pseudo labels, but the texture of adjacent pixels within the category looks similar. Thus, we can diffuse the pseudo labels to their adjacent similar pixels.
\textbf{Discovery 2 (Temporal Correspondence):} As shown in Fig.~\ref{motivation}(c), when sequential data is available in some cases, the pseudo labels of an object are more confident in the near-view image than in the corresponding far-view image (the near-view image is captured later than the far-view image), probably due to higher resolution and lighter fog. Taking the \textit{Traffic Sign} as an example, it is neglected in pseudo labels of the far-view image, but labeled in the near-view image. Thus we can diffuse its pseudo labels from near-view to far-view image. 

Based on these discoveries, we propose a self-training framework based on \textbf{T}arget \textbf{Do}main driven pseudo label \textbf{Dif}fusion (\textbf{TDo-Dif}), in which we first exploit the spatial similarity in terms of color and structures to propagate the sparsely labeled pixels to their similar neighbours. Since the sequential data is available, we also exploit the temporal correspondence to propagate labels from near-view images to far-view images. In particular, we develop spatial and temporal pseudo label diffusion modules to propagate the initial sparse pseudo labels, and integrate them into the self-training framework (Fig.~\ref{motivation}). 
In spatial diffusion, we employ a superpixel-based approach to discard the use of deep features from the source model, since the deep features are error-prone for domain gaps. We propagate pseudo labels in these small clusters containing confident pixels under the assumption that each superpixel tends to share the same semantic label. 
In temporal diffusion, we establish temporal correspondence between near-far image pairs and propagate reliable labels from the near-view image to the corresponding pixels in the far-view image. 
In this way, pseudo labels can be diffused by adaptively combining the initial predictions with spatially and temporally diffused labels. 

To further boost the accuracy of re-training stage using the diffused pseudo labels, we propose two spatial and temporal constraint losses to drive the proximity of deep features of the same class. The spatial loss measures local spatial similarity to encourage identical features between pixels within a superpixel, while the temporal loss encourages the features of corresponding pixels in a near-far image pair to be identical through a contrastive way. By imposing these two newly proposed losses, we are able to strengthen the pixel-level similarity in an unsupervised manner, encouraging the model to better identify pixel similarity in the target domain. Please note that if the images do not form a sequence in the target domain, we can independently implement the spatial label diffusion and loss for self-training. 
 
In summary, the main contributions of this paper are:


1) We are the first to exploit spatial and temporal knowledge of the target domain for unsupervised domain adaption, 
and propose a pseudo label diffusion method named \textbf{TDo-Dif} that propagates the highly confident pseudo labels to the unlabeled pixels by establishing spatial similarity and temporal correspondence. 

2) We also propose two losses in the re-training stage to encourage the learned features to be similar within the same semantic category, which can effectively help the model to be better adapted to the target domain. 

3) Experimental results demonstrate that \textbf{TDo-Dif} can effectively balance the density and accuracy of diffused pseudo labels, and outperforms the state-of-the-art methods on real foggy scene understanding.

In the rest of the paper, we present the related work in Section II and the proposed pseudo label diffusion scheme and the self-training framework in Section III. In section IV, we elaborate experimental settings and extensive experimental results. Finally, conclusions are drawn in Section V.

\section{Related work}
In this section, we briefly review related work in each of the three sub-fields: semantic segmentation, semantic foggy scene understanding, unsupervised domain adaptive semantic segmentation.

\subsection{Semantic Segmentation}
As a widely studied means of inferring the semantic content of images, semantic segmentation can predict labels at pixel level. After the typical works such as FCN~\cite{long2015fully} and SegNet~\cite{badrinarayanan2017segnet} that employ CNNs to boost the semantic segmentation performance by extracting deep semantic features, most of the top-performing methods are built on CNNs. The subsequent works attempt to aggregate scene context to improve the performance of semantic segmentation, for example, by assembling multi-scale visual clues or convolutional features of different sizes~\cite{lin2019refinenet,mustafa2018msfd}, adopting atrous spatial pyramid pooling~\cite{chen2017deeplab}, employing neural attention to exchange context between paired pixels~\cite{conf/cvpr/ZhongLBHDLZLW20,journals/tmm/ZhangLCSSK19}, and iteratively optimizing the results using Markov decision process \cite{zhou2019context}. For a comprehensive overview of semantic segmentation algorithms, we point the reader to Taghanaki \emph{et al.}~\cite{journals/air/TaghanakiACCH21} or Shervin \emph{et al.}~\cite{seg_survey}.

Although impressive, most semantic segmentation models are trained with fully labeled and clear-weather datasets. Some general datasets include PASCAL VOC 2012 Challenge~\cite{everingham2010pascal} and MS COCO Challenge~\cite{conf/eccv/LinMBHPRDZ14} for visual object segmentation, and Cityscapes~\cite{cordts2016cityscapes} for urban scene understanding. However, the performance of current vision algorithms, even the best performing ones, undergoes a severe performance degradation under adverse conditions, which are crucial for outdoor applications. Therefore, this work focuses on  semantic foggy scene understanding, which is introduced below. 

\subsection{Semantic Foggy Scene Understanding}

An early work on semantic foggy scene understanding is SFSU \cite{sakaridis2018semantic}. It builds a synthetic foggy dataset by rendering the Cityscapes dataset~\cite{cordts2016cityscapes} with an atmosphere scattering model~\cite{journals/pami/He0T11}, and a follow-up semi-supervised learning approach based on RefineNet \cite{lin2017refinenet} is proposed. To improve the quality of synthetic fog, Hahner \emph{et al.} \cite{HahnerDSZG19} propose to utilize the pure synthetic data with segmentation labels, which is unrestricted in dataset size and employ accurate depth maps for fog synthesis. Similarly, Dai \emph{et al.} \cite{SakaridisDHG18} propose a curriculum model adaptation method called CMAda with two adaptation steps to learn from both synthetic and unlabeled real foggy images, and later extend this idea to accommodate multiple adaptation steps to gradually deal with light and dense fog  \cite{dai2020curriculum}. Although these methods improve model adaptation by generating pseudo labels in an easy-to-hard manner, their direct use of the noise predictions from models trained on the source domain as pseudo labels leads to degraded performance for re-training.

\subsection{Unsupervised Domain Adaptive Semantic Segmentation}
Unsupervised domain adaptation (UDA) has been extensively investigated in computer vision tasks~\cite{journals/pr/ChenH20,conf/cvpr/RoyChowdhuryCSJ19,conf/eccv/LiHQWG20}. 
The main idea for domain adaptation is to alleviate the gap between the source and target domains. For the semantic segmentation task, we expect to use domain adaptation to improve the segmentation performance of real images by models trained on synthetic images. To deal with this problem, adversarial training methods have received significant attention. These methods attempt to mitigate the domain gap from the image level, such as adversarial training based image-to-image translation~\cite{conf/cvpr/BousmalisSDEK17,DBLP:conf/cvpr/MurezKKRK18,Zhong2022GrayscaleEC}, or from the representation level, such as making features or network predictions indistinguishable between domains through a series of adversarial losses~\cite{conf/cvpr/Kang0YH19,conf/icml/LongZ0J17}. The focus of such approaches is on learning and aligning the shared knowledge between the source and target domains, but domain-specific information is usually ignored.

Self-training based UDA methods tend to improve segmentation performance in semi-supervised learning manner \cite{ZouYKW18,LianDLG19}. These works iteratively train the model by using both labeled source domain data and the generated pseudo labels from target domain to achieve alignment between the source and target domains. PyCDA~\cite{LianDLG19} constructs the pyramidal curriculum to provide various properties about the target domain to guide the training. To deal with the imbalance of pseudo labels between easy and hard classes, CBST~\cite{ZouYKW18} scheme is proposed, showing comparable domain adaptation performance to the adversarial training based methods. Later, CRST~\cite{conf/iccv/ZouYLKW19} is proposed to make self-training less sensitive to noise pseudo labels using soft pseudo labels and smoothing network predictions. It also integrates a variety of confidence regularizers to CBST. Furthermore, some works~\cite{conf/cvpr/LiYV19,conf/cvpr/KimB20a} resort to self-training as the second stage of training to further boost their adaptive performance.
However, the current self-training strategy relies heavily on the performance of predicted segmentation map of the pre-trained model, which is limited by the domain gap between the source and target domains.



\section{Proposed method}
\label{sec:method}
In this section, we describe the proposed pseudo label diffusion method for semantic foggy scene understanding. 
Below we first review the general self-training method and then provide an overview of our framework. Next, we dissect the individual modules and the associated objective functions for model re-training.

\subsection{Formulation of Self-training for UDA}
Following the common self-training setting in UDA, there are two datasets: a labeled dataset of source domain $\mathbb{X}_{S} = \{\mathbf{x}_{s}^{i}\}_{i=1}^{N_s}$ with its segmentation labels $\mathbb{Y}_{S} = \{\mathbf{y}_{s}^{i}\}_{i=1}^{N_s}$, and an unlabeled dataset of target domain $\mathbb{X}_{T} = \{\mathbf{x}_{t}^{i}\}_{i=1}^{N_t}$ without accessing its labels $\mathbb{Y}_{T}$, where $N_s$ and $N_t$ indicate the total number of images in the source and target domains, respectively. $\mathbf{y}_{s}^{i}(p)\in\{1,...,C\}$ is the label of pixel $p$ of image $\mathbf{x}_{s}^{i}$ and $C$ is the total number of classes. Unsupervised domain adaptive semantic segmentation assumes that the target domain shares the same $C$ classes with the source domain, and its aim is to train a segmentation network that transfers knowledge from the source dataset and achieves high performance on the target dataset.

Typically, a model trained on a source dataset with supervised labels can achieve satisfactory performance on test data from the same dataset, but it does not generalize well to the target data due to domain gaps. To transfer knowledge, self-training techniques use a pre-trained source model $\Phi$ to test on the target dataset and infer target pseudo labels,
$\hat{\mathbb{Y}}_{T} =\{\hat{\mathbf{y}}_{t}^{i}\}_{i=1}^{N_t}$. 
Then they re-train and optimize the model $\Phi$ by supervision of the source labels $\mathbb{Y}_{S}$ and the target pseudo labels $\hat{\mathbb{Y}}_{T}$. 

\begin{equation}
\label{segloss}
\begin{aligned}
\min _{w_\Phi} \mathcal{L}_{St}=&\frac{1}{N_s} \sum_{\mathbf{x}_{s} \in \mathbb{X}_{S}} \mathcal{L}_{seg}(\Phi(\mathbf{x}_{s}), \mathbf{y}_{s})+\\
&\alpha_t \frac{1}{N_t} \sum_{\mathbf{x}_{t} \in \mathbb{X}_{T}} \mathcal{L}_{seg}(\Phi(\mathbf{x}_{t}), \hat{\mathbf{y}}_{t}),
\end{aligned}
\end{equation}
where $\alpha_t$ is a hyper-parameter balancing the contributions of the two dataset, $w_\Phi$ is the model weights, and $\mathcal{L}_{seg}$ denotes the loss function for semantic segmentation.

Pseudo labels are typically generated from the model predictions by selecting the most confident labels. The original prediction $\tilde{\mathbf{y}}_{t}$ on the target image $\mathbf{x}_t$ is generated by: 

\begin{equation} 
\tilde{\mathbf{y}}_{t} = \mathop{\arg\max}_c p\left(c \mid \mathbf{x}_{t}, \Phi\right),
\end{equation}
where $p\left(c\mid\mathbf{x}_t, \Phi\right)$ is the probability of class $c$ in the model prediction. Since it only picks the most confident predictions as pseudo labels, which can be obtained finally by:

\begin{equation} 
\hat{\mathbf{y}}_{t}=
\begin{cases}
\tilde{\mathbf{y}}_{t} & \text{if}~p\left(c \mid \mathbf{x}_{t}, \Phi\right) > \lambda^c \\
0 &  \text{otherwise} 
\end{cases},
\end{equation}
where $\lambda^c$ denotes the confidence threshold for class $c$. This means that if the segmentation probability value of class $c$ is greater than $\lambda^c$, these pixels will be considered as confident regions (pseudo-labeled regions) and the rest will be ignored. $\lambda^c=0$ means that the whole predicted segmentation map is used as pseudo labels. For each class $c$, we determine $\lambda^c$ by the confidence value chosen from the most confident $p$ percentage of the prediction of class $c$ in the entire target set~\cite{ZouYKW18}. To ensure that the pseudo labels are mostly reliable, the hyper-parameter $p$ is usually set to a very low value, \emph{i.e.}, 0.2. 

\subsection{Framework Overview}
In this paper, we attempt to exploit the properties of the target domain data and introduce target domain knowledge into the self-training framework for UDA, which has been rarely exploited in current methods. In the driving scene, the cameras are usually mounted at the front window of a vehicle, facing forward. Along the travel path, the vehicle take a sequence of image. Thus, the objects along the road will appear in multiple images, and the object in a later image will be closer to the camera than in the previous images. In the foggy scene, the deep features of the pixels are changed by the fog, leading to a very sparse and discrete pseudo labels. Therefore, we exploit the spatial and temporal relations between pixels in the target domain images to diffuse the pseudo labels to unknown pixels so as to densify the pseudo labels. 

The framework is shown in Fig.~\ref{motivation}. The self-training process is solved by iteratively alternating the \textbf{pseudo-label generation} and \textbf{model re-training} steps in one self-training round.

\textbf{Step 1: Pseudo-label generation.} The pseudo labels of the target domain are initialized by inferring segmentation labels using the pre-trained model and selecting high confident labels from the predicted segmentation maps. Then, the initialized sparse pseudo labels are densified by label diffusion. We propose a novel label-diffusion method, \emph{i.e.}, \textbf{TDo-Dif}, to diffuse pseudo labels from spatial and temporal perspectives for improving the performance of self-training-based domain adaptive semantic segmentation. In \textit{spatial diffusion}, we adopt a superpixel-based clustering method to divide each image into small clusters. Under the assumption that the pixels within a superpixel are similar, the reliable pseudo labels is expanded in the superpixel regarding the presence of any reliable label in the clusters. In \textit{temporal diffusion}, we propose to register temporally adjacent frames by optical flow. Based on the property of the foggy data and established pixel-level correspondence, pseudo labels in the far-view images can be refined by their corresponding pixel labels in the near-view images.

\textbf{Step 2: Model re-training}. After the pseudo label diffusion, the diffused pseudo labels are then used to re-train the semantic segmentation model. In this step, based on the spatial similarity and temporal correspondence, two new loss functions, a local spatial similarity loss and a contrastive loss, are designed to constrain the feature similarity of the target domain to assist the unsupervised domain adaptive semantic segmentation.

In the following, we address each part of our framework in detail.

\begin{figure*}[tb]
    	\centering
    	\includegraphics[width=0.9\textwidth]{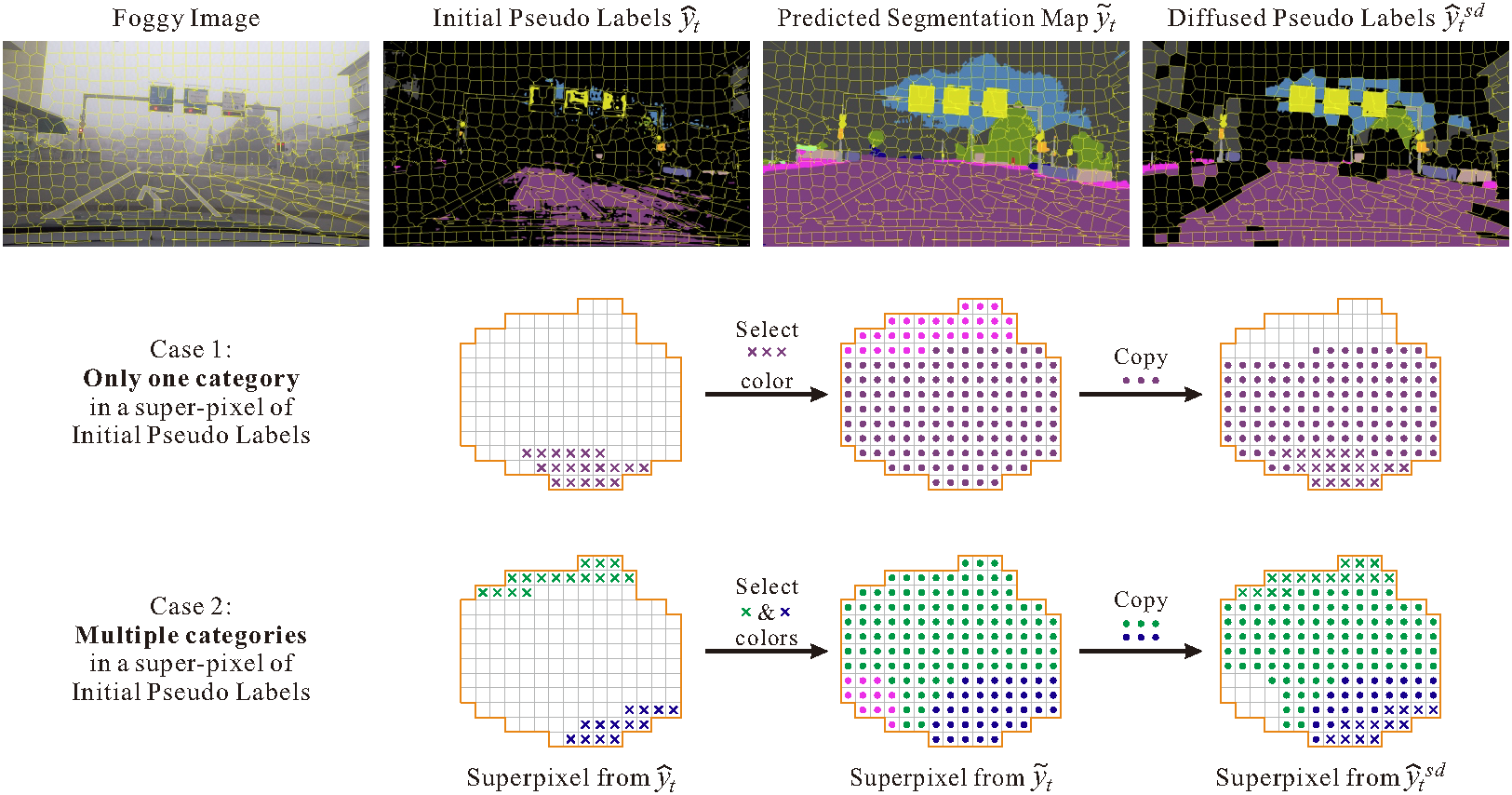}
	\caption{Illustration of the superpixel-based spatial diffusion, which is conducted in the ``pseudo-label generation'' step during each round of  re-training. Superpixels are first generated from the target foggy image, and then spatial diffusion is performed in a superpixel-wise order in the initial pseudo label map $\hat{\mathbf{y}}_{t}$. In \emph{Case 1}, if there is only one category in a superpixel, we copy all labels of that category within the superpixel from the initial predicted segmentation map $\tilde{\mathbf{y}}_{t}$ to the diffused superpixel map $\hat{\mathbf{y}}_{t}^{sd}$. This strategy is in line with the assumption that superpixel tends to cluster pixels of same category together, but also considers  the possible clustering errors in superpixels. In \emph{Case 2}, if there are more categories in a superpixel, we copy the labels of all existing categories from $\tilde{\mathbf{y}}_{t}$ to $\hat{\mathbf{y}}_{t}^{sd}$, considering that small objects tend to be mixed with large objects due to the set size of superpixels.}
    \label{fig:superpixel}
\end{figure*}


\subsection{Spatial Diffusion of Pseudo Labels}
In this section, we attempt to increase the number of pseudo labels by using spatial similarity within the target image. Considering that the learned model is not optimized for the target foggy scene, the discrimination of deep features cannot precisely reveal the semantic similarities in the foggy images. Therefore, we employ a superpixel-based clustering method to exploit the local spatial similarity present in target images.

\subsubsection{Spatial Clustering by Superpixels}
Superpixels~\cite{neubert2012superpixel} are defined by clustering a set of pixels with similar visual properties, \emph{e.g.}, pixel intensity, which have been used as pre-processing for lots of applications such as image stitching, image colorization and view synthesis~\cite{journals/tip/OliveiraSWJ21,journals/tgrs/YuanFZ21,journals/tvcg/FangWZZ20}. In this paper, we utilize the visual similarity property within a superpixel to extend reliable pseudo labels. Specifically, we adopt the well-known SLIC algorithm~\cite{achanta2012slic} to generate superpixels, which generates superpixels by $k$-means clustering strategy. SLIC measures the color and spatial difference between each pixel and its superpixel center is defined as:

\begin{equation}
D(p_{sp_i}, sp_i^{ctr})=(\frac{d_c}{M_c})^{2}+(\frac{d_s}{M_s})^{2},
\end{equation}
where $d_c$ and $d_s$ denote the color distance and spatial distance between pixel $p_{sp_i}$ and the superpixel center $sp_i^{ctr}$ in the $i$-th superpixel $sp_i$, respectively. $M_c$ is the maximum color distance, which is usually set as a constant. $M_s=\sqrt{M / K}$, $M$ denotes the total number of pixels of this image and $K$ is a hyper-parameter indicating the number of superpixels, which is the only parameter we need to set.


After the spatial clustering, the target image $\mathbf{x}_{t}$ is divided into a superpixel set $\mathcal{SP}=\{sp_i\}_{i=1}^S$, where $S$ is the total number of superpixels.

\subsubsection{Superpixel-based Spatial Label Diffusion}

The spatial label diffusion process starts with the initial pseudo labels, which are collected from the high confident labels of the original predicted segmentation map $\tilde{\mathbf{y}}_{t}$.
The superpixels containing initial pseudo labels in the set $\mathcal{SP}$ are picked out for further spatial label diffusion. We then densify the labels in each picked superpixel by diffusing the labels from the original predicted segmentation map, under the assumption that visually clustered pixels within a superpixel have a high probability of belonging to the same class.

The particular diffusion process of each superpixel $sp_i$ is presented in Fig.~\ref{fig:superpixel}, and can be formulated as follows:

\begin{equation}
\hat{\mathbf{y}}_{sp_i}^{sd}(p)=
\begin{cases}
\tilde{\mathbf{y}}_{sp_i}(p) & \text{if}~\exists~ \hat{\mathbf{y}}_{sp_i}(q)=\tilde{\mathbf{y}}_{sp_i}(p)\\
0 & \text{otherwise}
\end{cases}
\end{equation}
where $\tilde{\mathbf{y}}_{sp_i}$ and $\hat{\mathbf{y}}_{sp_i}$ are the predicted labels and initial pseudo labels of superpixel $sp_i$, and $p$ and $q$ denote a pixel within $sp_i$ of these maps, respectively. All the pixels in the target superpixel $sp_i$ are traversed to check whether the label of a pixel $p$ in $\tilde{\mathbf{y}}_{sp_i}$ is the same as the label of any pixel $q$ in the initial pseudo labels $\hat{\mathbf{y}}_{sp_i}$. If yes, we copy the label of $\tilde{\mathbf{y}}_{sp_i}(p)$ to $\hat{\mathbf{y}}_{sp_i}^{sd}(p)$ so as to diffuse the initial pseudo label on $q$ to $p$ in $\hat{\mathbf{y}}_{sp_i}^{sd}$. The densified pseudo labels of all superpixels $\hat{\mathbf{y}}_{sp_i}^{sd}$ form the diffused pseudo label $\hat{\mathbf{y}}^{sd}$. 


Considering that there could still be some clustering errors in a superpixel, especially in the case of small objects, we do not directly diffuse the class of high-confidence pixels to the entire superpixel, but only pass the same class present in the predicted segmentation map $\tilde{\mathbf{y}}_{sp_i}$ to the pseudo labels $\hat{\mathbf{y}}_{sp_i}^{sd}$ (Fig.~\ref{fig:superpixel} \textit{Case 1}). 
In case there exist pseudo labels for multiple classes in one superpixel, we diffuse all existing classes. This is especially designed to handle small objects which are usually smaller than the size of a superpixel (Fig.~\ref{fig:superpixel} \textit{Case 2}).

\begin{figure*}[tb]
     \centering
     \includegraphics[width=\textwidth]{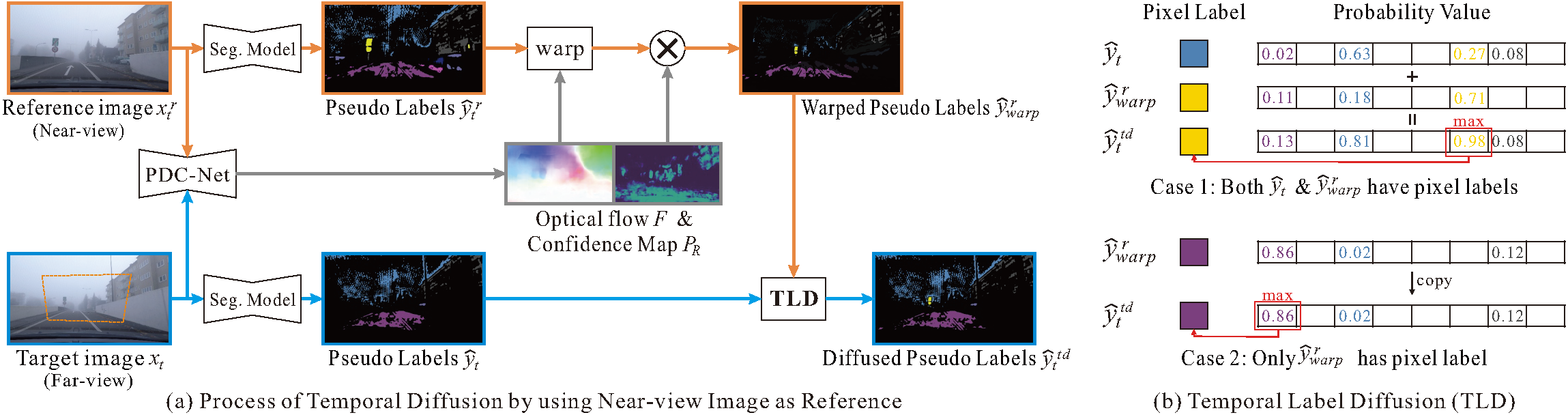}
     \caption{Illustration of the optical flow-based temporal diffusion. The near-view image is taken as the reference image to propagate its warped pseudo labels to optimize the pseudo labels of the target image, under the assumption that the near-view image is less affected by fog. (a) shows the overall process the temporal diffusion, and (b) shows the label diffusion on each pixel.}

     \label{temporal}
\end{figure*}


\subsection{Temporal Diffusion of Pseudo Labels}
We exploit temporal correspondence between neighboring images to densify pseudo labels. It is motivated by the assumption that the closer an object is to the camera, the higher resolution and the less it is affected by fog, thus resulting in a more reliable prediction. 
On this basis, we diffuse the labels of an object from a closer view to a more distant view. 

\subsubsection{Temporal Correspondence by Optical Flow}
We take a near-far view pair for the temporal diffusion, where the far-view image is the target image denoting as $\mathbf{x}_t$ and the close-view image is the diffusion reference denoting as $\mathbf{x}^{r}_t$. The first step for temporal diffusion is building the temporal correspondence between the image pair. We employ one of the latest optical flow estimation methods, PDC-Net~\cite{truong2021learning}, to estimate the dense flow field with a coupled pixel-wise confidence map to indicate the reliability and accuracy of the flow field prediction. 
The optical estimation process can be simplified to the following form:

\begin{equation}
    F, P_F = \phi(\mathbf{x}^{r}_t, \mathbf{x}_t),
\end{equation}
\begin{equation}
M_F(p)=
\begin{cases}
1 &  P_F(p)>T \\
0 &  \text{otherwise} 
\end{cases},
\end{equation}
where $F=(u,v)\in\mathbb{R}^2$ is the predicted flow vector from $\mathbf{x}^{r}_t$ to $\mathbf{x}_t$. $P_F$ and $M_F$ are the confidence map and the confident binary map of the true flow, respectively. $p$ is the pixel index. The threshold $T$ is set to 0.5, the same as in \cite{truong2021learning}.


\subsubsection{Flow-based Temporal Label Diffusion}

The illustration of flow-based temporal diffusion is shown in Fig.~\ref{temporal}. After generating the temporal correspondence, we warp the pseudo labels $\hat{\mathbf{y}}_t^r$ and the predicted segmentation probability map $p_t^r(\mathbf{x}_t^{r}, \Phi)$ of $\mathbf{x}_{t}^{r}$ by the guidance of the estimated flow $F$ and its confident binary map $M_F$.

\begin{equation}
    \hat{\mathbf{y}}^r_{warp} = M_F \odot warp(\hat{\mathbf{y}}_t^r, F),
\end{equation}
\begin{equation}
    p^r_{warp}\left(\mathbf{x}_t^{r}, \Phi\right) = M_F \odot warp(p_t^r\left(\mathbf{x}_t^{r}, \Phi\right), F),
\end{equation}
where $warp(\cdot)$ denotes a warping function and $\odot$ denotes element-wise multiplication.  $M_F$ acts as a binary mask.

After aligning the pseudo labels of the reference image to the target image, we handle the temporal diffusion according to the following situations: 

\begin{enumerate}
\item If the warped pseudo label in $\hat{\mathbf{y}}^r_{warp}$ falls on a pixel $p$ that already has a label in $\hat{\mathbf{y}}_t$, we update the pseudo label for $\hat{\mathbf{y}}_t(p)$ by fusing the soft segmentation prediction of semantic segmentation and re-select the class label with the highest confidence score. 
\begin{equation}
\hat{\mathbf{y}}_t^{td} = \mathop{\arg\max}_c (p(c \mid \mathbf{x}_t, \Phi)+p^r_{warp}\left(c \mid \mathbf{x}_t^{r}, \Phi\right)).
\end{equation}

\item If the warped pseudo label in $\hat{\mathbf{y}}^r_{warp}$ falls on a pixel $p$ and there is no specified pseudo label in $\hat{\mathbf{y}}_t$, we copy the label from $\hat{\mathbf{y}}^r_{warp}(p)$ to $\hat{\mathbf{y}}_{t}(p)$. It increases the number of pseudo labels. 

\begin{equation}
\hat{\mathbf{y}}_{t}^{td} = \hat{\mathbf{y}}^r_{warp}.
\end{equation}

\end{enumerate}

\begin{algorithm}[t]
\caption{{ Training and Testing Step of \textbf{TDo-Dif} (Temporal Diffusion Followed by Spatial Diffusion)}}\label{algorithm}
\SetKwInOut{Input}{{Input}}\SetKwInOut{Output}{ {Output}}
\tcp{{Training procedure}}
\Input{{1. Unlabeled target dataset $\mathbb{X}_{T} = \{\mathbf{x}_{t}^{i}\}_{i=1}^{N_t}$;\\
\vspace{1mm}
2. Labeled source dataset $\mathbb{X}_{S} = \{\mathbf{x}_{s}^{i}\}_{i=1}^{N_s}$ with \\ its labels $\mathbb{Y}_{S} = \{\mathbf{y}_{s}^{i}\}_{i=1}^{N_s}$;\\
3. Semantic segmentation model $\Phi$ that has been pretrained on \{$\mathbb{X}_{S}$, $\mathbb{Y}_{S}$\}.}}
\Output{ {Domain adaptive model $\Phi$ on target domain.}}

\SetKwFor{While}{ {while}}{ {do}}{ {end}}

\While{ {Iter=1 to num\_round}}
{
 {1) Generate the original prediction $\tilde{\mathbf{y}}_{t}$ for target training image $\mathbf{x}_t$ with $\Phi$ by Eq.~(2);\\
2) Pick the confident pseudo labels $\hat{\mathbf{y}}_{t}$ according to $\tilde{\mathbf{y}}_{t}$ and $\lambda^c$ by Eq.~(3);\\
3) Perform temporal diffusion on $\hat{\mathbf{y}}_{t}$, and generate the temporal densified pseudo labels $\hat{\mathbf{y}}_t^{td}$ by Eq.~(10) and Eq.~(11);\\
4) Perform spatial diffusion on $\hat{\mathbf{y}}_t^{td}$, and generate the final densified pseudo labels $\hat{\mathbf{y}}_t$ by Eq.~(5), then target pseudo labels $\mathbb{\hat{Y}}_{T}=\{\hat{\mathbf{y}}_{t}^{i}\}_{i=1}^{N_t}$};

 {5)} \While{ {epoch=1 to num\_epoch}}
{ {a) Calculate $\mathcal{L}_{St}$, $\mathcal{L}_{Spa}$, and $\mathcal{L}_{Tem}$ based on $\mathbb{X}_{T}$ and $\mathbb{\hat{Y}}_{T}$;\\
b) Update $\Phi$ according to Eq.~(16);}}}
\BlankLine

\tcp{ {Testing procedure}}

\Input{ {1. Testing image $\mathbf{x}_{tst}$ of target domain;\\
        2. Domain adaptive model $\Phi$.}}
\Output{ {The predicted segmentation map $\mathbf{y}_{tst}$.}}
 {1) Feed target testing image $\mathbf{x}_{tst}$ into model $\Phi$;\\
2) Generate the semantic segmentation map $\mathbf{y}_{tst}$.}\\
\end{algorithm}

\subsection{Re-training Strategy}
After obtaining the diffused pseudo labels, we re-train the segmentation model to adapt it to the target domain. To further utilize the knowledge of the target domain, we propose two new losses for the model re-training. 

We first propose a local spatial similarity loss based on the assumption that a superpixel should belong to a single object. In particular, we encourage the distances between the pixel features in each superpixel to be small. Let $f(x_t)$ represent the deep feature map of image $\mathbf{x}_t$, and $sp_i\downarrow$ is the downsamped version of the superpixel with the same spatial resolution of $f(x_t)$. $sp_i\downarrow$ is used to indicate the coressponding locations of the feature map for the superpixel $sp_i$. We define the feature distances in a superpixel as follows:

\begin{equation}
\begin{aligned}
    D_{sp_i\downarrow}&=\frac{1}{|sp_i\downarrow|}\sum_{p\in sp_i\downarrow} D_c(f(p), \eta_{sp_i\downarrow})\\
    &=\frac{1}{|sp_i\downarrow|}\sum_{p\in sp_i\downarrow} <\frac{f(p)}{||f(p)||},\frac{\eta_{sp_i\downarrow}}{||\eta_{sp_i\downarrow}||}>,
\end{aligned}
\end{equation}

\begin{equation}
    \eta_{sp_i\downarrow}=\frac{1}{|sp_i\downarrow|}\sum_{p\in sp_i\downarrow} f(p),
\end{equation}
where $f(p)$ is the deep feature at pixel $p$, and $\eta_{sp_i\downarrow}$ is the feature centroid of $D_{sp_i\downarrow}$. We adopt $D_c$ as the cosine similarity between features to measure the pixel-wise similarity, so that $D_{sp_i\downarrow}$ is the average feature distance. Then we define the local spatial similarity loss as:

\begin{equation}
    \mathcal{L}_{Spa}=\frac{1}{S}\sum_{i=1}^S D_{sp_i\downarrow}.
\end{equation}

Temporally, we try to maximize the similarity of associated pixels via a contrastive loss. The objective is to distinguish between the tight correspondence (different views of the same pixel) and incompatible one (different views of different pixels). For each pixel $p$ which has found a correspondence $p'$, we construct a set of random negative pixels $\mathcal{N}$. Then the temporal loss function for pixel $p$ and $\mathcal{N}$ can be written as:

\begin{equation}
        \begin{aligned}
    &\mathcal{L}_{Tem}=\\
     &-\frac{1}{||\mathcal{P}||}\sum_{p\in\mathcal{P}} log\frac{e^{D_c(f(p),f(p'))}}{e^{D_c(f(p),f(p'))}+\sum_{n\in \mathcal{N}}e^{D_c(f(p),f(n)))}}
    \end{aligned},
\end{equation}
where $\mathcal{P}$ is the set of positive pixels.

Therefore, the overall training loss function for model re-training is defined as the weighted sum of the segmentation loss $\mathcal{L}_{St}$, 
the spatial loss $\mathcal{L}_{Spa}$ and the temporal loss $\mathcal{L}_{Tem}$:

\begin{equation}
    \mathcal{L}_{Final} = \mathcal{L}_{St}+\alpha_{spa}\mathcal{L}_{Spa}+\alpha_{tem}\mathcal{L}_{Tem},
\end{equation}
where $\alpha_{spa}$ and $\alpha_{tem}$ are the weights for the local spatial similarity loss and temporal contrastive loss, respectively.  {The details about the overall training and testing
procedure can be seen in Algorithm~\ref{algorithm}}.

\section{Experimental results}
\label{sec:results}

\subsection{Experimental Settings}

\subsubsection{Datasets}
We validate our method on real foggy datasets, \textbf{Foggy Zurich}~\cite{dai2020curriculum} and \textbf{Foggy Driving}~\cite{sakaridis2018semantic}, as the target domain datasets. The source domain for the pre-training is the synthetic dataset \textbf{Foggy Cityscapes}~\cite{sakaridis2018semantic} that derived from \textbf{Cityscapes}~\cite{cordts2016cityscapes} with fine segmentation labels. 


\textbf{Foggy Cityscapes}~\cite{sakaridis2018semantic}. It is synthesized with the atmospheric scattering model~\cite{journals/pami/He0T11} for fog simulation based on \textbf{Cityscapes}. Three distinct fog densities are generated by controlling a constant simulated attenuation coefficient $\beta\in \{0.005,0.01,0.02\}$, where higher $\beta$ corresponding to denser fog. Each version shares the same semantic labels with \textbf{Cityscapes}, which contains 5000 images ($2048\times1024$) belonging to 20 categories with 2975 images for training, 500 for validation, and 1525 for testing.

\textbf{Foggy Zurich}~\cite{dai2020curriculum}. It is a realistic foggy scene dataset consisting of 3808 images with a resolution of $1920\times1080$ collected in the city of Zurich and its suburbs. It was recorded as four large video sequences, and sequential images were extracted at a rate of one frame per second. It provides pixel level annotations for 40 images with dense fog. We use the 40 annotated images for testing and the rest without annotations for training.

\textbf{Foggy Driving}~\cite{sakaridis2018semantic}. It consists of 101 color images depicting real-world foggy driving scenes. 51 of these images were taken with a cell phone camera under foggy conditions at different areas of Zurich, and the remaining 50 images were collected from the web. Image resolutions in the dataset range from $500\times339$ to $1280\times960$.

\begin{table*}[htb]
\small
\renewcommand\tabcolsep{4pt}
\centering
\caption{ {Quantitative comparison on \textbf{Foggy Zurich} dataset. Notice that we exclude the statistics of class \emph{Train} because it is not included in the test set, and we calculate class \emph{Truck} although it cannot be detected by any method.} The numbers in \textbf{\rc{red}} and \textbf{\bc{blue}} represent the best and second best scores.}
\linespread{1.3}
\resizebox{1.0\textwidth}{!}
{
\begin{tabular}{c|cccccccccccccccccc|c}
\hline            
Method& Road & SW & Build & Wall & Fence & Pole & TL & TS & Veg. & Terrain & Sky & PR & Rider & Car & Truck & Bus & Motor & Bike & mIoU\\
\hline
SFSU~\cite{sakaridis2018semantic} & 64.94 & 51.25 & 37.95 & 28.46  & 20.87 & 41.69   & \textbf{\bc{60.13}}  & 55.11 & 34.20  & 31.22 & 27.03 & 3.55 & 38.04  & 77.54 & 0.00 & 11.08  &  13.63 & 4.75  & 33.41            \\
CMAda~\cite{dai2020curriculum} & 84.68  &	57.65 &	42.27 &	27.03 &	21.28 &	\textbf{\rc{47.77}} &	\textbf{\rc{61.88}} &	\textbf{\bc{62.95}} &	62.26 &	35.73 &	70.93 &	8.47 &	35.92 &	85.03 & 0.00 & \textbf{\bc{45.32}} &	37.65 &	9.97  & 44.27 \\
AdSegNet~\cite{conf/cvpr/TsaiHSS0C18} & 21.31 &	31.53 &	26.11 &	14.84 &	23.45 &	30.64 &	48.52 &	46.75 &	56.69 &	22.78 &	43.52 &	3.51 &	20.34 &	10.86 &	0.00 &	4.20 	& 38.49 &	4.00 & 24.86 \\
CBST~\cite{ZouYKW18} & \textbf{\rc{91.64}} & 57.09 & 29.92 &	55.96 &	31.54 &	42.32 &	54.88 & 	60.07 &	73.35 &	53.61 &	52.62 &	\textbf{\bc{8.71}} &	43.06 & \textbf{\bc{87.52}} & 0.00 &	16.60 &53.90 &	11.48  & 45.79  \\
CRST~\cite{zou2019confidence} & 91.16 &	57.81 &	36.23 &	54.53 &	31.19 &	41.99 &	51.43 & \textbf{\rc{63.29}}	 &	75.04 &	54.40 &	61.21 &	7.84 &	40.92 &	86.89 &	0.00 &	25.36 &		45.01 &	12.21 & 46.47  \\

CuDA-Net~\cite{xianzheng}&91.47&51.64&40.07&55.99&28.37&\textbf{\bc{46.38}}&58.22&63.07&\textbf{\rc{77.38}}&59.47&67.9&2.87&\textbf{\rc{45.74}}&86.74&0&\textbf{\rc{54.52}}&50.72&3.98&49.14\\

FIFO~\cite{fifo22}&-&-&-&-&-&-&-&-&-&-&-&-&-&-&-&-&-&-&48.40\\
CMDIT~\cite{iccvw21}&-&-&-&-&-&-&-&-&-&-&-&-&-&-&-&-&-&-&41.69\\
FogAdapt+~\cite{Iqbal2020Invariance}&-&-&-&-&-&-&-&-&-&-&-&-&-&-&-&-&-&-&49.80\\
\hline
\textbf{TDo-Dif} (SD) & 91.01 & 55.03 &	66.95 &	52.11 &	36.14 &	35.53 & 	54.24 &	60.05 &	73.35 &	57.96 &	\textbf{\rc{90.58}} &	6.67 &	42.70 &	84.44 & 0.00 &	40.61 &57.42 &	7.23 & 50.67  \\
\textbf{TDo-Dif} (SD+SL) & 89.52 &	53.29 &	66.69 &	\textbf{\bc{56.65}} &	39.97 &	36.92 & 	53.22 &	59.20 &	73.94 &	58.48 &	\textbf{\bc{90.26}} &	4.45 &	33.19 &	83.79 & 0.00 &	42.03  & 57.86 &	\textbf{\rc{18.10}} & 50.92 \\
\textbf{TDo-Dif} (TD) & \textbf{\bc{91.57}} & \textbf{\rc{58.72}} &	34.12 &	54.28 &	31.86 &	41.00 & 	50.87 &	60.91 &	74.64 &	\textbf{\bc{59.10}} &	58.45 &	\textbf{\rc{9.49}} &	44.32 &	\textbf{\rc{87.60}} & 0.00 &	26.21 &	47.03 &	12.19 & 46.80 \\
\textbf{TDo-Dif} (TD+TL) & 90.89 &	\textbf{\bc{58.58}} &	41.14 &	54.86 &	28.59 &	42.19 & 	51.83 &	59.49 &	74.93 &	57.40 &	68.78 &	6.17 &	41.22 &	86.66 & 0.00 &	40.72 &		42.17 &	12.56 & 47.68 \\
\makecell[c]{\textbf{TDo-Dif} \\ (SD$\rightarrow$TD+SL+TL)} & 89.81 &	54.38 &	\textbf{\rc{68.79}} &	\textbf{\rc{56.70}} &	\textbf{\bc{44.25}} &	36.54 & 	51.72 &	58.85 &	74.48 &	\textbf{\rc{59.66}} &	90.12 &	4.19 &	44.09 &	84.82 & 0.00 & 41.51 &	\textbf{\bc{58.06}}	 &	13.85 & \textbf{\bc{51.77}} \\
\makecell[c]{\textbf{TDo-Dif} \\ (TD$\rightarrow$SD+SL+TL)} & 89.41 &	51.43 &	\textbf{\bc{67.49}} &	55.93 &	\textbf{\rc{44.64}} &	37.70 &	52.06 &	60.10 &	\textbf{\bc{75.23}} &	58.48 &	89.79 &	4.88 &	\textbf{\bc{44.68}} &	86.17  & 0.00 & 42.03 &	\textbf{\rc{59.15}} &	\textbf{\bc{15.38}} & \textbf{\rc{51.92}} \\\hline

\end{tabular}
}
\label{tab:cityscape2z}
\end{table*}

\begin{table*}[htb]
\small
\renewcommand\tabcolsep{4pt}
\centering
\caption{{Quantitative comparison on \textbf{Foggy Driving} dataset}. The numbers in \textbf{\rc{red}} and \textbf{\bc{blue}} represent the best and second best scores.}

\linespread{1.3}
\resizebox{1.0\textwidth}{!}
{
\begin{tabular}{c|ccccccccccccccccccc|c}
\hline            
Method& Road & SW & Build & Wall & Fence & Pole & TL & TS & Veg. & Terrain & Sky & PR & Rider & Car & Truck & Bus & Train & Motor & Bike & mIoU\\
\hline
SFSU~\cite{sakaridis2018semantic} & 90.26 &	28.83 &	72.13 &	25.23 &	13.41 &	42.84 &	52.03 & 	58.97 &	64.27 &	5.78 &	76.71 &	57.26 &	44.02 &	70.41 &	13.42 & 27.73 &	58.48 &	19.29 &	46.48 & 45.66             \\
CMAda~\cite{dai2020curriculum} & 91.51 &	29.24 &	74.77 &	28.37 &	15.10 &	\textbf{\rc{49.36}} &	51.35 & 59.26 &	\textbf{\bc{74.76}} &	7.82 &	\textbf{\bc{92.29}} &	62.63 &	47.67 &	72.90 &	19.38 & 	\textbf{\bc{32.48}} &	52.05 &	24.62 &	\textbf{\bc{52.81}} & 49.39 \\
AdSegNet~\cite{conf/cvpr/TsaiHSS0C18} & 45.82 &	13.52 &	43.34 &	0.63 &	8.94 &	25.97 &	37.57 & 	35.92 &	54.12 &	0.53 &	80.70 &	30.73 &	27.08 &	56.74 &	0.73 &	12.58 & 0.40 &	11.19 &	26.47 &27.00 \\

CBST~\cite{ZouYKW18} & 91.68 & 31.35 & 68.63 &	25.61 &	15.98 &	48.14 &	49.48 & 	\textbf{\bc{60.02}} &	67.85 &	10.37 &	82.18 & 62.22 &	41.62 &	73.30 &	36.96 & 	15.69 & 31.69 & 29.90 &	46.95  & 46.82  \\
CRST~\cite{zou2019confidence} & 91.82 &	36.34 &	70.59 &	23.93 &	16.33 &	46.02 &	49.66 & 	56.92 &	70.84 &	12.68 &	86.36 &	64.25 &	42.17 &	\textbf{\rc{75.07}} &	30.72 & 	13.24 &	31.32 &	\textbf{\bc{35.06}} &	45.70 & 47.32  \\

CuDA-Net~\cite{xianzheng}&90.14&\textbf{\rc{45.52}}&71.47&\textbf{\rc{43.63}}&\textbf{\rc{44.23}}&43.83&46.30&52.24&72.63&\textbf{\rc{36.18}}&91.19&59.90&47.90&72.04&\textbf{\rc{48.58}}&\textbf{\rc{40.96}}&32.81&33.47&44.09&\textbf{\bc{53.50}}\\

FIFO~\cite{fifo22}&-&-&-&-&-&-&-&-&-&-&-&-&-&-&-&-&-&-&-&50.70\\
CMDIT~\cite{iccvw21}&-&-&-&-&-&-&-&-&-&-&-&-&-&-&-&-&-&-&-&45.35\\
FogAdapt+~\cite{Iqbal2020Invariance}&-&-&-&-&-&-&-&-&-&-&-&-&-&-&-&-&-&-&-&52.40\\ \hline

\makecell[c]{\textbf{TDo-Dif$^\dag$} \\ (SD$\rightarrow$TD+SL+TL)} & \textbf{\bc{92.12}} & 31.84 &	74.82 &	28.44 &	17.91 &	43.72 &	\textbf{\bc{52.45}} &	56.12 &	71.40 &	12.86&	86.71 &	\textbf{\bc{64.31}} &	\textbf{\bc{47.98}} &	73.11 &	38.40 &	23.19 &	55.34 &	27.43 &	\textbf{\rc{53.46}} & 50.08  \\
\makecell[c]{\textbf{TDo-Dif$^\dag$} \\ (TD$\rightarrow$SD+SL+TL)} & 92.09 &	31.80 &	\textbf{\bc{74.87}} &	27.72 &	17.93 &	45.01 &	\textbf{\rc{52.77}} &	57.92 &	71.33 &	\textbf{\bc{13.73}} &	86.63 &	\textbf{\rc{64.62}} &	\textbf{\rc{48.09}} &	\textbf{\bc{73.69}} &	38.69 &	24.59 &	\textbf{\bc{60.25}} &	28.11 &	52.45 & 50.65  \\\hline

\makecell[c]{\textbf{TDo-Dif$^\star$} (SD+SL)}&\textbf{\rc{93.03}}&\textbf{\bc{39.26}}&\textbf{\rc{76.72}}&\textbf{\bc{33.35}}&\textbf{\bc{18.77}}&\textbf{\bc{48.35}}&50.17&\textbf{\rc{64.41}}&\textbf{\bc{79.99}}&2.32&\textbf{\rc{92.66}}&61.87&46.64&78.31&\textbf{\bc{44.63}}&28.22&\textbf{\rc{70.78}}&\textbf{\rc{41.58}}&51.58&\textbf{\rc{53.84}} \\ \hline
\multicolumn{21}{l}{\makecell[l]{{\textbf{TDo-Dif$^\dag$} and \textbf{TDo-Dif$^\star$} denote the results from the model trained on \textbf{Foggy Zurich} and \textbf{Foggy Driving}, respectively.} \\{Note that the images in \textbf{Foggy Driving} dataset are non-sequential images, thus we only use the spatial diffusion and spatial loss.}}}\\\hline
\end{tabular}
}
\label{tab:cityscape2d}
\end{table*}

\subsubsection{Baseline Methods}
 {We compare our method with the state-of-the-art foggy scene understanding methods and domain adaptive segmentation methods. Among them, \textbf{SFSU}~\cite{sakaridis2018semantic} is a supervised learning-based approach on foggy scene understanding by training  segmentation model with the labeled source dataset \textbf{Foggy Cityscapes}. \textbf{CMAda}~\cite{dai2020curriculum} re-trains the segmentation model with the labeled source dataset \textbf{Foggy Cityscapes} and unlabeled real foggy weather dataset \textbf{Foggy Zurich}, gradually adapting the model from light fog to dense fog in multiple steps. \textbf{AdSegNet}~\cite{conf/cvpr/TsaiHSS0C18} is a representative adversarial learning strategy for domain adaptive semantic segmentation. \textbf{CBST}~\cite{ZouYKW18} and \textbf{CRST}~\cite{zou2019confidence} are two representative self-training strategies for domain adaptive semantic segmentation.
Recent methods \textbf{CuDA-Net}~\cite{xianzheng}, \textbf{FIFO}~\cite{fifo22} and \textbf{CMDIT}~\cite{iccvw21} focus on bridging the domain gap between the clear images and foggy images to improve the performance of foggy scene segmentation. \textbf{FogAdapt+}~\cite{Iqbal2020Invariance} combines the scale-invariance and uncertainty to minimize the domain shift in foggy scenes segmentation.}

 {We directly use the released models of \textbf{SFSU}\footnote{http://people.ee.ethz.ch/\textasciitilde csakarid/SFSU\_synthetic/} and \textbf{CMAda}\footnote{http://people.ee.ethz.ch/\textasciitilde csakarid/Model\_adaptation\_SFSU\_dense/} and re-train the segmentation models of the  three domain adaptive training strategies, including \textbf{AdSegNet}, \textbf{CBST} and \textbf{CRST}, on our dataset. We collect the results of \textbf{CuDA-Net} from its authors and the results of the last three methods from their papers.}

\begin{table}[tb]
\centering
\caption{Settings of different variants}
\setlength\tabcolsep{3pt}
\linespread{1.5}
\resizebox{1.0\columnwidth}{!}
{
\begin{tabular}{|c|c|l|}
\hline
~& Variants & \makecell[c]{Characteristics} \\ \hline
\multirow{4}{*}{\textbf{Group I}}&\textbf{TDo-Dif} (SD)&Superpixel-based spatial diffusion only \\\cline{2-3}
~&\textbf{TDo-Dif} (SD+SL)&Superpixel-based spatial diffusion plus spatial loss\\\cline{2-3}
~&\textbf{TDo-Dif} (TD)&Flow-based temporal diffusion only\\\cline{2-3}
~&\textbf{TDo-Dif} (TD+TL)&Flow-based temporal diffusion plus temporal loss\\ \hline
\multirow{2}{*}{\textbf{Group II}}&\makecell[c]{\textbf{TDo-Dif} \\ (SD$\rightarrow$TD+SL+TL)}&Spatial diffusion followed by temporal diffusion \\\cline{2-3}
~&\makecell[c]{\textbf{TDo-Dif} \\ (TD$\rightarrow$SD+SL+TL)}&Temporal diffusion followed by spatial diffusion\\ \hline
\end{tabular}}
\label{variants}

\end{table}

\subsection{Implementation Details}
Similar with \textbf{SFSU} and \textbf{CMAda}, we adopt the RefineNet \cite{lin2019refinenet} with ResNet-101 as the backbone in all experiments and initialize it with the weights pre-trained by \textbf{Foggy Cityscapes}. We implement our method using the Pytorch toolbox and optimize it using Adam algorithm with $\beta_1=0.5$, $\beta_2=0.999$, and a learning rate of 0.0001 following \cite{sakaridis2018semantic}.  {In all experiments, we use a batch size of 2 and set the self-training round number to 4 and the training epochs in each round to 10}. The loss weights $\alpha_t$, $\alpha_{spa}$, and $\alpha_{tem}$ are set to 1, 0.1 and 5, respectively. As similar to \textbf{CBST} and \textbf{CRST}, we set the hyper-parameter $p$ of confident portion to 0.2 so as to pick the top 20\% of high confidence predictions as pseudo labels. We set the number of superpixels $K$ in each images to 500. The threshold $T$ for reliable dense matching is set to 0.5, the same as in \cite{truong2021learning}. The number of positive samples used for computing the temporal contrasive loss is set to 20, which are randomly selected from pixels with temporal correspondence. The number of negative samples for each positive pixels is set to 1, chosen from pixels of different predicted categories.


\tabcolsep=0.5pt
\begin{figure*}[tb]
	\centering
	\linespread{0.5}
\small{
		\begin{tabular}{ccccc}
			\rotatebox{90}{~~~~Foggy image} &
			\includegraphics[width=0.235\textwidth]{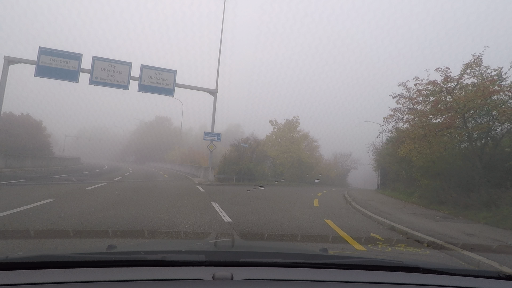} &
			\includegraphics[width=0.235\textwidth]{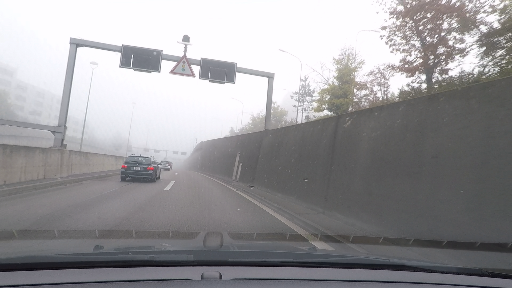} &
			\includegraphics[width=0.235\textwidth]{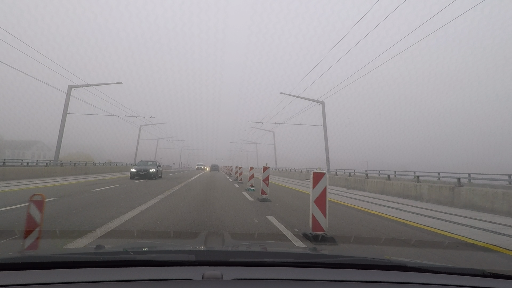} &
			\includegraphics[width=0.235\textwidth]{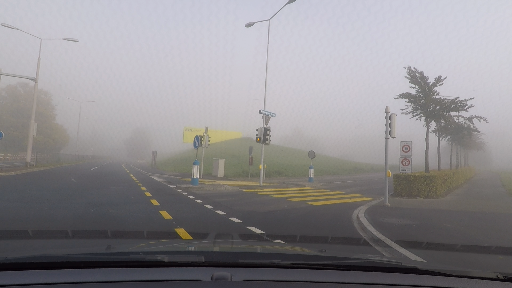} \\
			\rotatebox{90}{~~~~~SFSU~\cite{sakaridis2018semantic}} &
			\includegraphics[width=0.235\textwidth]{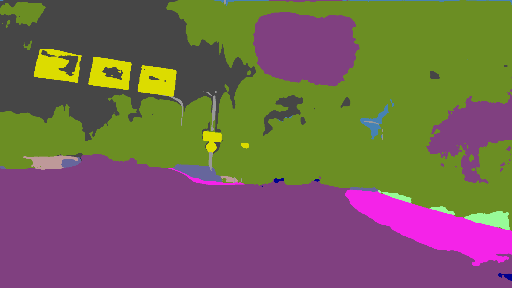} & 
			\includegraphics[width=0.235\textwidth]{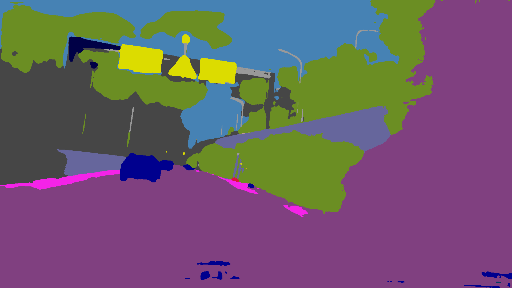} & 
			\includegraphics[width=0.235\textwidth]{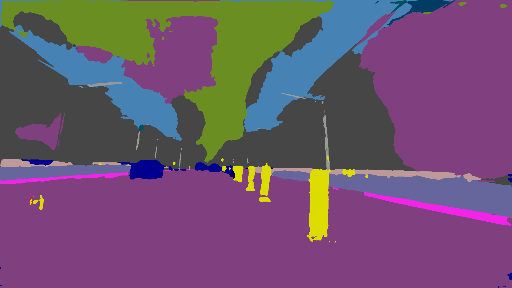} & 
			\includegraphics[width=0.235\textwidth]{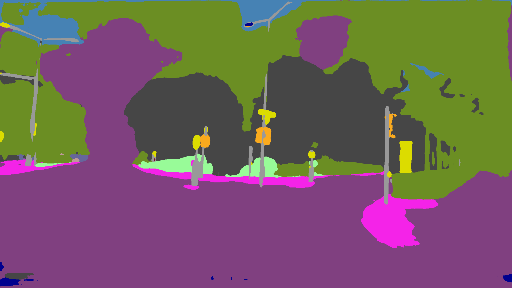} \\
			\rotatebox{90}{~~~~CMAda~\cite{dai2020curriculum}} &
			\includegraphics[width=0.235\textwidth]{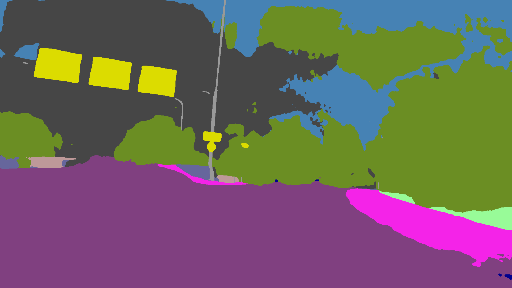} &
			\includegraphics[width=0.235\textwidth]{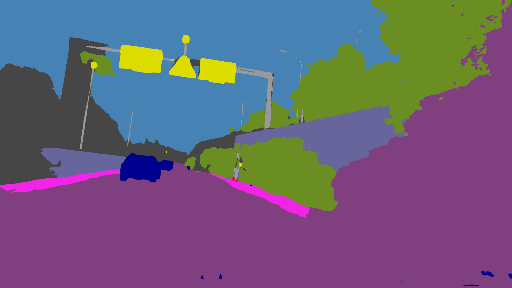} &
			\includegraphics[width=0.235\textwidth]{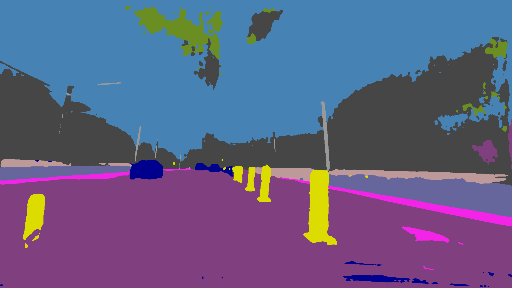} &
			\includegraphics[width=0.235\textwidth]{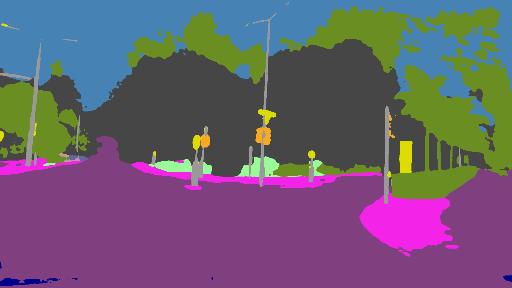} \\
			\rotatebox{90}{~~~~~CRST~\cite{zou2019confidence}} &
			\includegraphics[width=0.235\textwidth]{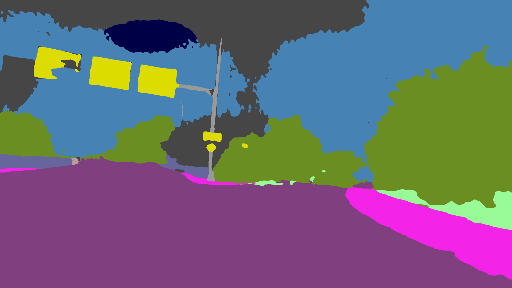} &
			\includegraphics[width=0.235\textwidth]{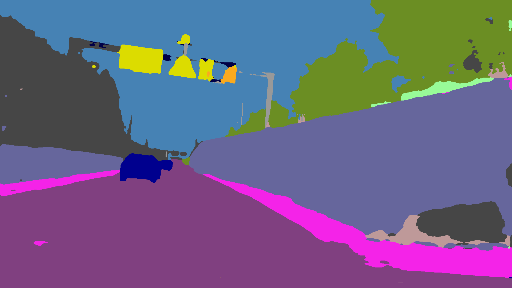} &
			\includegraphics[width=0.235\textwidth]{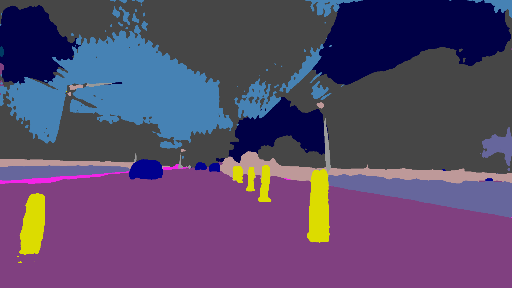} &
			\includegraphics[width=0.235\textwidth]{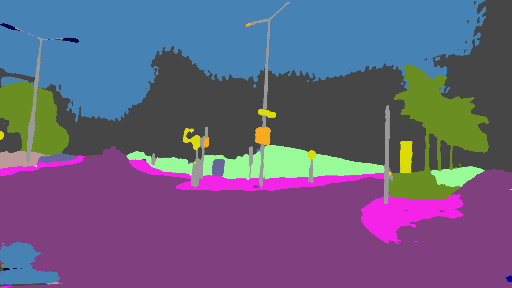} \\
			\rotatebox{90}{ {~CuDA-Net~\cite{xianzheng}} } &
			\includegraphics[width=0.235\textwidth]{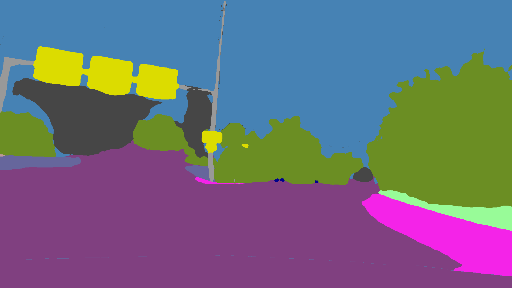} &
			\includegraphics[width=0.235\textwidth]{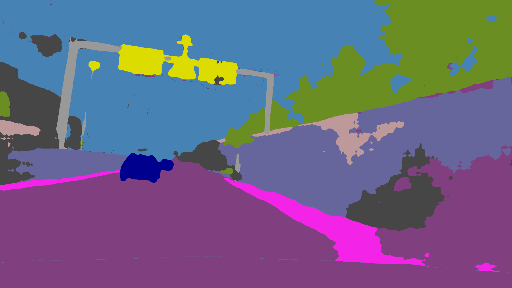} &
			\includegraphics[width=0.235\textwidth]{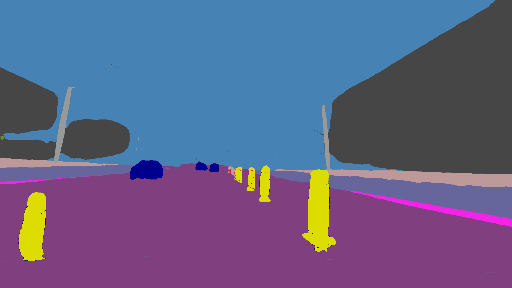} &
			\includegraphics[width=0.235\textwidth]{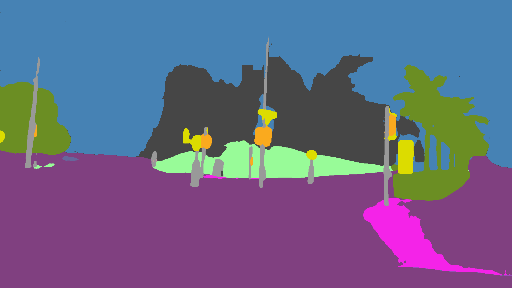} \\
			\rotatebox{90}{~~~~~~\textbf{TDo-Dif}} &
			\includegraphics[width=0.235\textwidth]{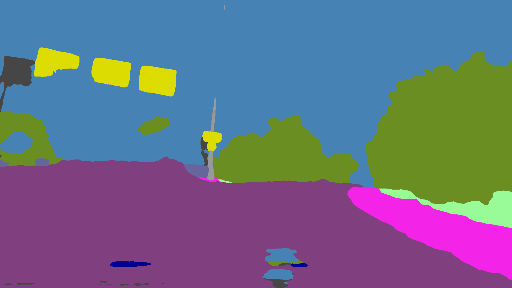} &
			\includegraphics[width=0.235\textwidth]{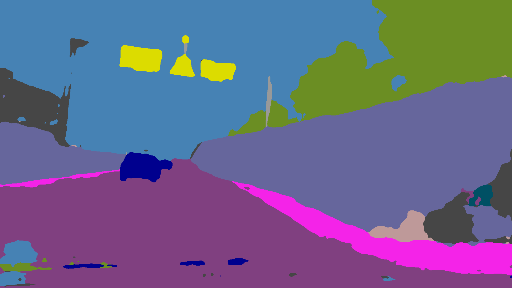} &
			\includegraphics[width=0.235\textwidth]{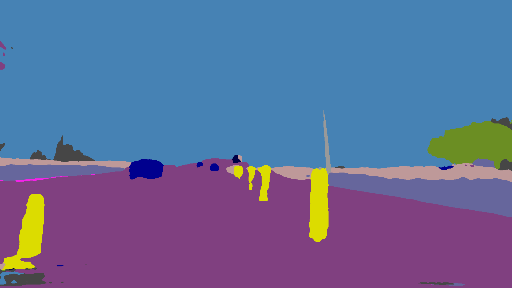} &
			\includegraphics[width=0.235\textwidth]{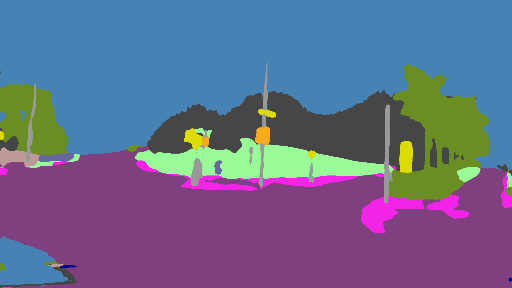} \\
			\rotatebox{90}{~~~~Ground-truth} &
			\includegraphics[width=0.235\textwidth]{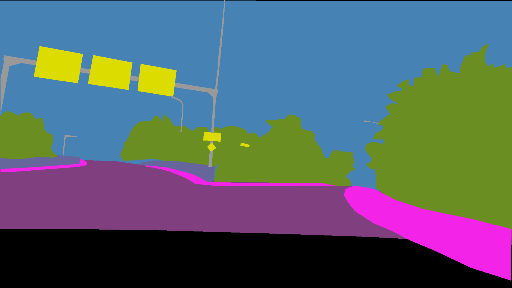} &
			\includegraphics[width=0.235\textwidth]{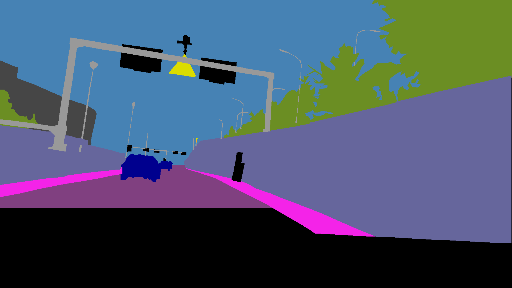} &
			\includegraphics[width=0.235\textwidth]{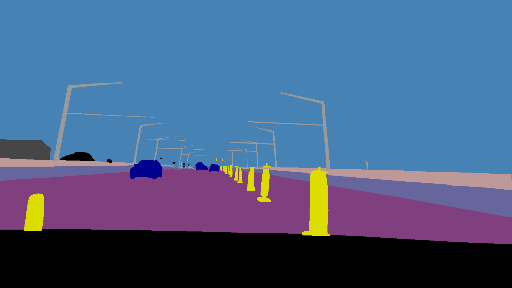} &
			\includegraphics[width=0.235\textwidth]{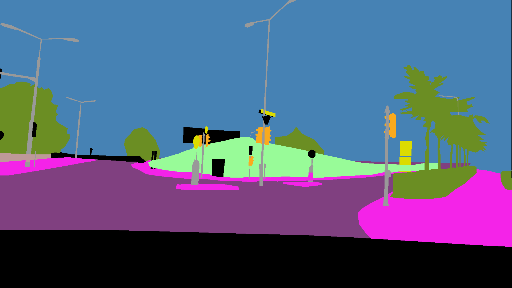} \\
			~&\multicolumn{4}{c}{\includegraphics[width=0.95\textwidth]{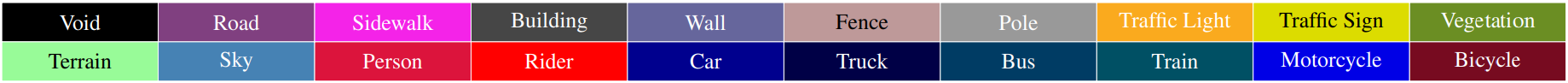} }\\
	\end{tabular}}
	
   \caption{Subjective quality comparison of test results on image samples from \textbf{Foggy Zurich}~\cite{dai2020curriculum}}
\label{cityscape2zurich}
\end{figure*}

\tabcolsep=0.5pt
\begin{figure*}[!htb]
	\centering
	\linespread{0.5}

\small{
		\begin{tabular}{ccccc}
			\rotatebox{90}{~~~~~~~Foggy image} &
			\includegraphics[width=0.235\textwidth]{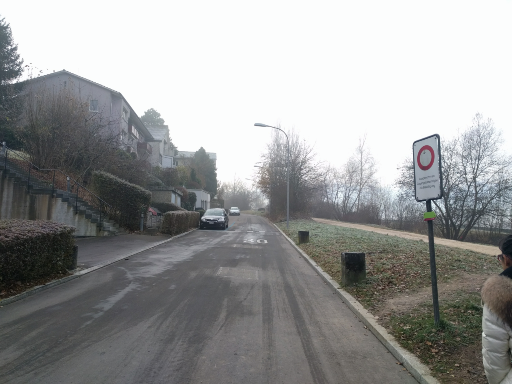} &
			\includegraphics[width=0.235\textwidth]{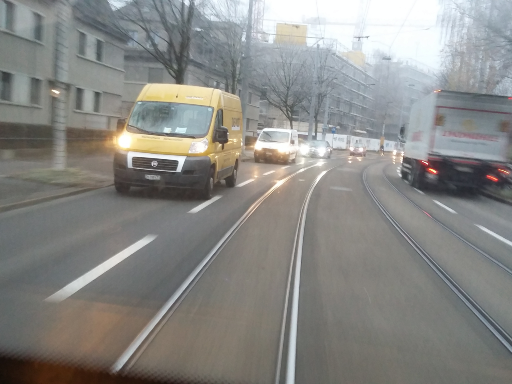} &
			\includegraphics[width=0.235\textwidth]{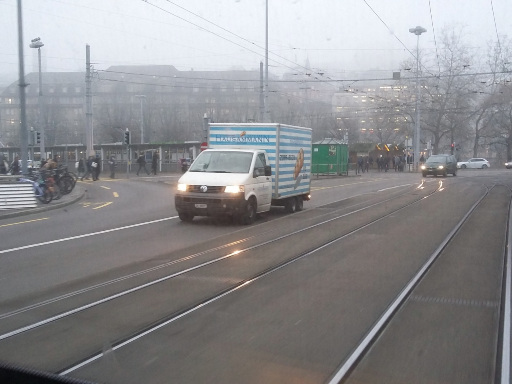} &
			\includegraphics[width=0.235\textwidth]{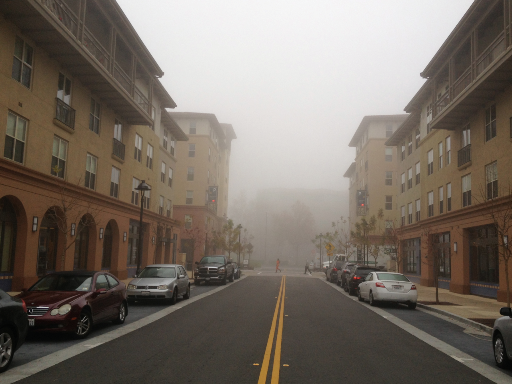} \\
			\rotatebox{90}{~~~~~~~~~SFSU~\cite{sakaridis2018semantic}} &
			\includegraphics[width=0.235\textwidth]{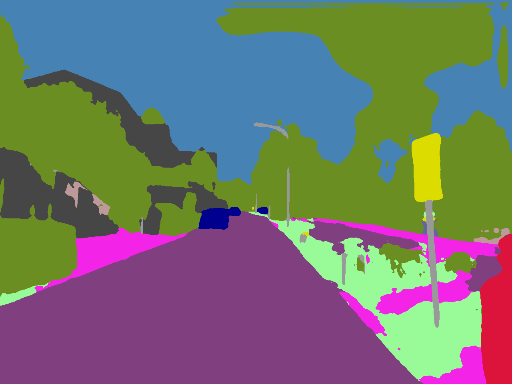} & 
			\includegraphics[width=0.235\textwidth]{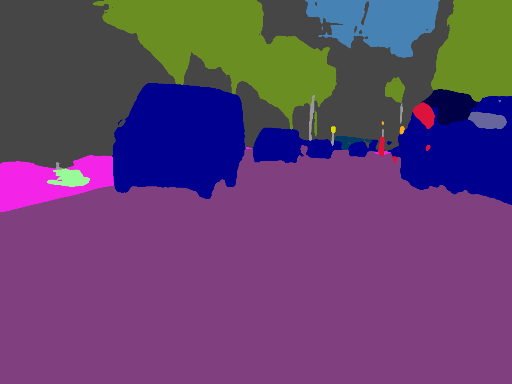} & 
			\includegraphics[width=0.235\textwidth]{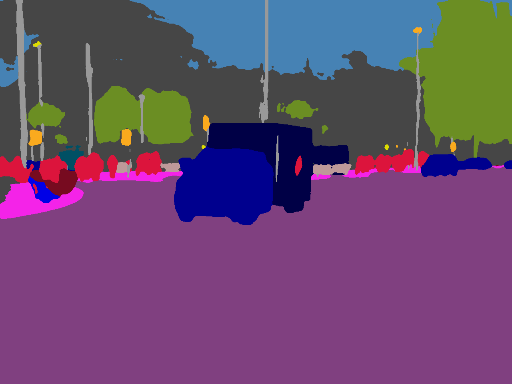} & 
			\includegraphics[width=0.235\textwidth]{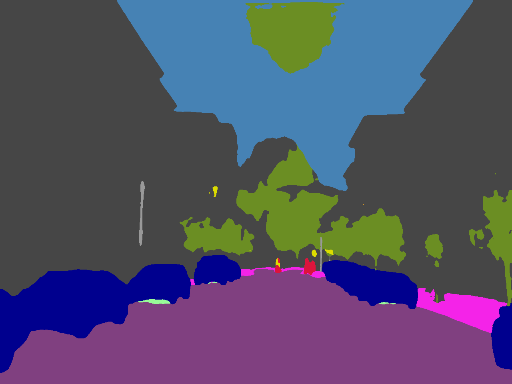} \\
			\rotatebox{90}{~~~~~~~CMAda~\cite{dai2020curriculum}} &
			\includegraphics[width=0.235\textwidth]{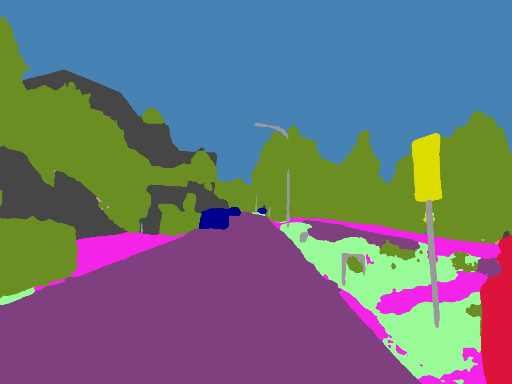} &
			\includegraphics[width=0.235\textwidth]{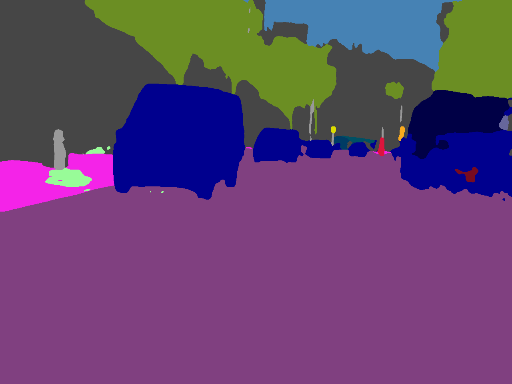} &
			\includegraphics[width=0.235\textwidth]{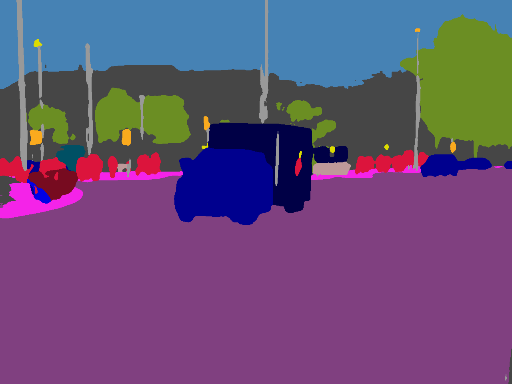} &
			\includegraphics[width=0.235\textwidth]{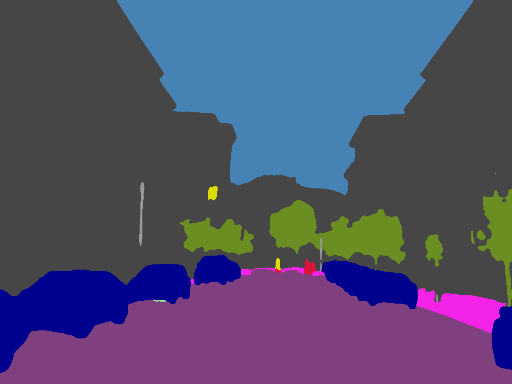} \\
			\rotatebox{90}{~~~~~~~~CRST~\cite{zou2019confidence}} &
			\includegraphics[width=0.235\textwidth]{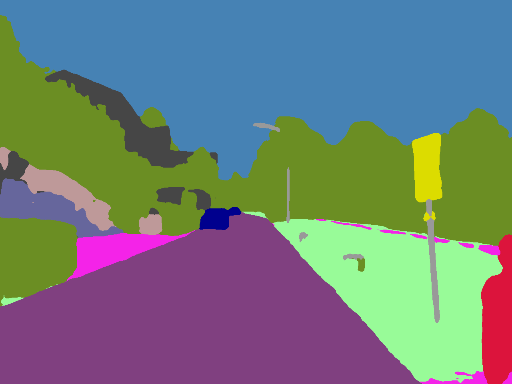} &
			\includegraphics[width=0.235\textwidth]{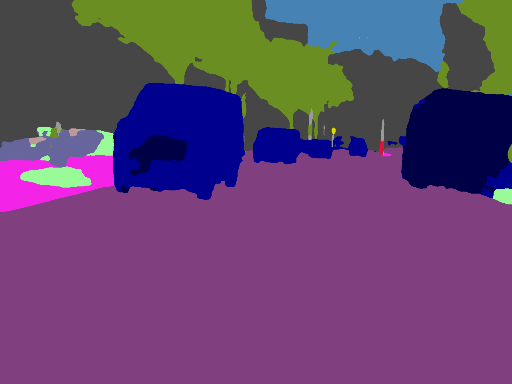} &
			\includegraphics[width=0.235\textwidth]{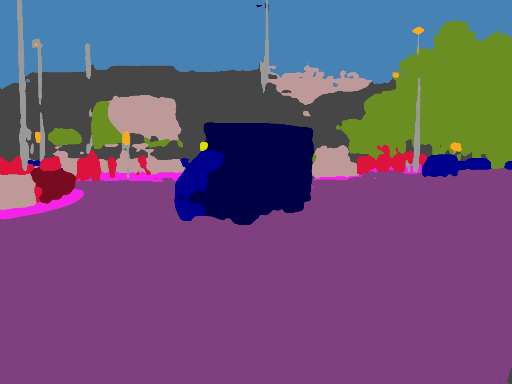} &
			\includegraphics[width=0.235\textwidth]{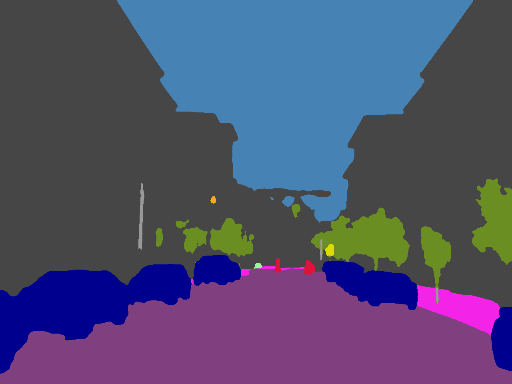} \\
			\rotatebox{90}{ {~~~~CuDA-Net~\cite{xianzheng}} } &
			\includegraphics[width=0.235\textwidth]{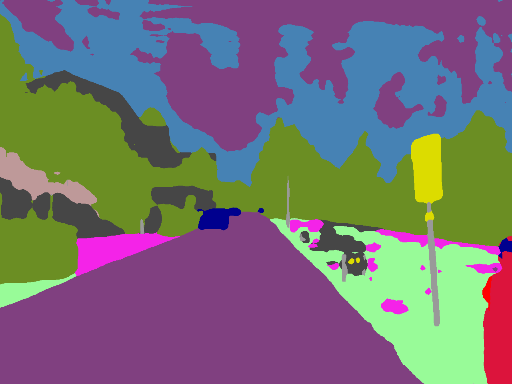} &
			\includegraphics[width=0.235\textwidth]{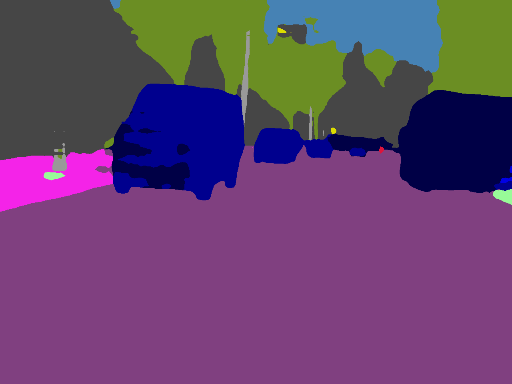} &
			\includegraphics[width=0.235\textwidth]{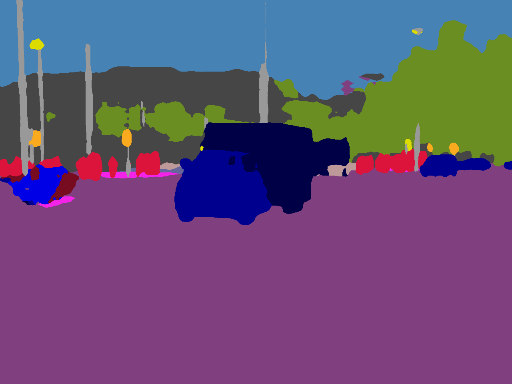} &
			\includegraphics[width=0.235\textwidth]{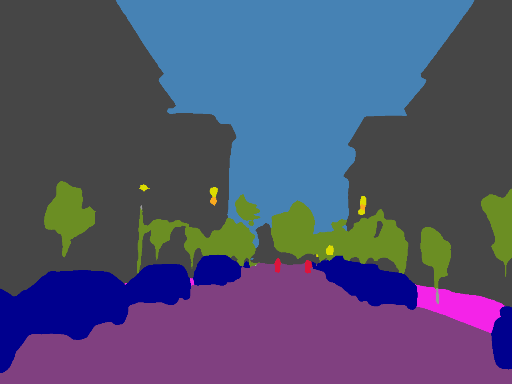} \\
			\rotatebox{90}{~~~~~~~~~~\textbf{TDo-Dif}} &
			\includegraphics[width=0.235\textwidth]{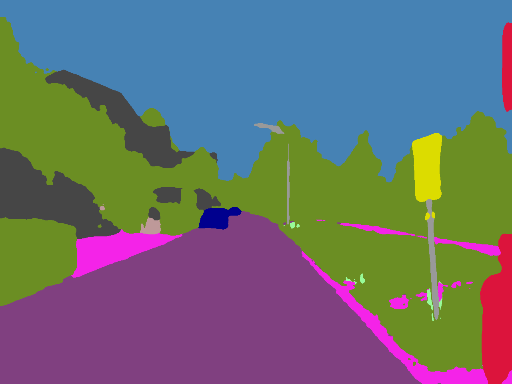} &
			\includegraphics[width=0.235\textwidth]{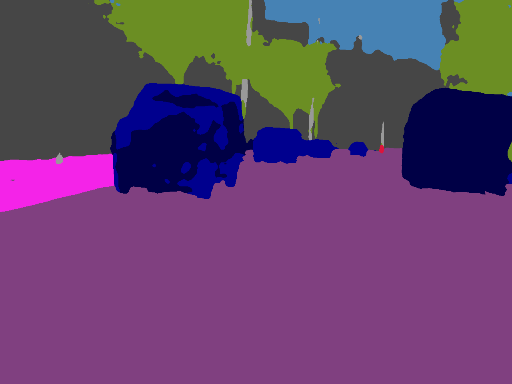} &
			\includegraphics[width=0.235\textwidth]{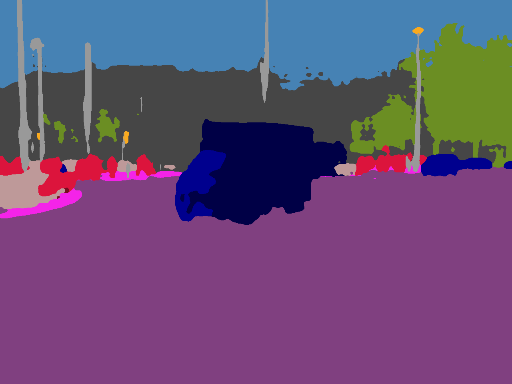} &
			\includegraphics[width=0.235\textwidth]{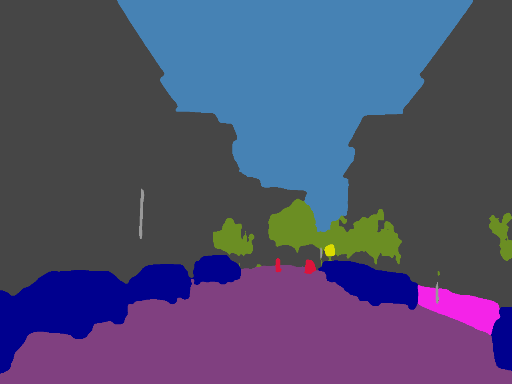} \\
			\rotatebox{90}{~~~~~~~Ground-truth} &
			\includegraphics[width=0.235\textwidth]{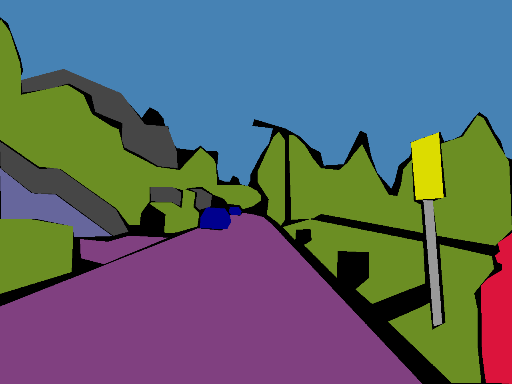} &
			\includegraphics[width=0.235\textwidth]{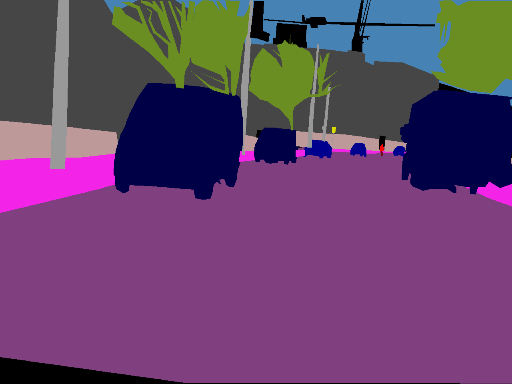} &
			\includegraphics[width=0.235\textwidth]{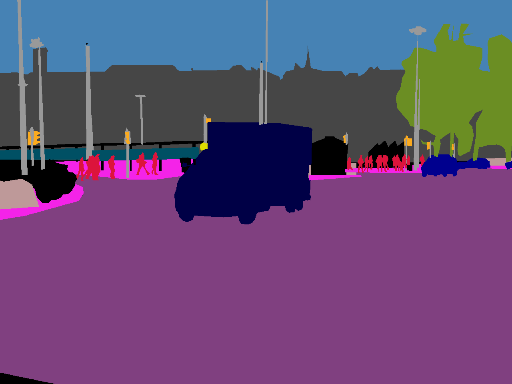} &
			\includegraphics[width=0.235\textwidth]{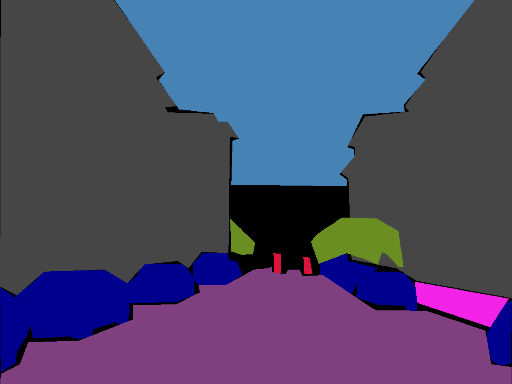} \\
			~&\multicolumn{4}{c}{\includegraphics[width=0.95\textwidth]{IEEEtran/SOTA_Zurich/label.png} }\\
	\end{tabular}}
   \caption{Subjective quality comparison of test results on image samples from \textbf{Foggy Driving}~\cite{sakaridis2018semantic}. }
\label{cityscape2driving}
\end{figure*}

\subsection{Comparisons with State-of-the-Art Methods}


In this section, we compare the proposed method with the baselines on the semantic segmentation of foggy scene, from both quantitative and qualitative aspects.

\subsubsection{Quantitative Comparisons}
We first focus on the testing of the proposed four components based on two core motivations: superpixel-based spatial label diffusion (SD) and local spatial similarity loss (SL) for \textit{spatial similarity}, and flow-based temporal label diffusion (TD) and temporal contrastive loss (TL) for \textit{temporal correspondence}. In general, we design two groups of model variants to sort out the most optimal combination of the proposed diffusion schemes, and the details of the variants are listed in Table~\ref{variants}. The comparison results with the baselines on \textbf{Foggy Zurich} and \textbf{Foggy Driving} are shown in Table~\ref{tab:cityscape2z} and Table~\ref{tab:cityscape2d}, respectively.

\textbf{Comparison with SOTA}. In general,  {the best variant of TDo-Dif with label diffusion outperform all baselines on \textbf{Foggy Zurich} and \textbf{Foggy Driving}, owning to the introduction of the target-domain knowledge for domain adaptation.}
On \textbf{Foggy Zurich}, the best variant of TDo-Dif reaches 51.92\% mIoU, surpassing both the \textbf{SFSU} \cite{sakaridis2018semantic} and the curriculum learning model \textbf{CMAda}\cite{dai2020curriculum}, which are specifically designed for foggy scene understanding, with significant gains of 18.51\% and 7.65\%, respectively. It indicates that while the curriculum learning strategy is effective in adapting the model to foggy data, the large number of false pseudo labels in the whole segmentation predictions leads to a degradation in performance. 
Although \textbf{CBST}~\cite{ZouYKW18} and \textbf{CRST}~\cite{zou2019confidence} adopt a small portion of pseudo labels for model re-training to avoid false labels, our method obtains additional gains of 6.13\% and 5.45\% mIoU, respectively, with a significant positive impact on some large objects (\emph{e.g.}, \textit{Building}, \textit{Fence} and \textit{Sky}) and small objects (\emph{e.g.}, \textit{Rider}, \textit{Motor} and \textit{Bike}).  {In contrast to methods that attempt to bridge the gap between the source and target domains, \textit{i.e.,} CuDA-Net, FIFO, CMDIT, and FogAdapt+, our proposed TDo-Dif explicitly exploits the self-similarity present in the target data to increase the correct labels for these classes and therefore achieves the best performance on the foggy scene segmentation}.

\tabcolsep=0.5pt
\begin{figure*}[htb]
	\centering
\small{
		\begin{tabular}{cc}
			\includegraphics[width=0.49\textwidth]{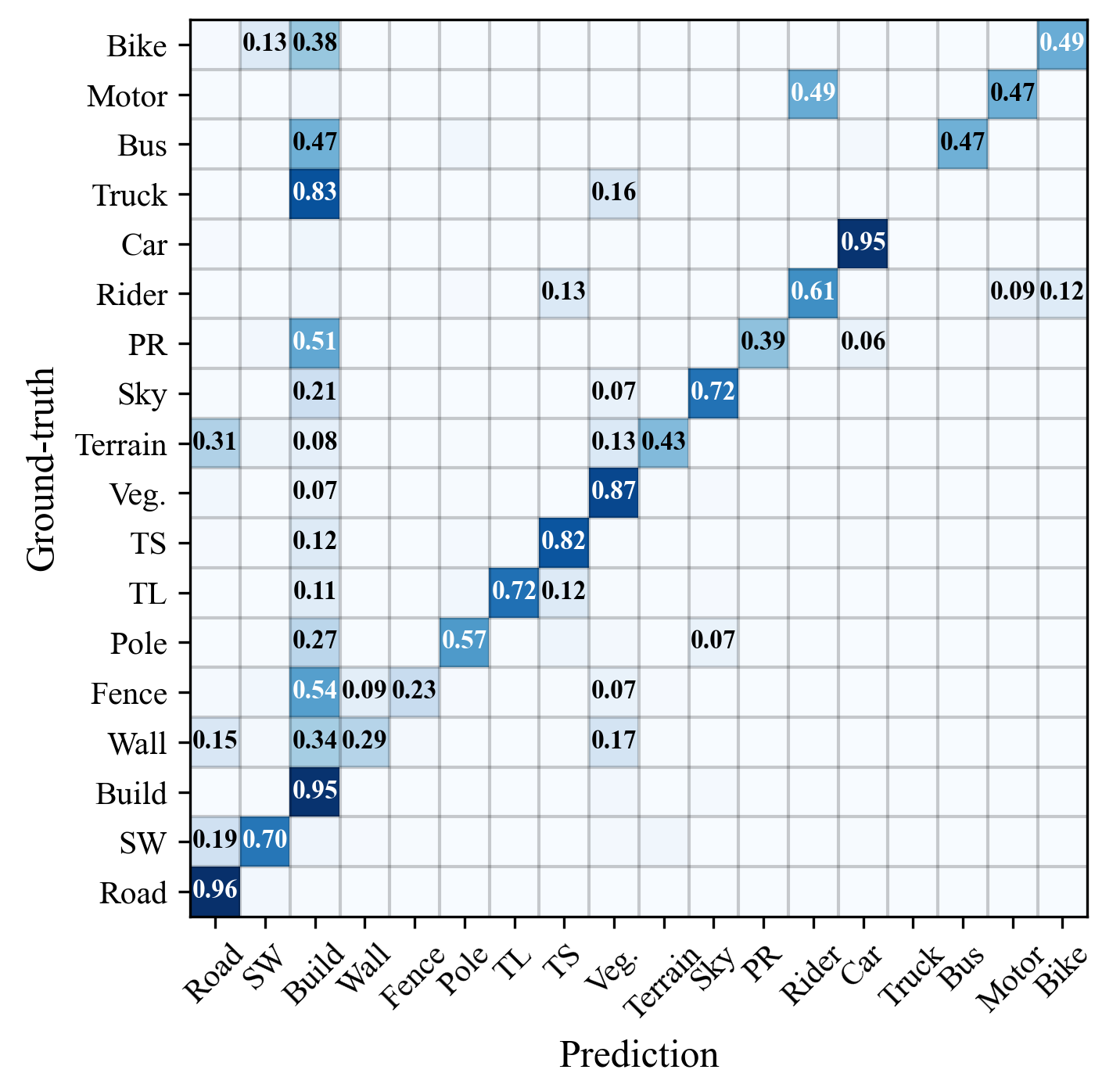} &
			\includegraphics[width=0.49\textwidth]{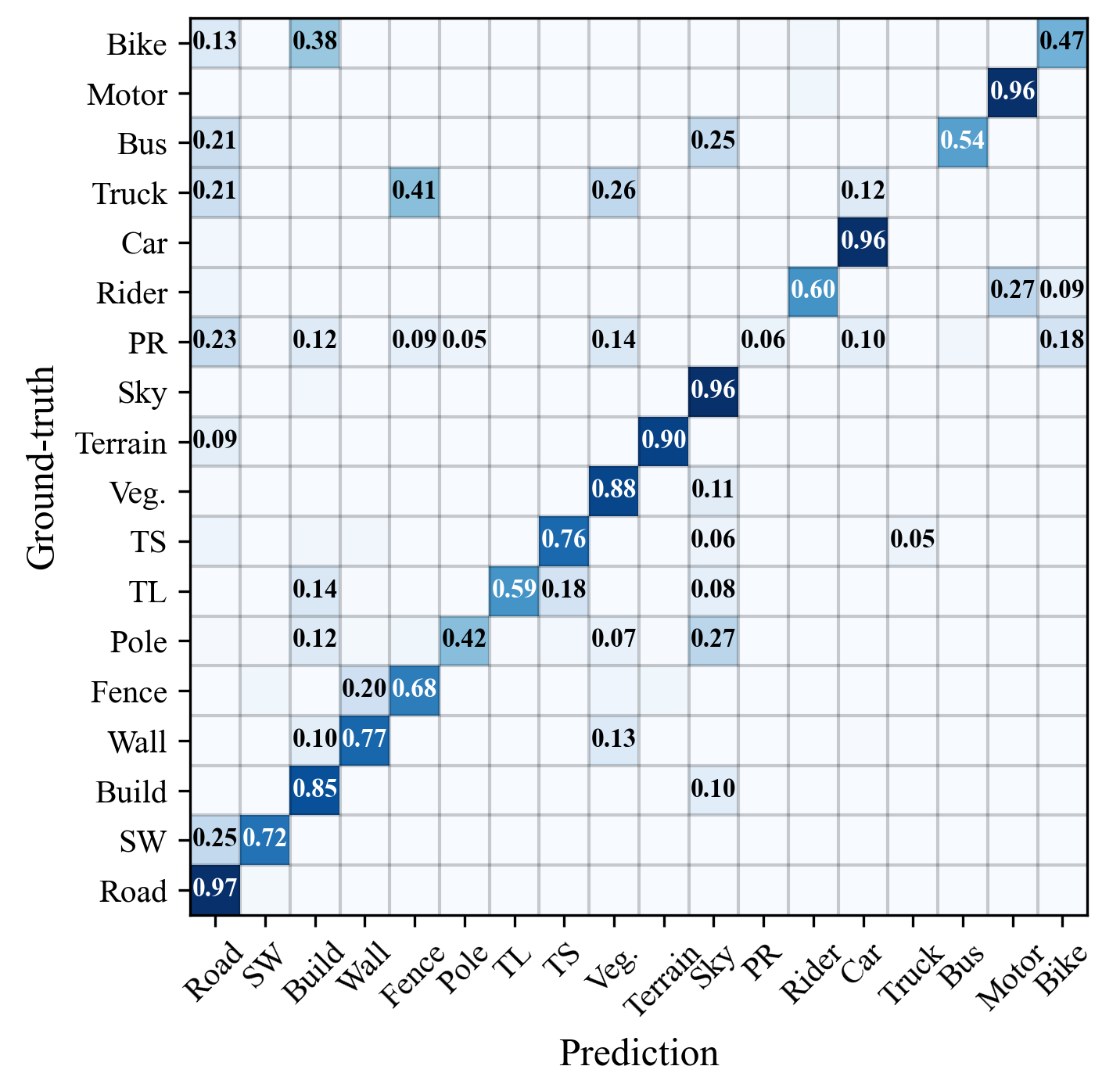} \\
	\end{tabular}}
   \caption{Confusion matrices for \textbf{CMAda} (left) and Ours (\textbf{TDo-Dif}) (right) for the semantic segmentation on \textbf{Foggy Zurich}.}
\label{fig:confusion}
\end{figure*}

\begin{figure}[tb]
     \centering
     \includegraphics[width=1.0\columnwidth]{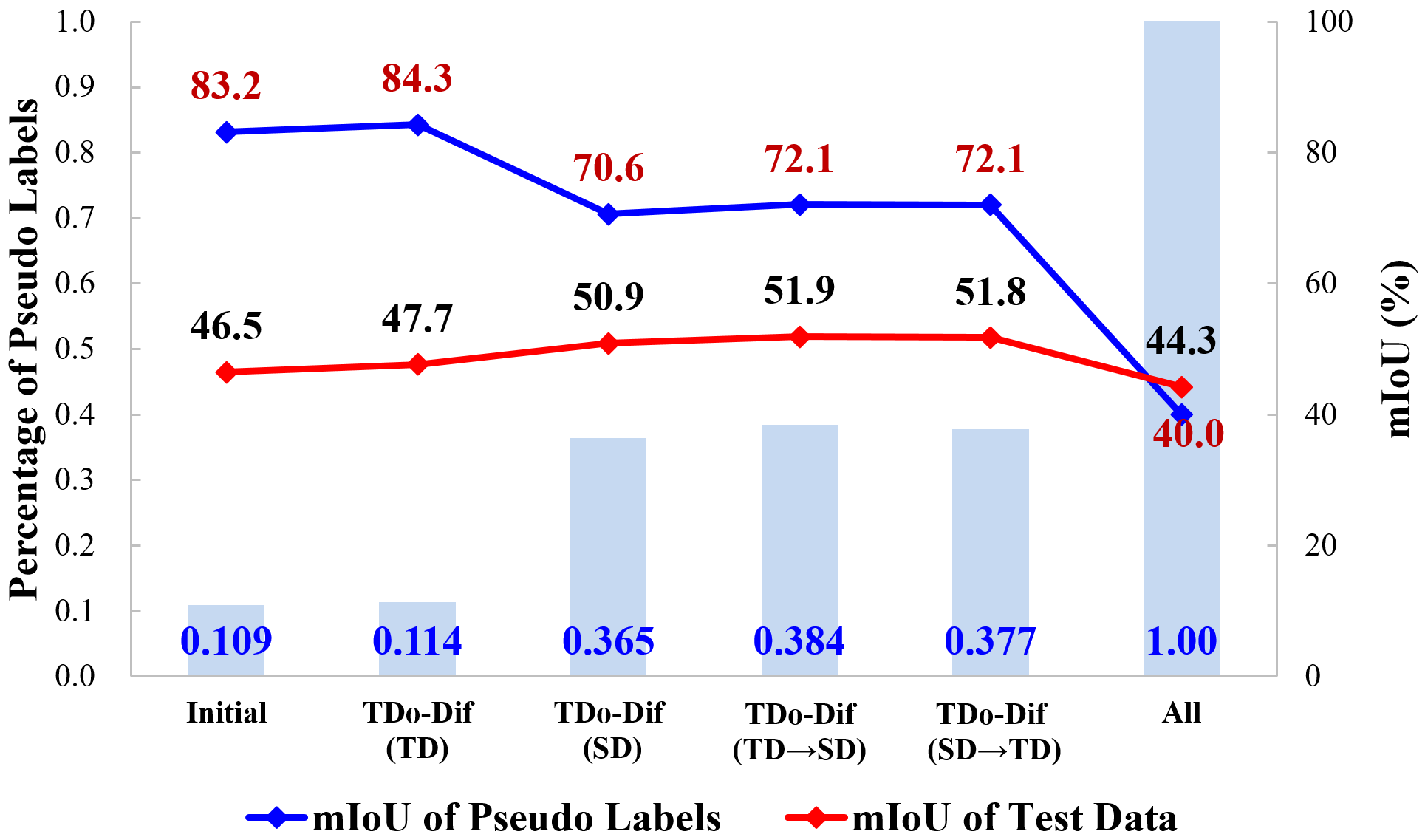}
     \caption{Statistical comparison of the effect of diffused pseudo labels quality on segmentation performance. The bar and blue line show the average percentage of pixels and the average accuracy of pseudo labels at each diffused stage. The red line shows the segmentation performance of the model trained with the corresponding diffused pseudo labels.}
     \label{portion}
\end{figure}

\textbf{Comparison among variants of TDo-Dif.} By comparing the performance among variants, we evaluate the effectiveness of each component on semantic segmentation performance. The variants in the first group show that all components have a positive impact on the understanding of foggy scenes. Comparing TDo-Dif (SD+SL) with TDo-Dif (TD+TL), spatial similarity shows a higher performance improvement over temporal correspondence, with a gain of 2.87\% mIoU. This can be attributed to the fact that the spatial diffused area is much larger than the temporal one. The local spatial similarity loss and temporal contrastive loss also contribute additional 0.25\% and 0.88\% mIoU, repectively, through comparing TDo-Dif (SD+SL) with TDo-Dif (SD) and TDo-Dif (TD+TL) with TDo-Dif (TD). We also investigate the impact of different spatial and temporal diffusion orders through the variants of the second group. The two strategies are comparable in terms of segmentation performance and selection of pseudo labels as well as mIoU of pseudo labels.

\textbf{Model Generalization.} Since the adaptive models with both spatial and temporal diffusion achieve the best performance on \textbf{Foggy Zurich}, we test them on the \textbf{Foggy Driving} dataset. As shown in Table \ref{tab:cityscape2d}, our model achieves 50.65\% mIoU, which exceeds most of state-of-the-art methods and demonstrates the good generality of our model. 

Moreover, since TDo-Dif (TD$\rightarrow$SD+SL+TL) reaches the highest scores on both datasets, we use this model as our final model and denote it as \textbf{TDo-Dif} in the following experiments.

\subsubsection{Qualitative Comparisons}

Fig.~\ref{cityscape2zurich} and Fig.~\ref{cityscape2driving} give some examples of qualitative comparison on \textbf{Foggy Zurich} and \textbf{Foggy Driving}, respectively. Trained on a synthetic foggy dataset,  
\textbf{SFSU} \cite{SakaridisDHG18} has limited generalization ability on the real-world foggy scenarios, \emph{e.g.},  \textit{Sky} and \textit{Building} are incorrectly predicted as \textit{Road} and \textit{Vegetation}. The performance of \textbf{CMAda} \cite{dai2020curriculum} is relatively good by re-training on the target domain data, but its results also have many segmentation errors, such as categories of \textit{Sky}, \textit{Wall} and \textit{Sidewalk}. \textbf{CRST} \cite{zou2019confidence}, trained with sparse and confident pseudo labels, presents better results than \textbf{CMAda} in terms of semantic mis-recognition.  {\textbf{CuDA-Net} achieves the excellent performance in some small objects, such as the \textit{Pole} and \textit{Traffic Sign}, but the performance of the large objects still needs to be improved, especially \textit{Building} and \textit{Road}}. Benefiting from the spatial and temporal pseudo label diffusion, the proposed method achieves the best results, especially for large objects, \emph{e.g.}, \textit{Sky} and \textit{Vegetation}. According to the underlying characteristics of fog, \textit{Sky} is most affected by fog at the farthest distance from the camera, which explains the poor performance of all baselines on this class, but our method performs well on it owning to the label diffusion.
  
\subsubsection{Confusion between Classes}
In this part, we provide more informative analysis by comparing the confusion matrices of the baseline and ours in Fig.~\ref{fig:confusion}. Considering the space limitation, we only show the comparison between \textbf{CMAda} \cite{dai2020curriculum} (the one with best performance among all baselines) and ours. An obvious observation from the confusion matrices of \textbf{CMAda} is that most categories tend to be misclassified into \textit{Building}, especially the classes of \textit{Truck}, \textit{Fence}, \textit{Person}, and \textit{Bus}, which is consistent with the visualization in Fig.~\ref{cityscape2zurich}. This may be due to the fact that noisy pseudo labels from the source model are used to re-train the model for the foggy scenes. This issue is greatly alleviated by our method, by which most of the classes are accurately classified with much smaller probability of belonging to other classes than to the same class. Some classes such as \textit{Bike}, \textit{Truck}, and \textit{Person} are difficult to be correctly classified, probably due to their small portion of pixel numbers in the overall dataset. 




\subsection{Ablation Study}

\tabcolsep=0.5pt
\begin{figure*}[htb]
	\centering
\small{
		\begin{tabular}{ccccc}
			\includegraphics[width=0.19\textwidth]{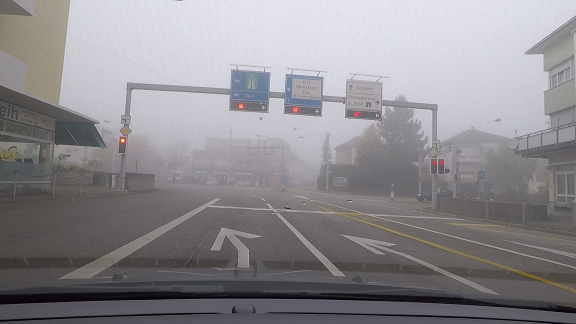} &
			\includegraphics[width=0.19\textwidth]{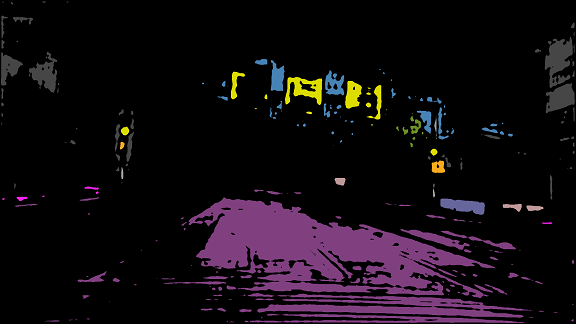} &
			\includegraphics[width=0.19\textwidth]{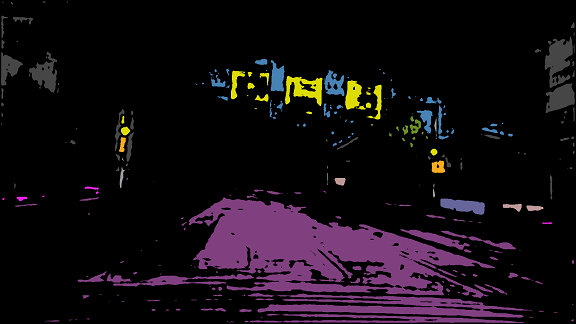} &
			\includegraphics[width=0.19\textwidth]{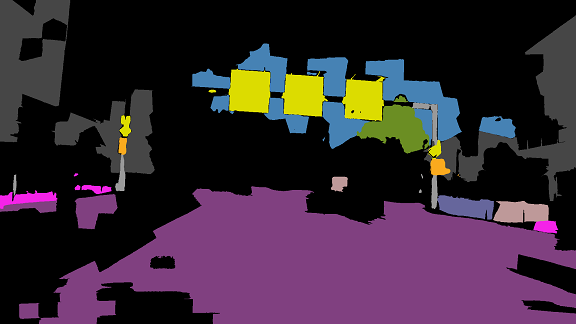} &
			\includegraphics[width=0.19\textwidth]{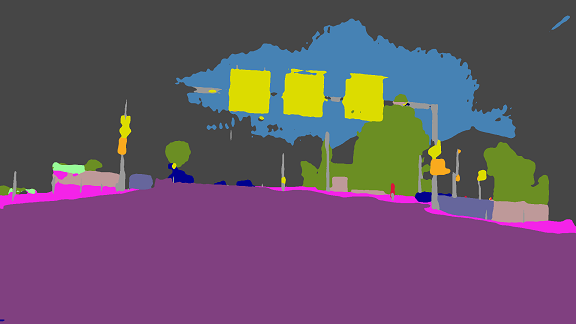} \\
			\includegraphics[width=0.19\textwidth]{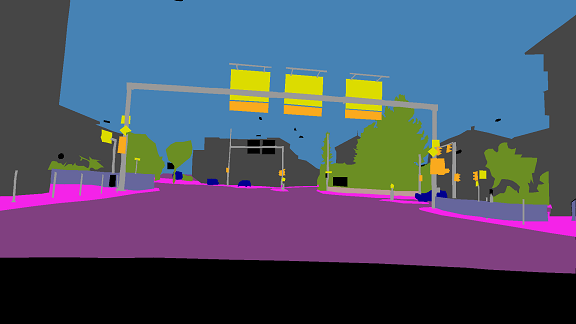} & 
			\includegraphics[width=0.19\textwidth]{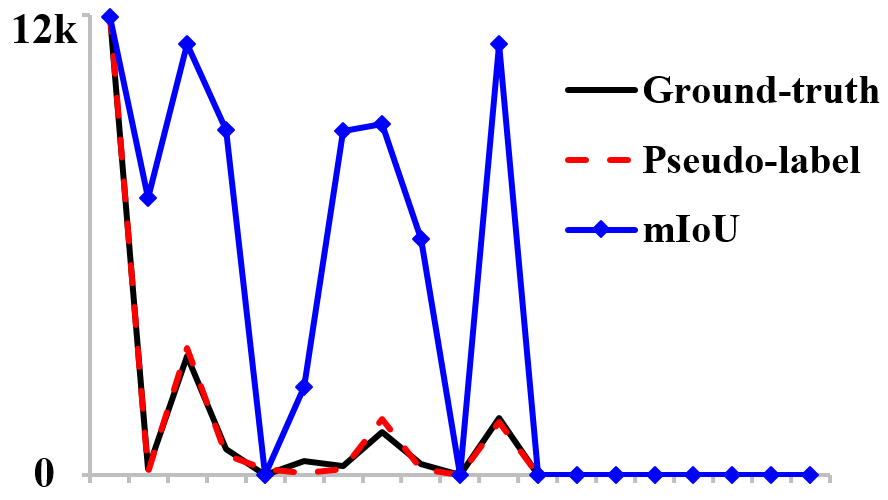} &
			\includegraphics[width=0.19\textwidth]{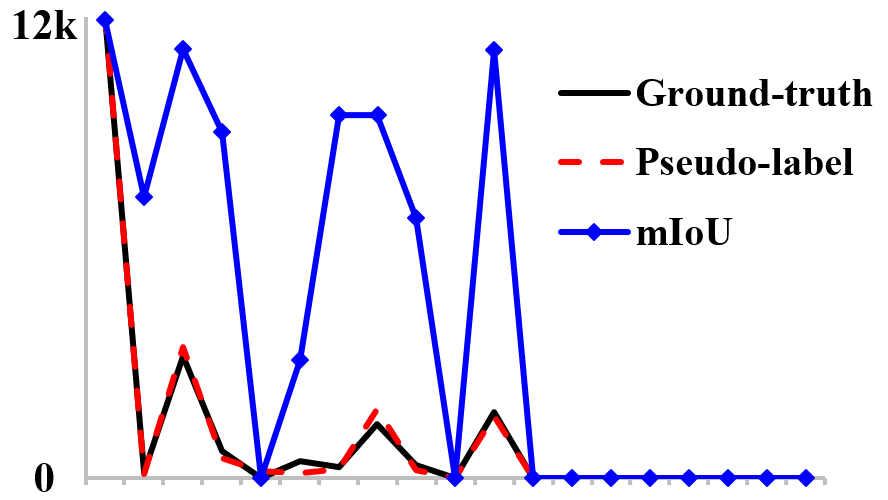} & 
			\includegraphics[width=0.19\textwidth]{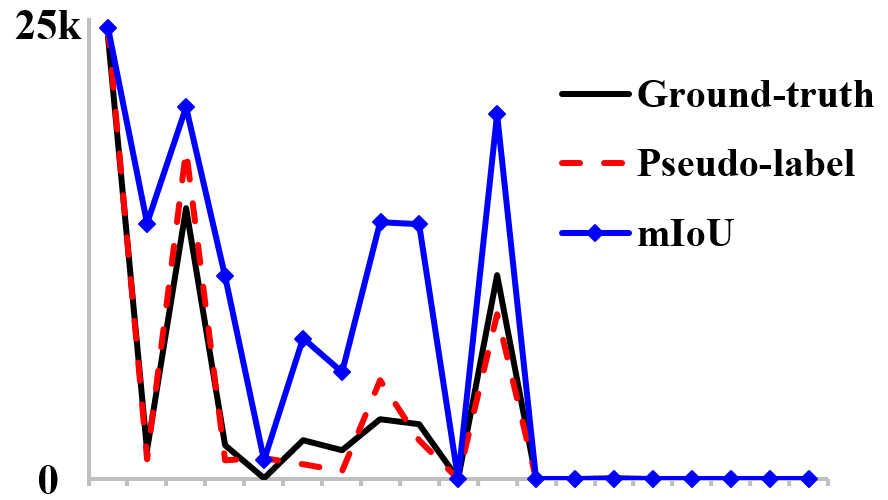} & 
			\includegraphics[width=0.19\textwidth]{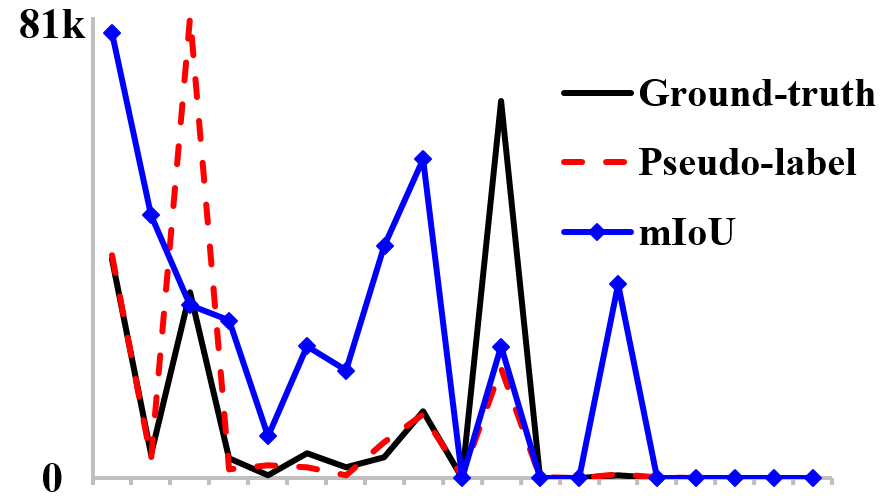} \\
			\includegraphics[width=0.19\textwidth]{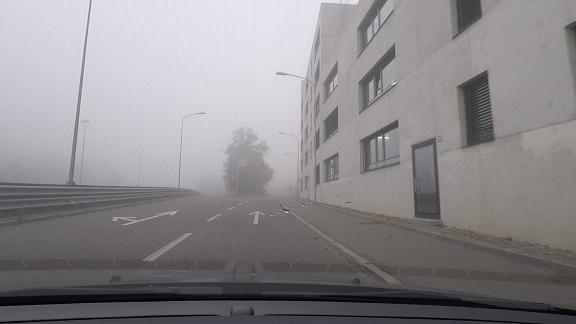} &
			\includegraphics[width=0.19\textwidth]{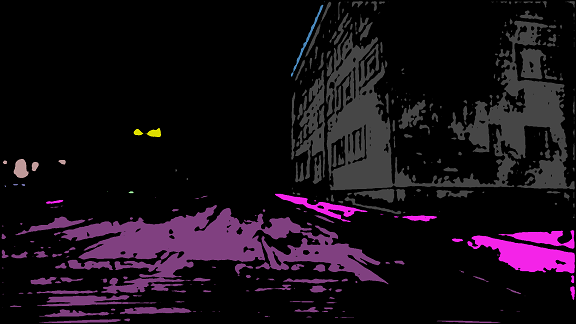} &
			\includegraphics[width=0.19\textwidth]{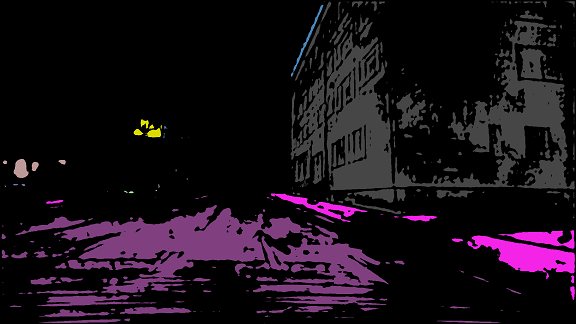} &
			\includegraphics[width=0.19\textwidth]{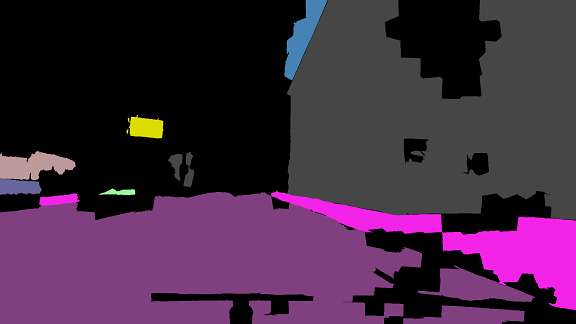} &
			\includegraphics[width=0.19\textwidth]{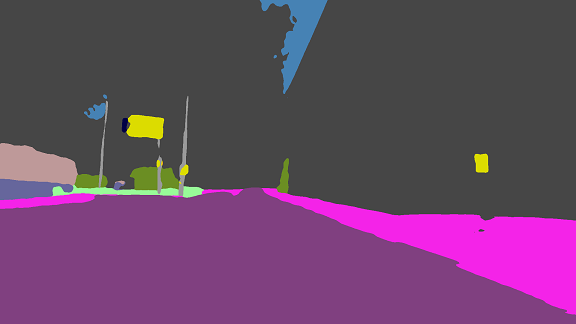} \\
			\includegraphics[width=0.19\textwidth]{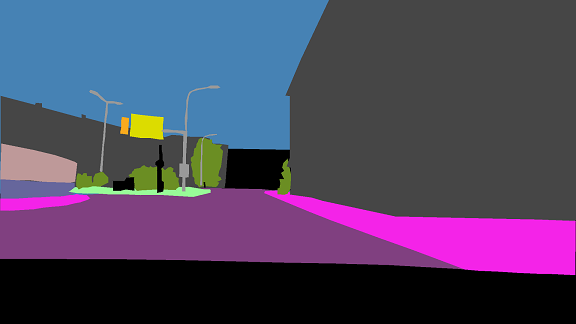} & 
			\includegraphics[width=0.19\textwidth]{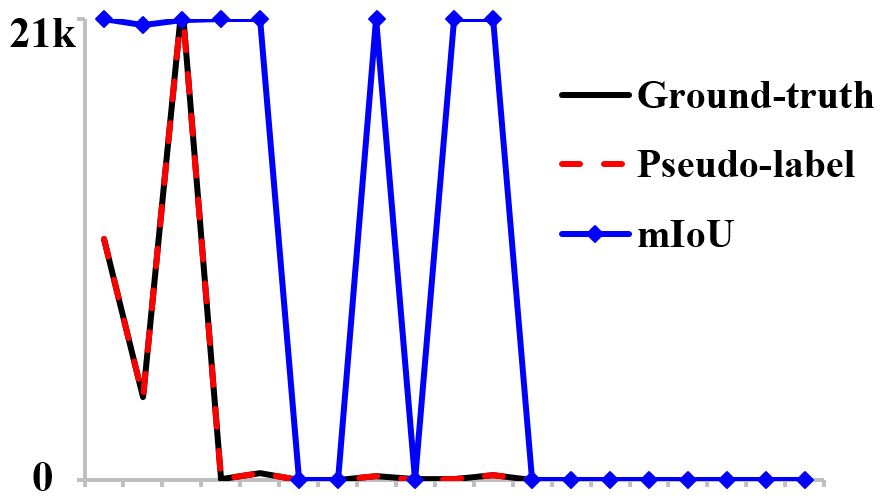} &
			\includegraphics[width=0.19\textwidth]{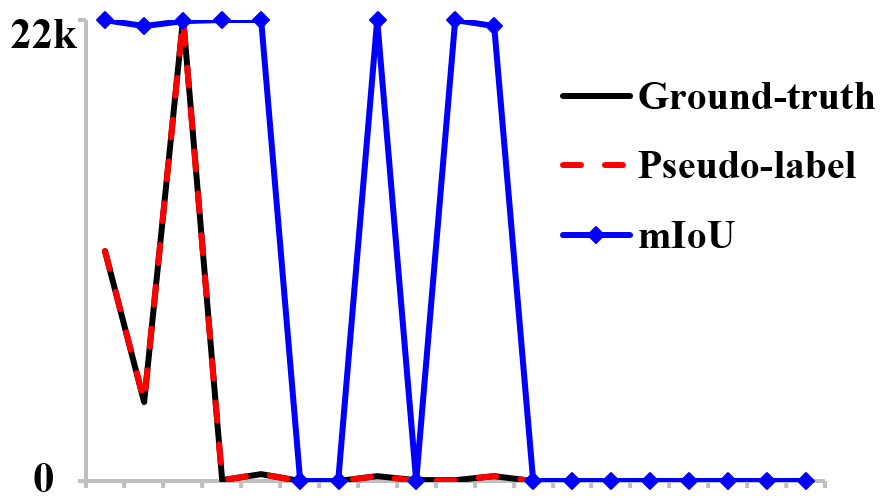} & 
			\includegraphics[width=0.19\textwidth]{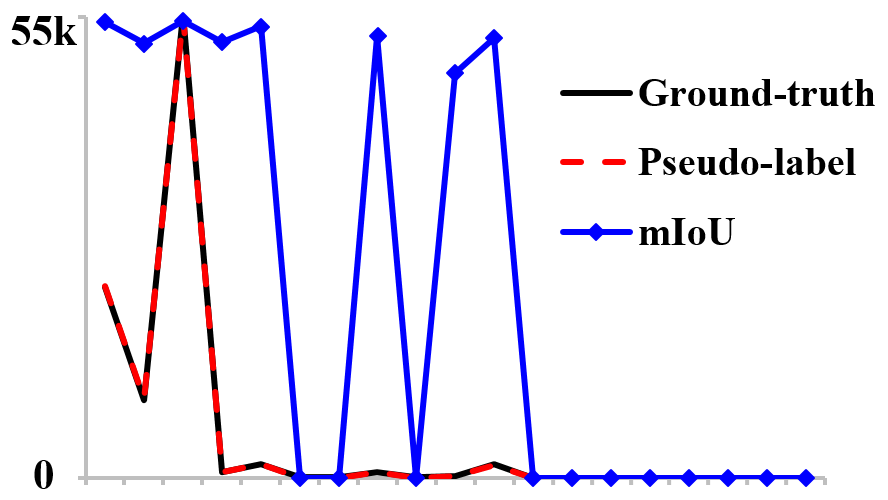} & 
			\includegraphics[width=0.19\textwidth]{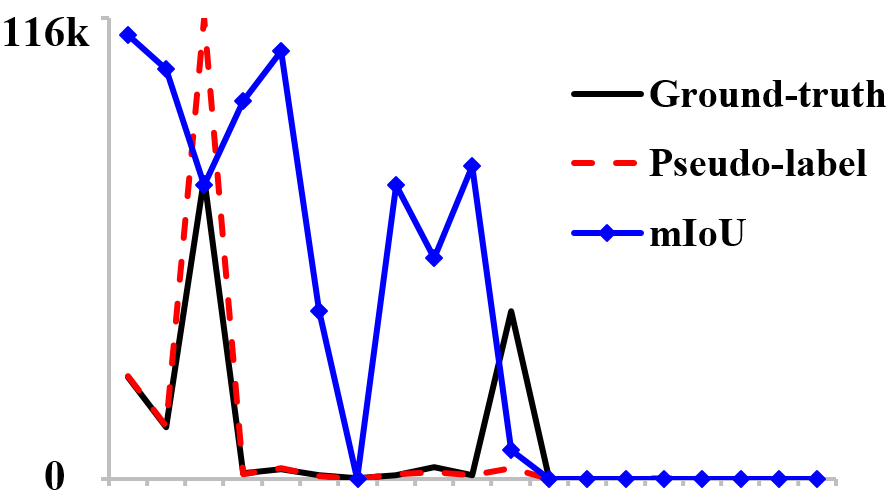} \\
			(a) Foggy image & (b) Initial labels & (c) TDo-Dif (TD)& (d) TDo-Dif (SD)& (e) Whole labels\\
	\end{tabular}}
   \caption{Comparison of subjective results for various pseudo labels on image samples from \textbf{Cityscapes}. The black solid line and red dashed line represent the pixel number distribution of every category, and the blue line represent the value of mIoU of every pixel.}
\label{fig:distribution}
\end{figure*}

\subsubsection{Quality of Diffused Label and Its Impacts on Segmentation}
The success of self-training-based domain adaptation relies heavily on the quality of pseudo labels (\textit{i.e.}, the percentage of valid pixels and the accuracy of pseudo labels). In this section, we assess the quality of diffused labels and how the quality of diffusion labels will affect the final performance of semantic segmentation. 

\textbf{Assessment on Overall Statistics}. The statistics for quality assessment of pseudo labels are calculated from all the 40 test images of \textbf{Foggy Zurich}. As shown in Fig.~\ref{portion}, we adopt three pseudo-label generation strategies: 1) initial sparse labels without diffusion, 2) diffused labels from four variants of our method, and 3) an entire predicted labels with noise. all variants in this evaluation are with their corresponding losses, but loss notations are omitted for short in the figure. 

The accuracy of the pseudo labels is measured by two terms: the average percentage of pseudo labels in the image (blue bar) and the mIoU of pseudo labels (blue line). In general, temporal diffusion (TD) results in a small increase in the number of labeled pixels (percentage increases by 0.5\%-1.2\%) and improvement in label accuracy (mIoU increases by 1.1\%-1.5\%), while spatial diffusion (SD) results in a large growth of the pseudo labels (percentage increases by 25\%-27\%) but a drop of label accuracy (mIoU decreases by 12\%-13\%). The original predicted labels covers the entire image, but the accuracy is nearly half of the selected pseudo labels, namely more than half of the labels are wrongly predicted. Note that the percentage of initial pseudo labels in the test set (\emph{i.e.}, 0.109), which captured in dense fog, is typically much lower than 0.2 since their confidence scores are much lower than that of light foggy images.

The impact of pseudo labels quality on the final semantic segmentation is measured by the mIoU of the test data (red line). The accuracy of the final segmentation is the result of the combined effect of the number and accuracy of pseudo labels. As shown in the figure, within the proposed label diffusion strategies, the final performance improves with the percentage increment of labels, but the performance gains are slowed down by the accuracy drop. The extremely noisy labels from the original prediction lead to a dramatic decrease in segmentation accuracy. This proves that although our label diffusion strategy may introduce some noise into the labels, the positive effects from the newly added correct labels plays a dominant role on the final performance.
 

\textbf{Assessment on Sample Images}. Two samples demonstrating the detailed assessment on the quality of pseudo label diffusion are shown in Fig.~\ref{fig:distribution}, whose statistical patterns are quite consistent with the overall statistics. The initial pseudo labels are quite accurate (almost overlapped number distribution of ground-truth and pseudo labels in the distribution diagram), but they are quite sparse. The temporal diffusion (TD) brings slight increment in the valid pixel labels and maintains a high accuracy, bringing additional reliable labels for further amplification by spatial diffusion. Compared to the initial labels, the spatial diffused (SD) labels are much denser. The number of pixels per class and their accuracy shown in the distribution diagram indicate that the valid labels are largely expanded, while the accuracy is still kept to some extent. The distribution of pixel counts in the pseudo labels is almost identical to that in the ground-truth labels. For the entire original prediction, the accuracy is too low to be used for re-training, although the number of valid labels is large.









\subsubsection{Superpixel Versus Deep Feature for Spatial Diffusion}
\begin{figure}[tb]
	\centering
\footnotesize{
		\begin{tabular}{cccc}
			\rotatebox{90}{Foggy image} &
			\includegraphics[width=0.31\columnwidth]{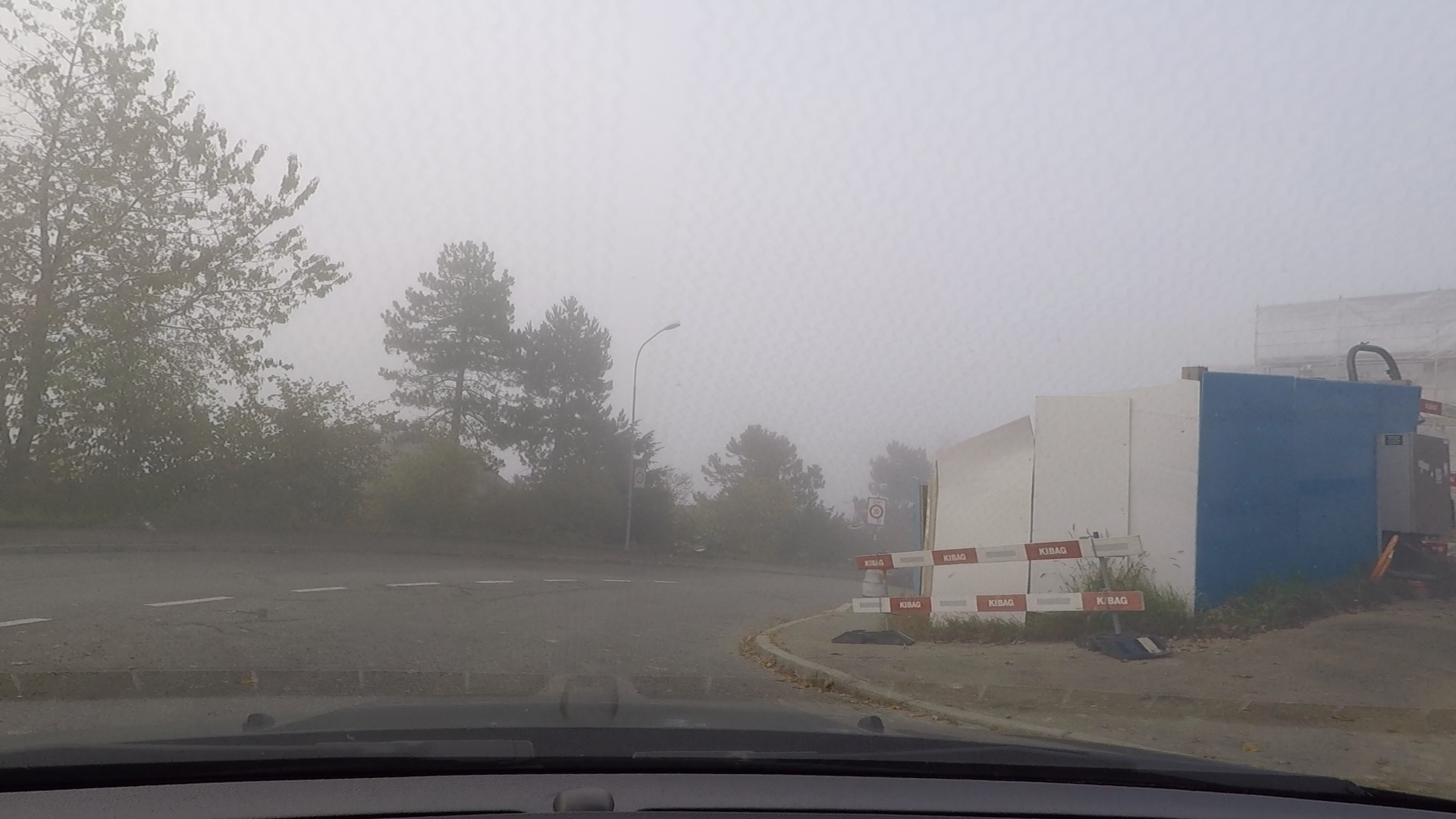} &
			\includegraphics[width=0.31\columnwidth]{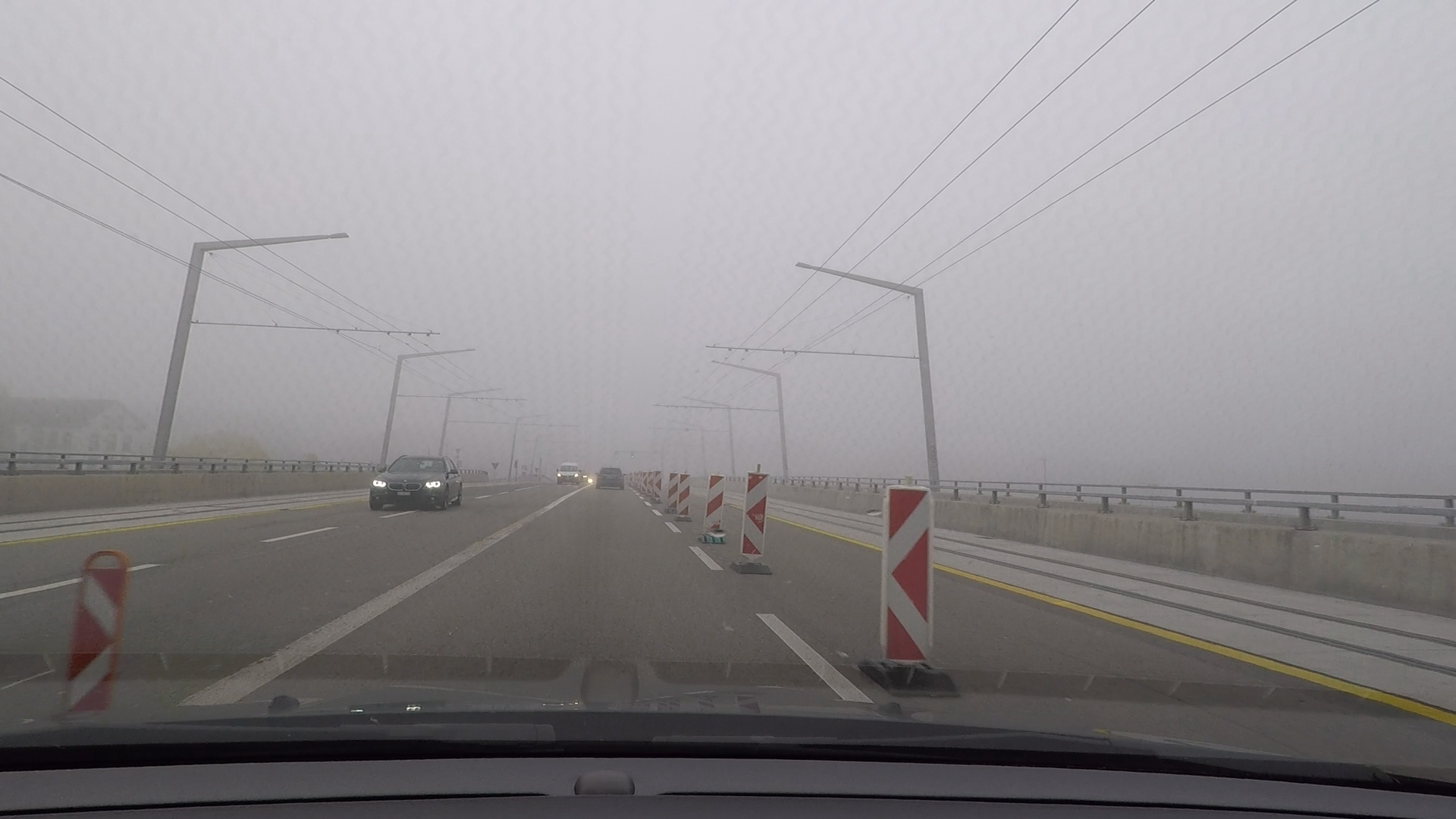} &
			\includegraphics[width=0.31\columnwidth]{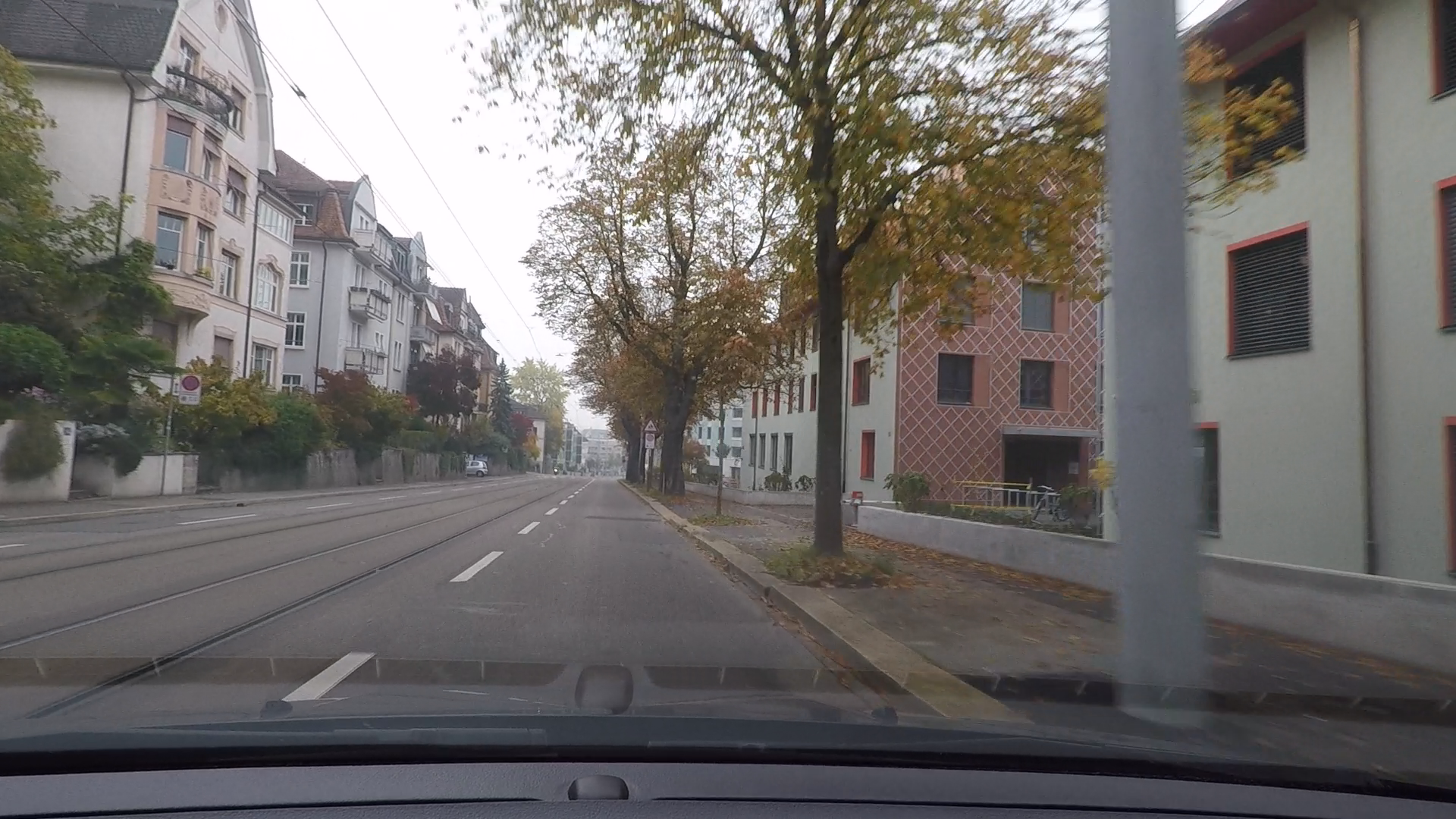} \\
				\rotatebox{90}{Prediction} &
			\includegraphics[width=0.31\columnwidth]{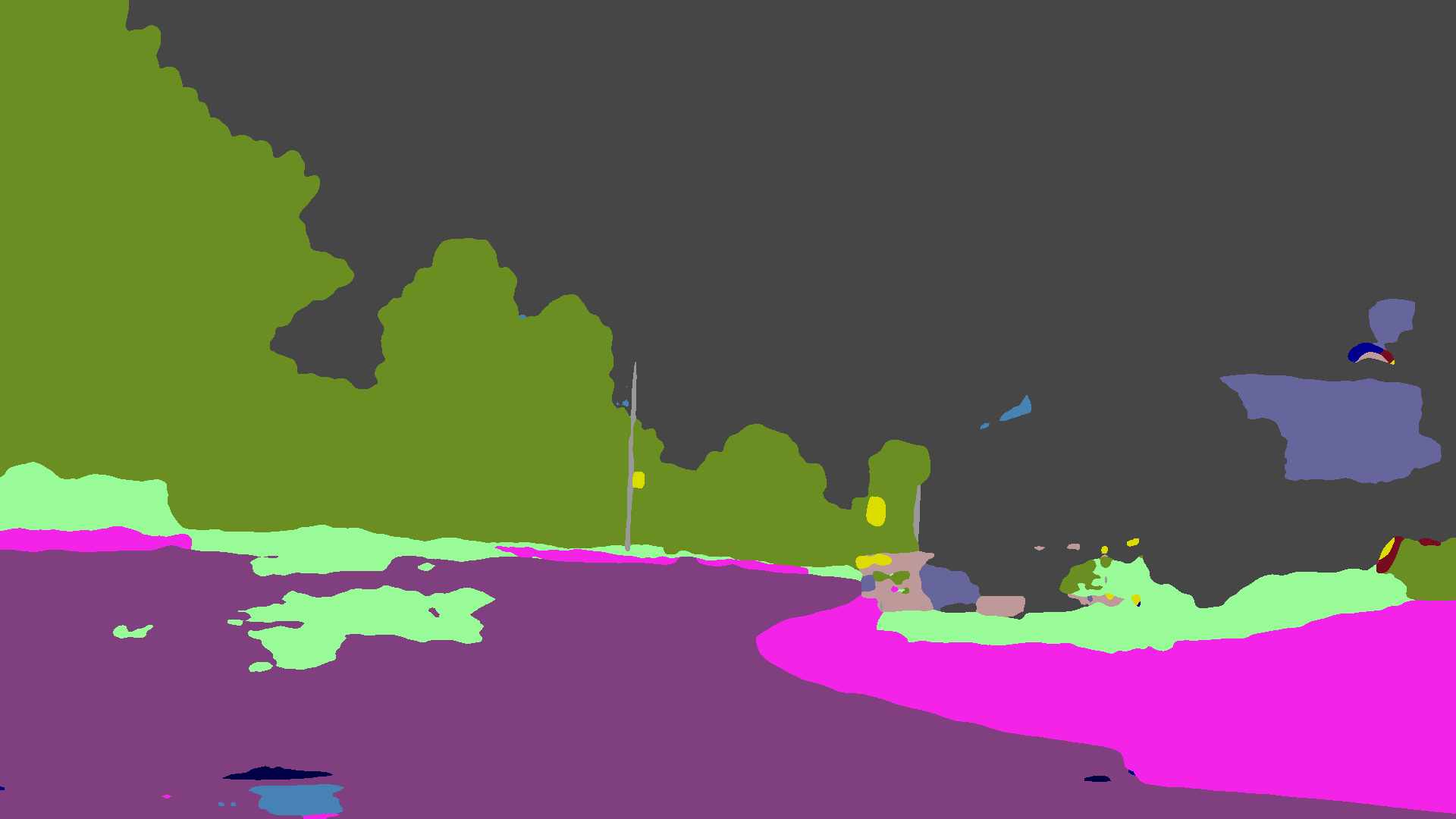} & 
			\includegraphics[width=0.31\columnwidth]{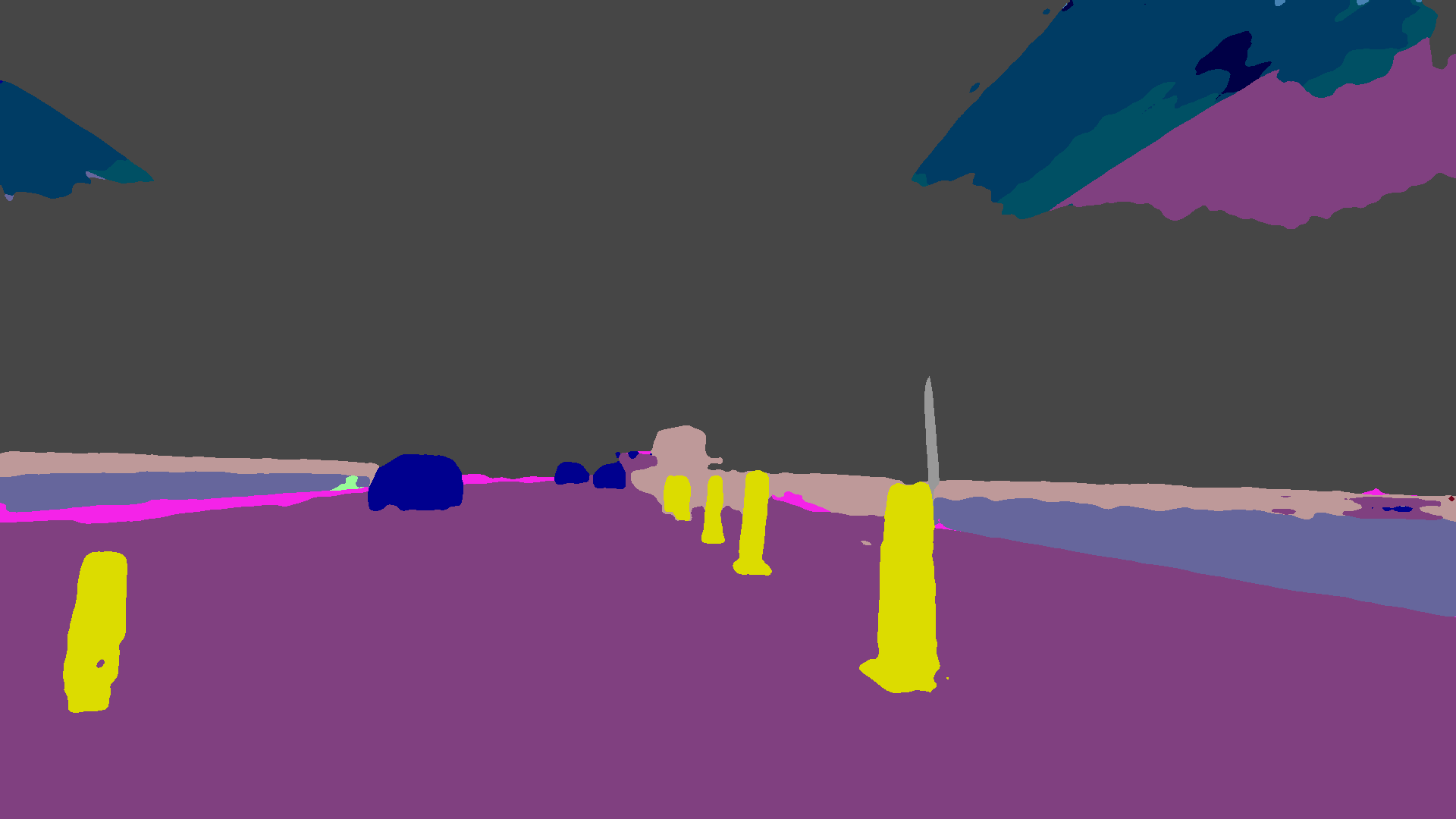} &
			\includegraphics[width=0.31\columnwidth]{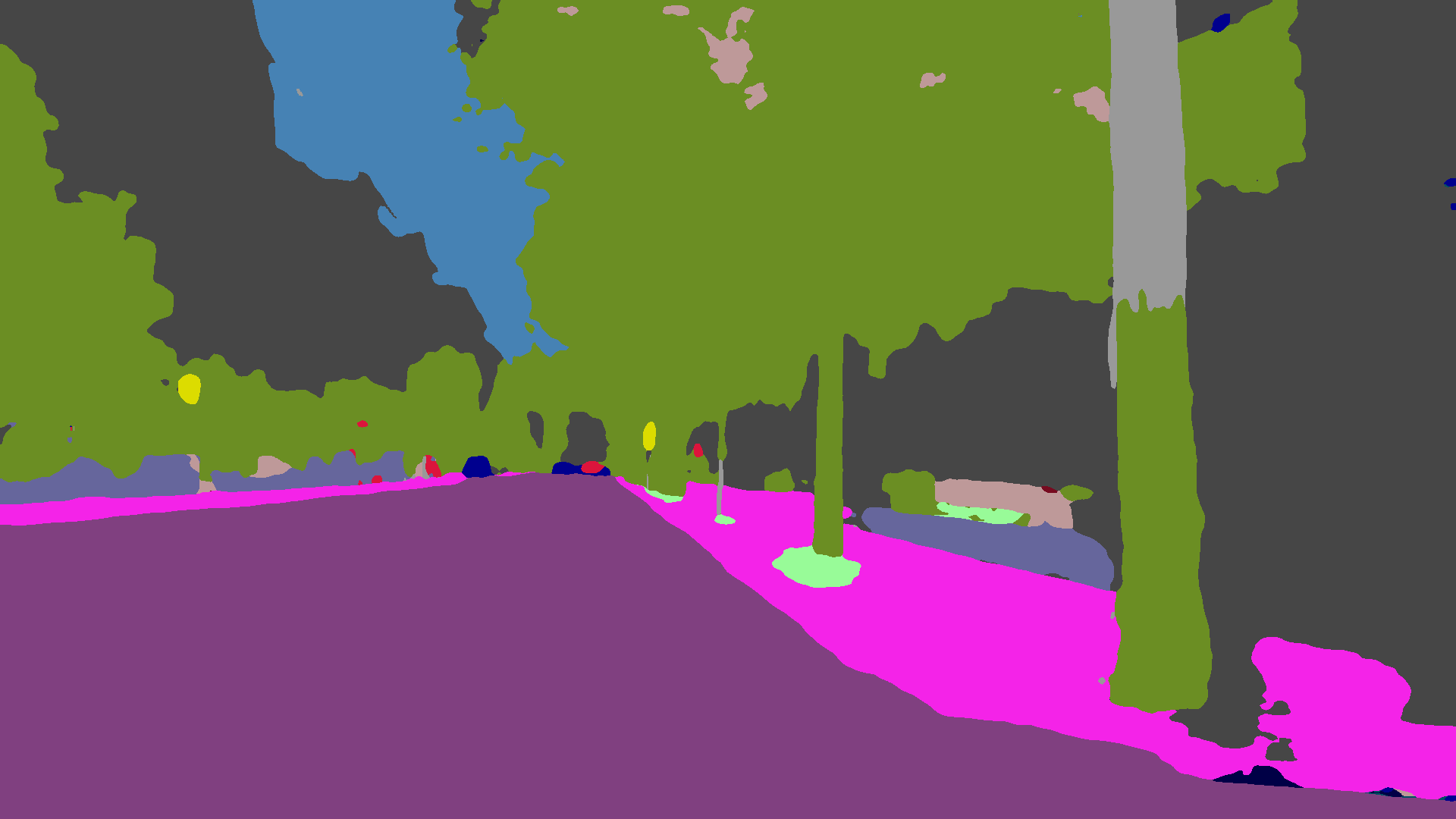} \\	\rotatebox{90}{Initial label} &
			\includegraphics[width=0.31\columnwidth]{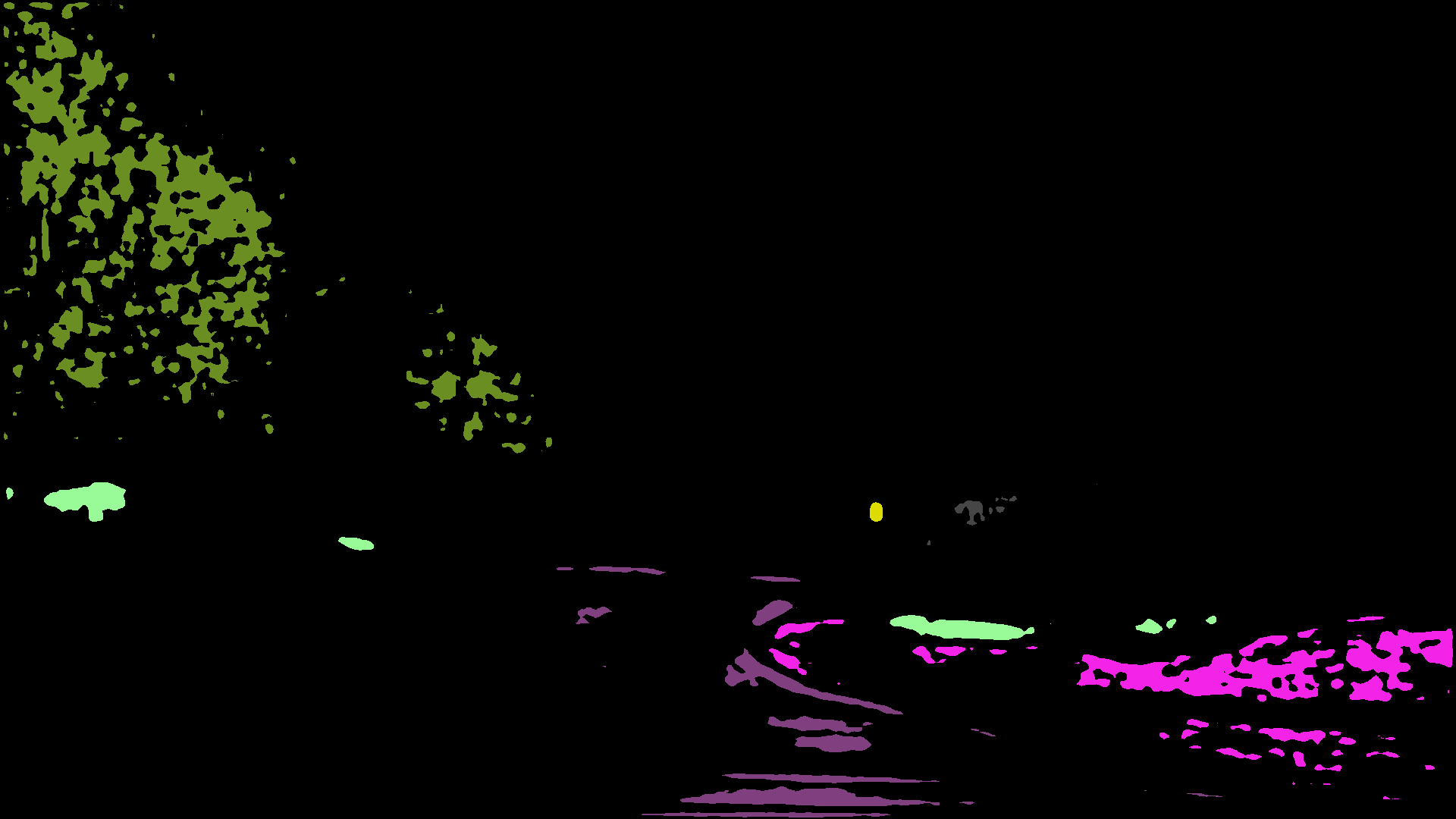} &
			\includegraphics[width=0.31\columnwidth]{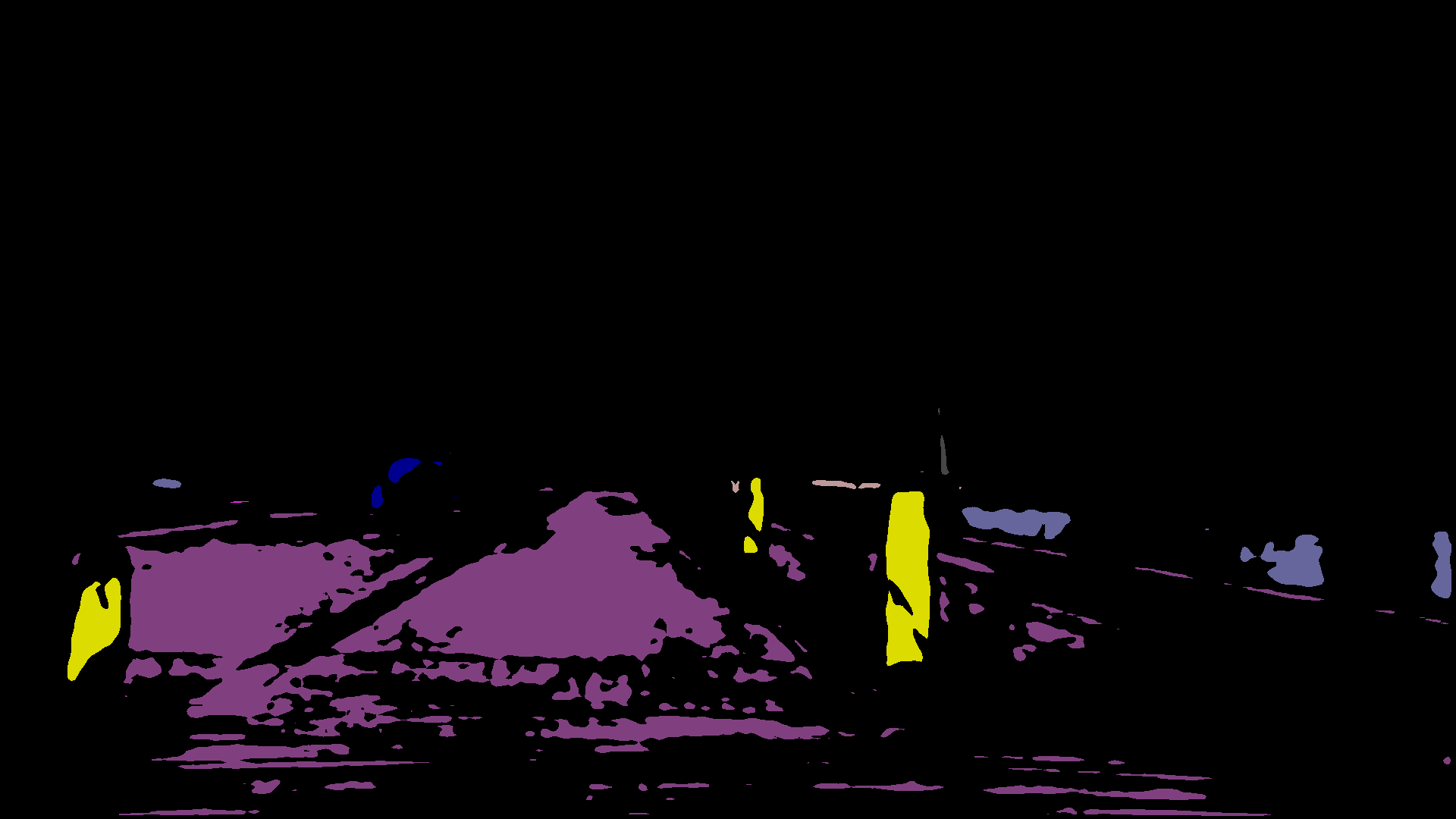} &
			\includegraphics[width=0.31\columnwidth]{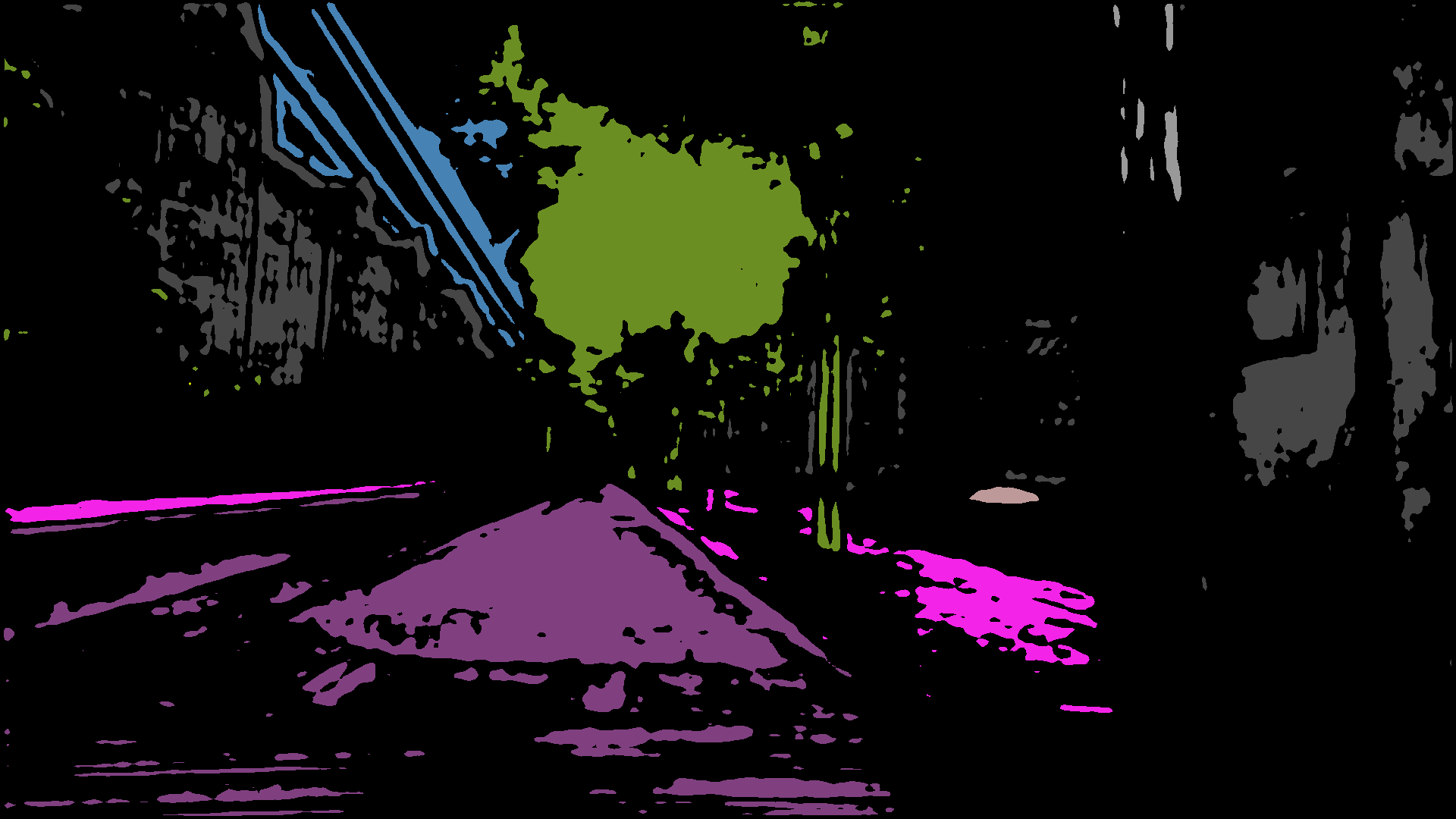}\\	\rotatebox{90}{\makecell[c]{Result from \\ superpixels}} &
			\includegraphics[width=0.31\columnwidth]{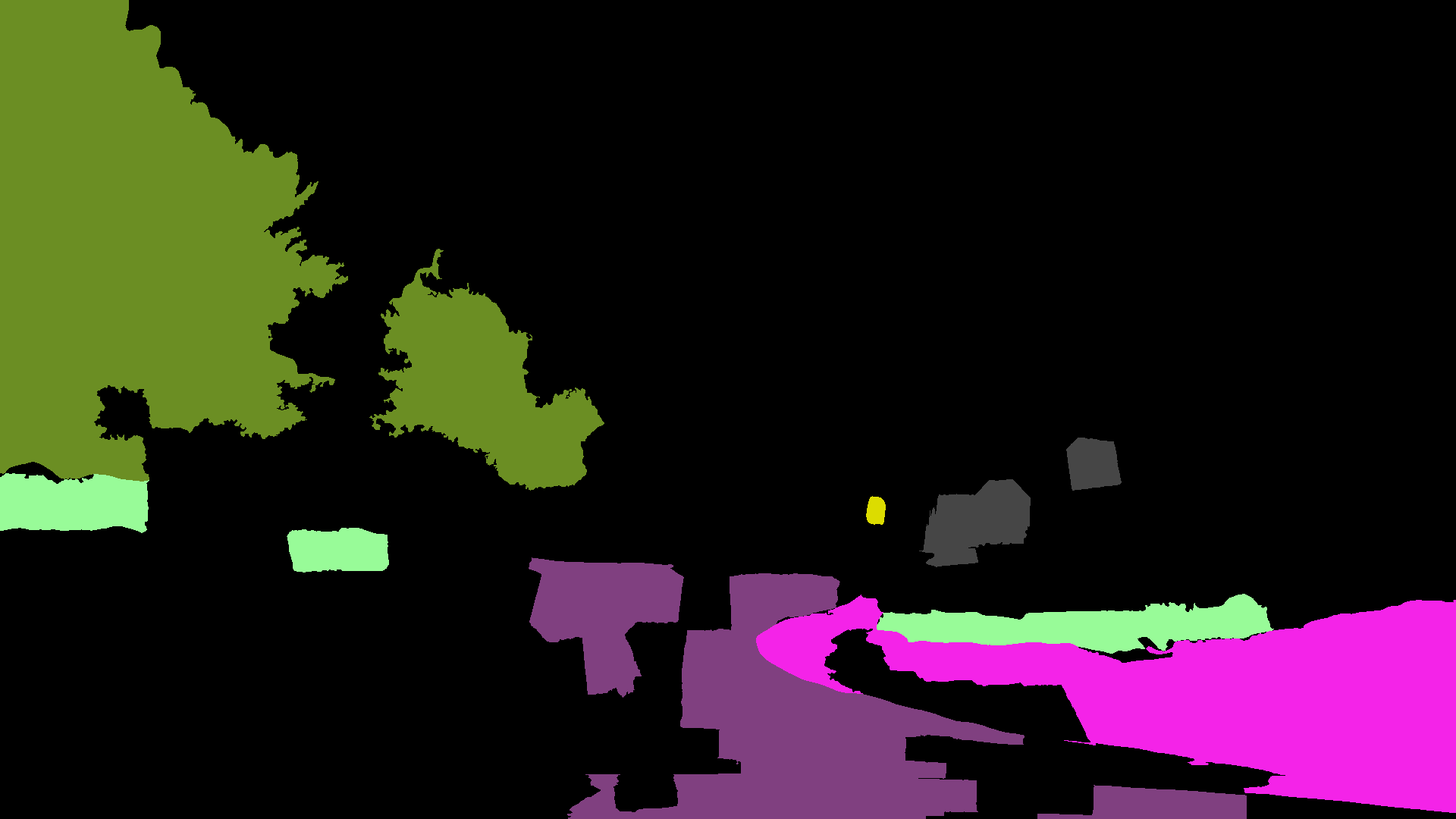} &
			\includegraphics[width=0.31\columnwidth]{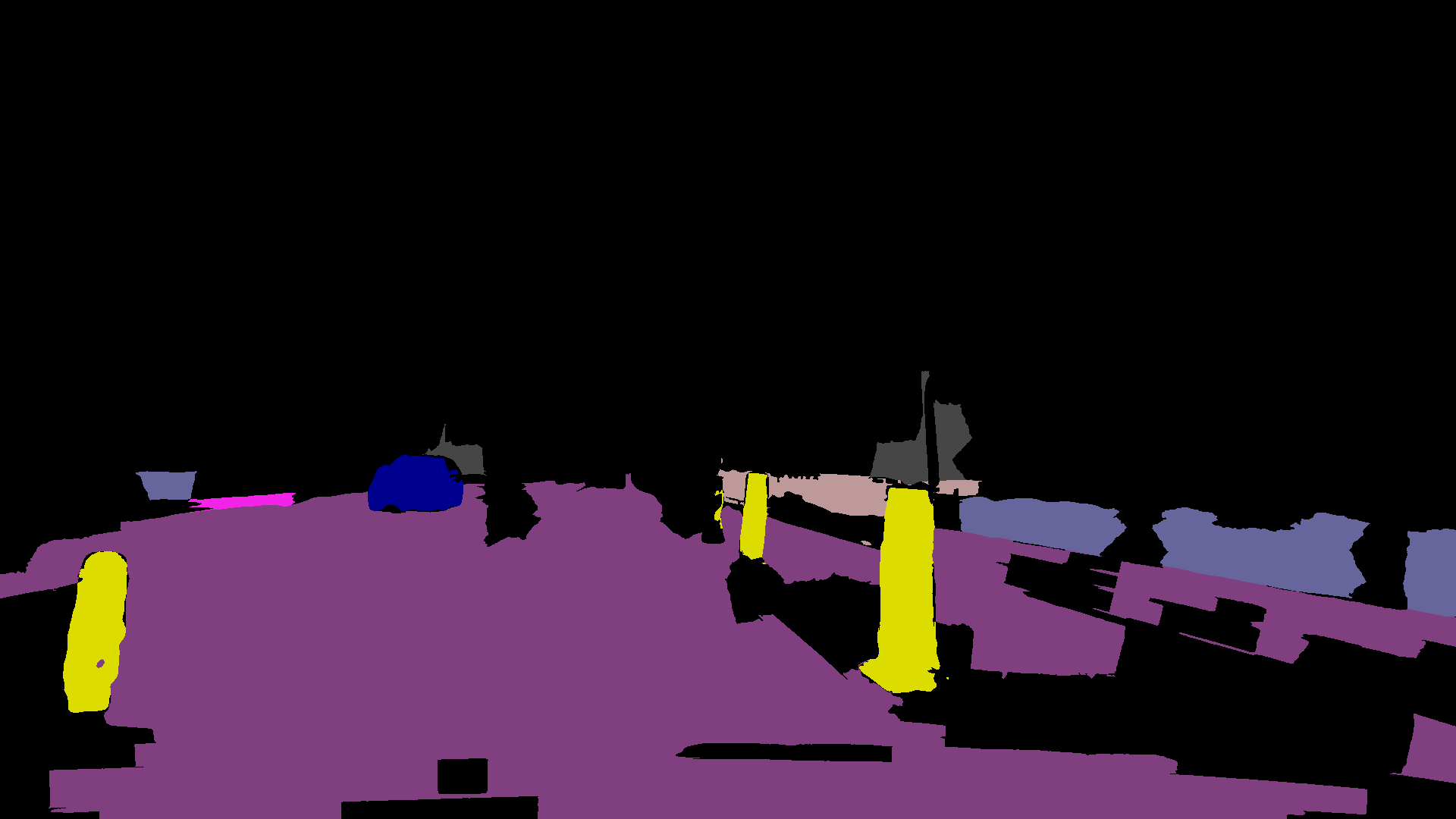} &
			\includegraphics[width=0.31\columnwidth]{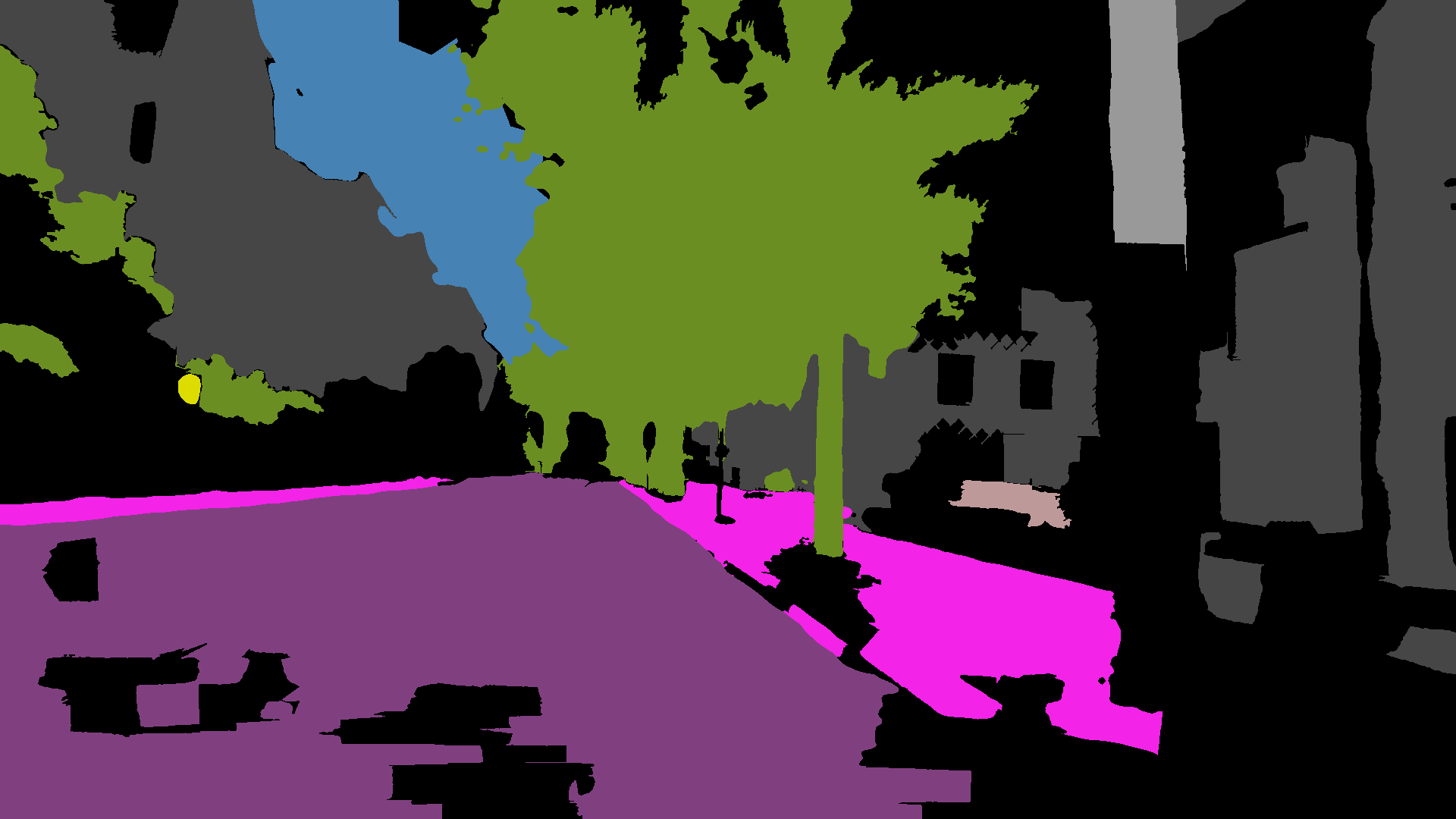}  \\	\rotatebox{90}{\makecell[c]{Result from \\ deep features}} &
			\includegraphics[width=0.31\columnwidth]{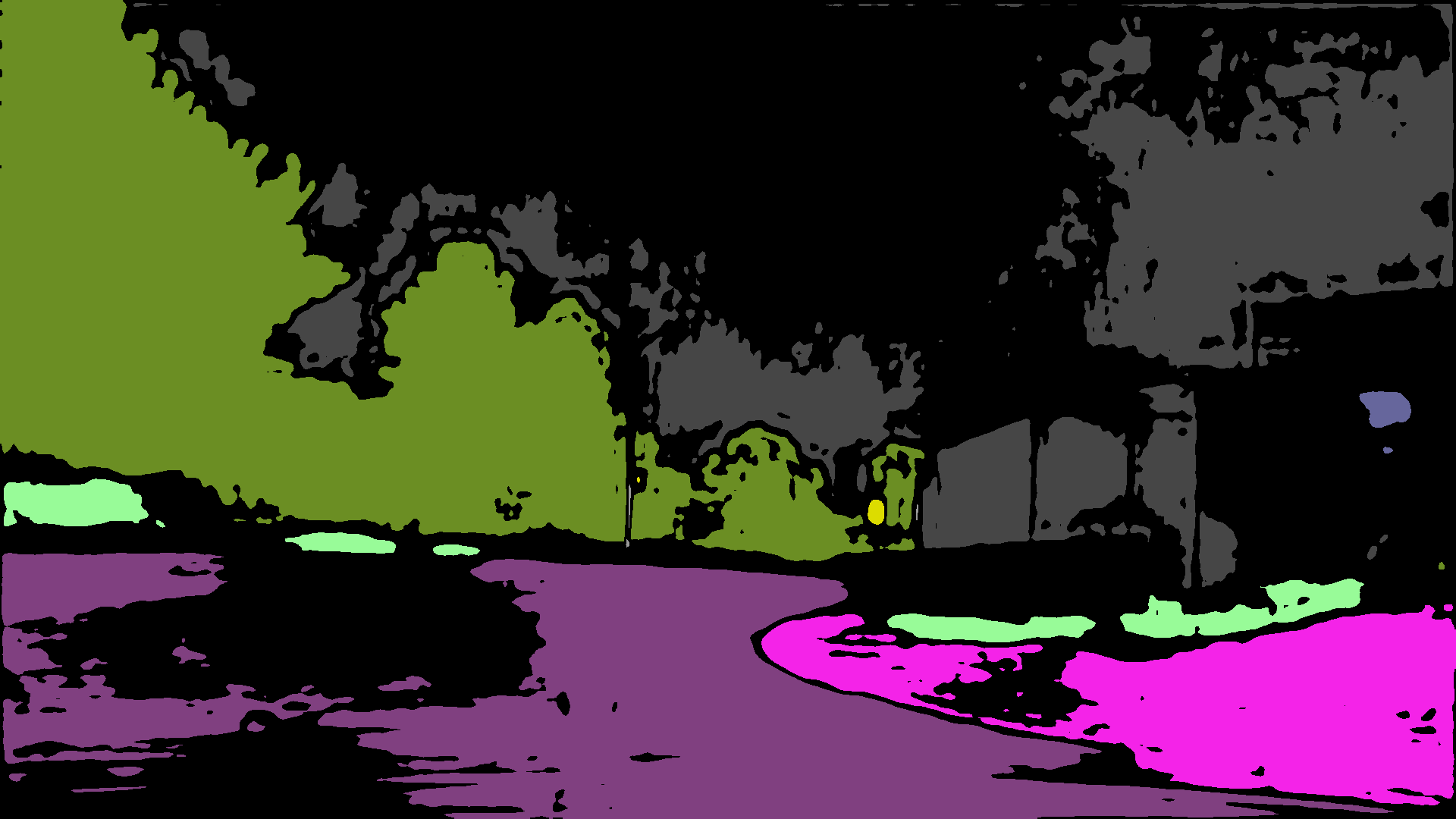} &
			\includegraphics[width=0.31\columnwidth]{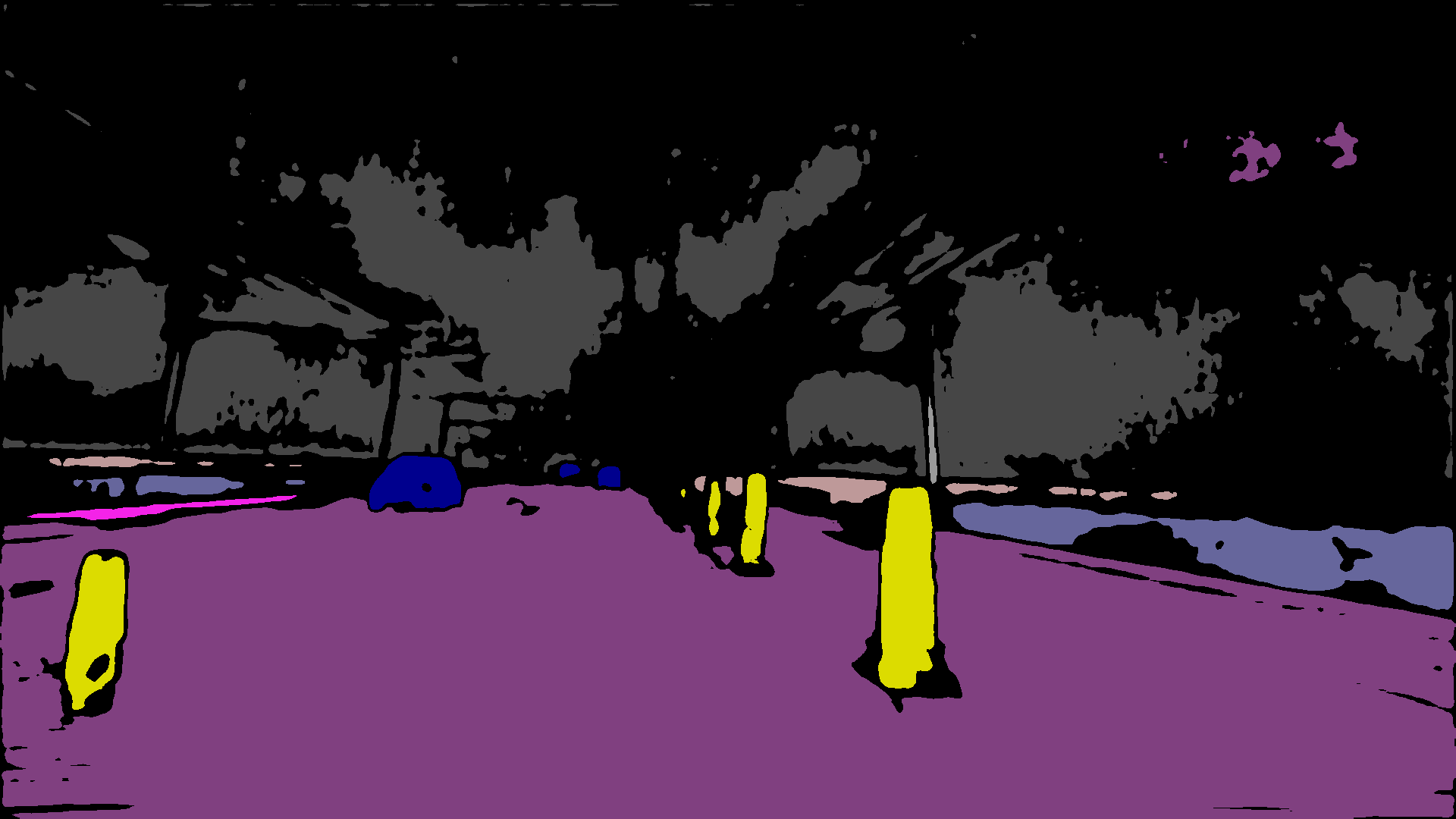} &
			\includegraphics[width=0.31\columnwidth]{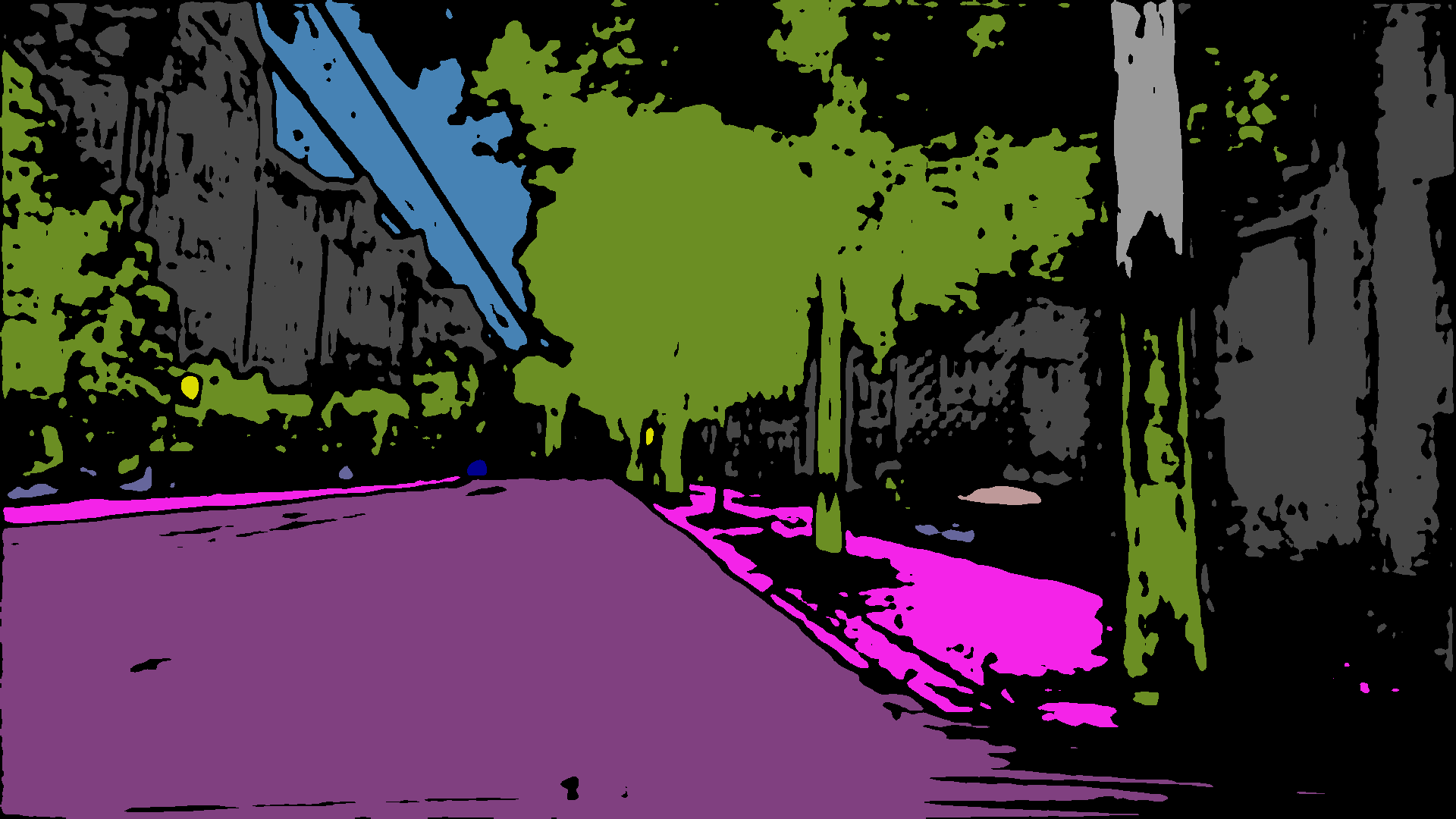}  \\
	\end{tabular}}
   \caption{Visual comparisons on spatial diffused pseudo labels between superpixels and deep feature.}
\label{fig:deepfeature}
\end{figure}

The superpixel-based spatial diffusion is performed under the assumption that local spatial similarity in the target domain can be measured more reliably  by superpixels than by deep feature extracted by segmentation models. Here we attempt to justify this assumption by visually comparing the pesudo labels in Fig.~\ref{fig:deepfeature}. The original predictions represents the semantic clustering from the deep features. In order to compare with the \textit{result from superpixels}, we select 50\% of the best predictions as the \textit{result from deep features}. We can notice that the there are obvious incorrect labels, which is much better in the superpixel-based spatial diffusion. Another advantage of adopting super-pixels is that it adaptively adjust the extent of diffusion according to the difficulty of the scene. For example, the diffused labels are fewer for a heavy fog in the first column than for a light fog in the third column, but a fixed selection ratio from deep feature will inevitably introduce wrongly predicted labels.

\tabcolsep=0.5pt

\subsubsection{Impact of the Number of Superpixels}
\begin{figure}[tb]
	\centering
	\includegraphics[width=\columnwidth]{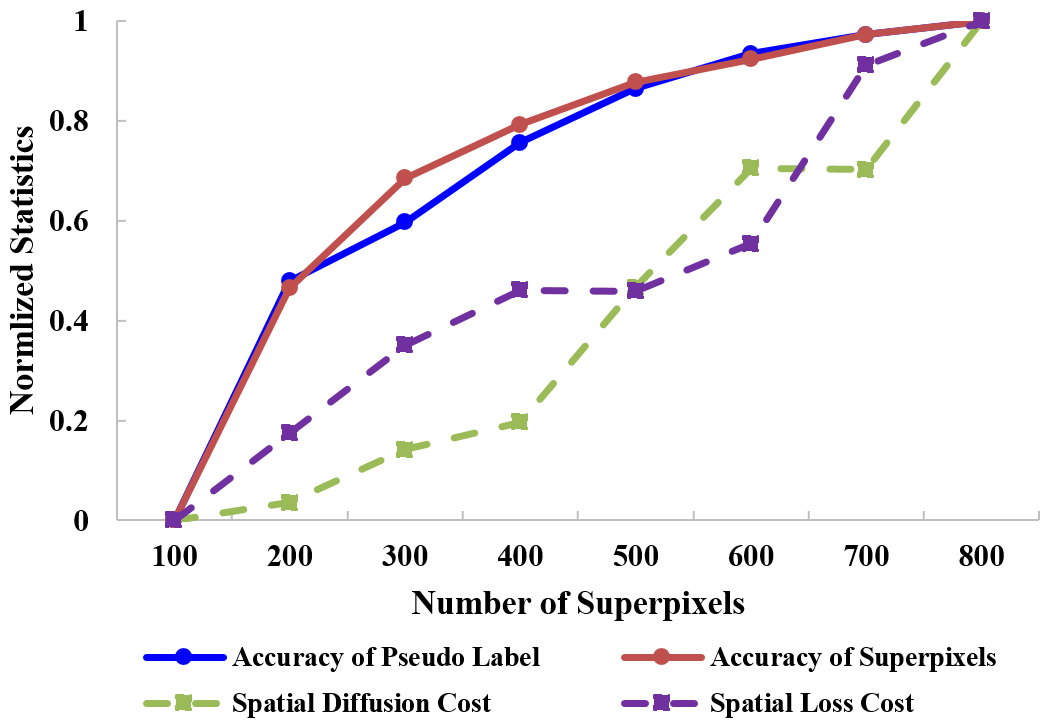}
   \caption{Statistical comparisons of the number of superpixels.}
\label{fig:numbersuperpixel}
\end{figure}

The success of spatial label diffusion largely relies
on the clustering of superpixels. Here we conduct experiment to validate the impact of hyper-parameter $K$ for number of superpixels on the pseudo labels diffusion. With experiments ranging from 100 to 800, we plot the normalized distribution of accuracy of pseudo labels, accuracy of superpixels (\emph{i.e.}, the percentage of superpixels with uniquely segmentation labels), computational cost of spatial diffusion and spatial loss in Fig.~\ref{fig:numbersuperpixel}. The plot shows that the proportion of good-fidelity labels and superpixels generally increases with the number of superpixels, while computational complexity for diffusing labels and calculating losses also increases. In our experiments, we adopt 500 as the number of superpixels per image, considering that the increment of accuracy is approaching saturation at 500, but computational cost is linearly growing for full range of $K$.

\subsubsection{Impact of the Range of Confident Flows}
For a sanity check, we study the impact of the threshold of confident flows by changing $T$ from 0.1 to 0.8. The percentages of pixels and mIoU of pseudo labels are shown in Table~\ref{temporal_range}, which shows the inclement of valid pixels and the degradation of mIoU are not significant because the initial pseudo labels is very sparse. In our experiment, we adopt the radius $T$ of 0.5, the same as in \cite{truong2021learning}.

\tabcolsep=0.5pt

\begin{table}[tb]
\centering
\caption{Quantitative comparisons of the Range of Confident Flows}
\setlength\tabcolsep{4pt}
\linespread{1.5}
\resizebox{1.0\columnwidth}{!}
{
\begin{tabular}{|c|cccccccc|}
\hline
$T$ & 0.1&0.2& 0.3&0.4&0.5&0.6&0.7&0.8 \\ \hline
Percentage& 0.122 & 0.119 & 0.117 & 0.116 & 0.114 & 0.110 & 0.110 & 0.110 \\
mIoU& 0.841 & 0.843 & 0.843 & 0.844 & 0.844 & 0.845 & 0.845 & 0.845 \\ \hline
\end{tabular}}
\label{temporal_range}

\end{table}

\tabcolsep=0.5pt
\begin{figure}[tb]
	\centering
	\includegraphics[width=\columnwidth]{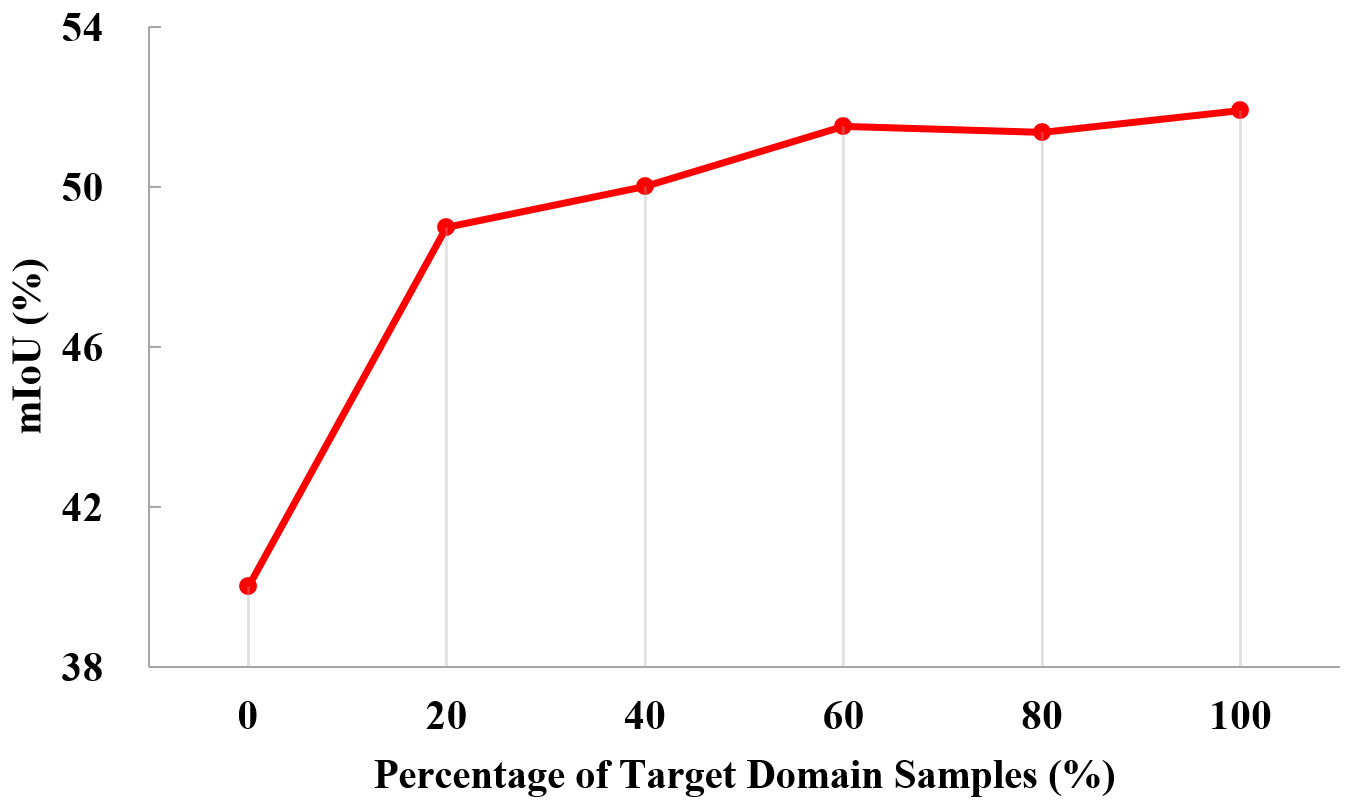}
   \caption{ {Effectiveness of number of target domain samples. The total number of target domain samples is 3808.}}
\label{fig:targetsample}
\end{figure}

\subsubsection{ {Effects of Number of Target Domain Samples}}
 {To investigate the effectiveness of the number of target domain samples on the segmentation performance, we test six variants of randomly selected target sample numbers, which are 0\%, 20\%, 40\%, 60\%, 80\%, 100\% of all target domain samples. The results in Fig.~\ref{fig:targetsample} shows that  increasing the target domain samples will highly improve the segmentation performance. With only 761 sample images from target domain (20\% of the total number), the performance in mIoU increases from 40.02\% to 49.00\%, and it continues to increase with the number increment of target domain samples. But when the number of samples exceeds 60\% (around 2284 target domain samples), the performance almost reaches the saturation and fluctuates between 51.50\% to 52.00\%.}

\subsubsection{ {Feature Visualization}}
\begin{figure*}[tb]
	\centering
\footnotesize{
		\begin{tabular}{ccccc}
		\makecell[c]{(a) Pre-trained model}&\makecell[c]{(b) Adaptive model after \\  round 1}&\makecell[c]{(c) Adaptive model  after \\  round 2}&\makecell[c]{(d) Adaptive model  after \\  round 3}&\makecell[c]{(e) Adaptive model  after \\ round 4}\\
			\includegraphics[width=0.19\textwidth]{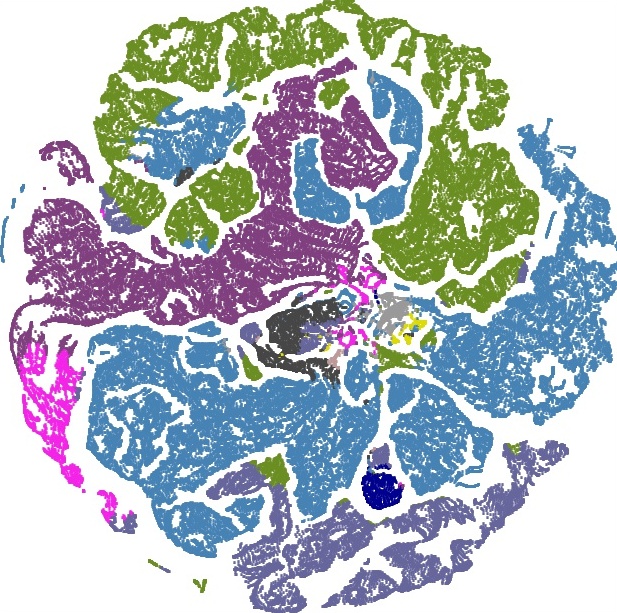}~ &
			\includegraphics[width=0.19\textwidth]{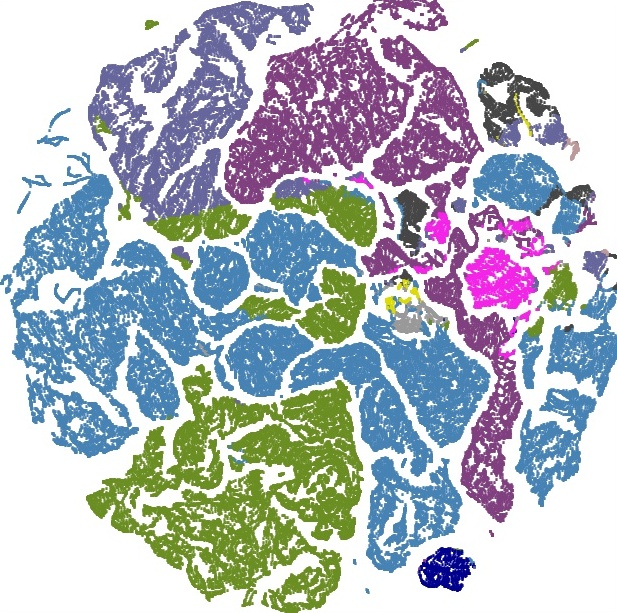}~ &
			\includegraphics[width=0.19\textwidth]{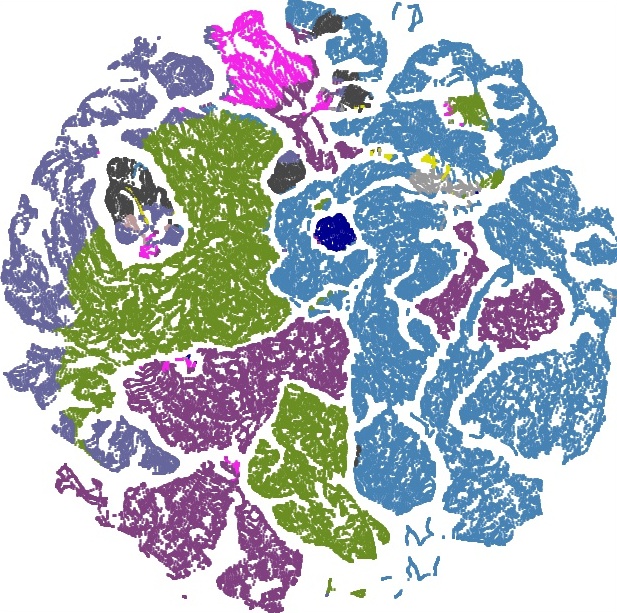}~ &
			\includegraphics[width=0.19\textwidth]{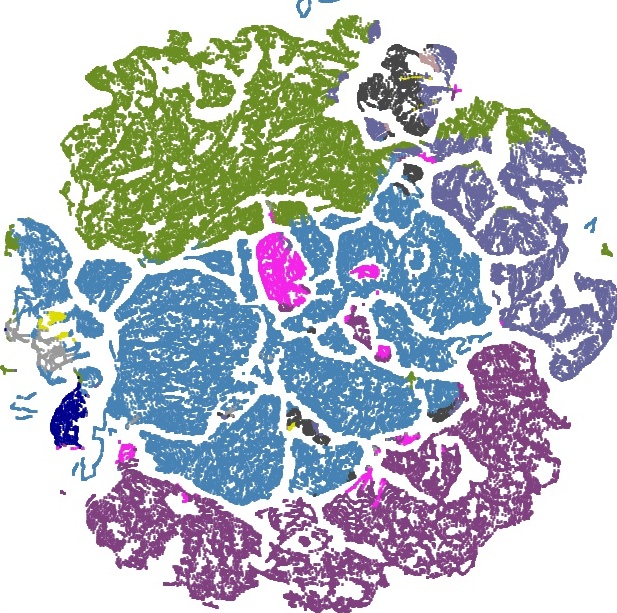}~ &
			\includegraphics[width=0.19\textwidth]{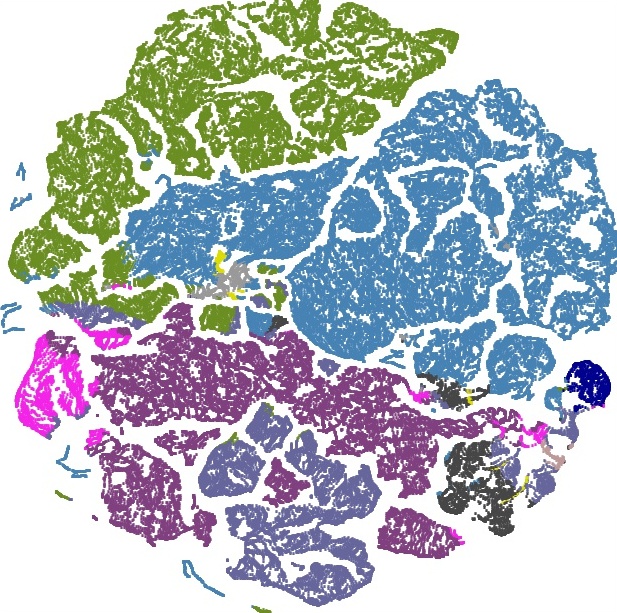} \\
			\includegraphics[width=0.19\textwidth]{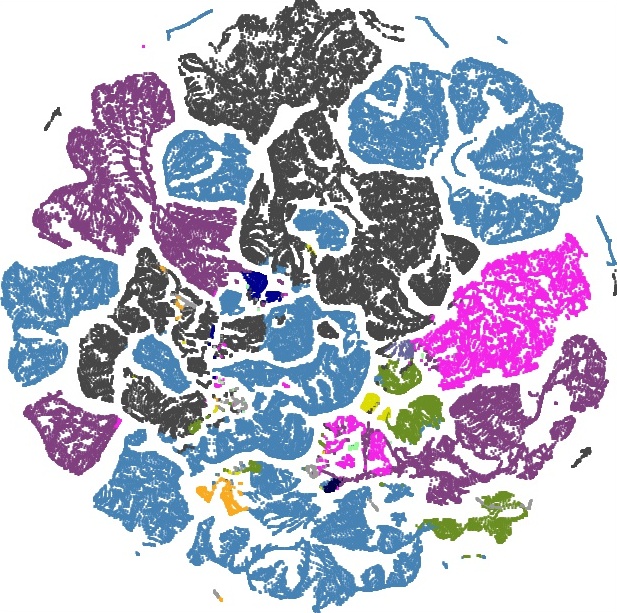}~ &
			\includegraphics[width=0.19\textwidth]{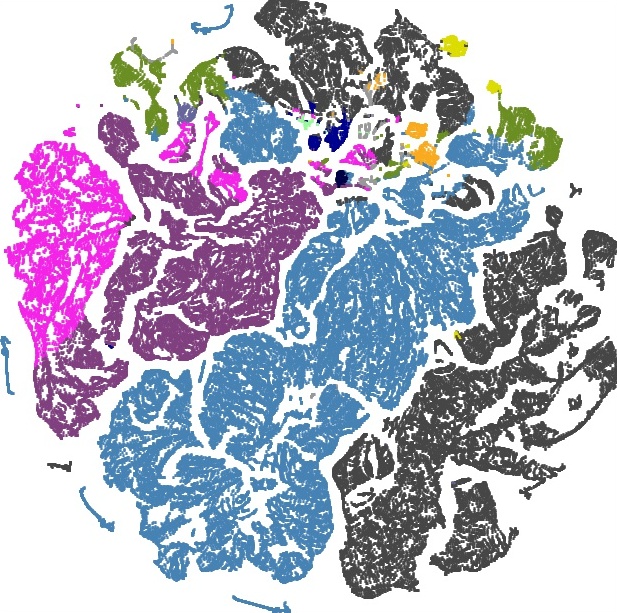}~ &
			\includegraphics[width=0.19\textwidth]{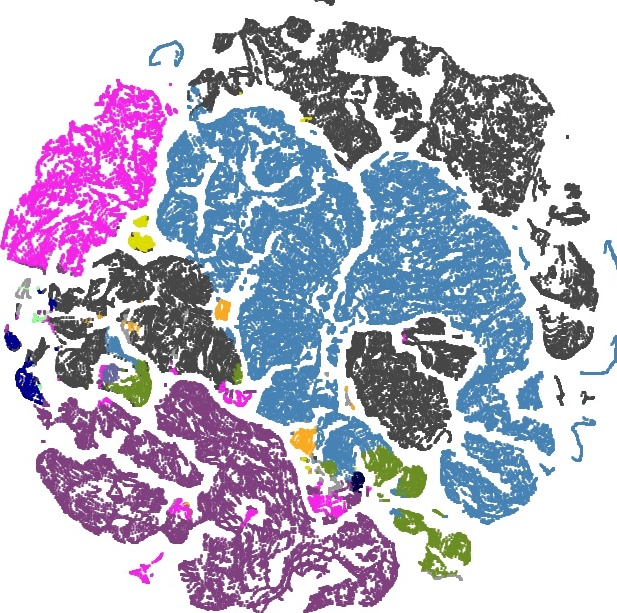}~ &
			\includegraphics[width=0.19\textwidth]{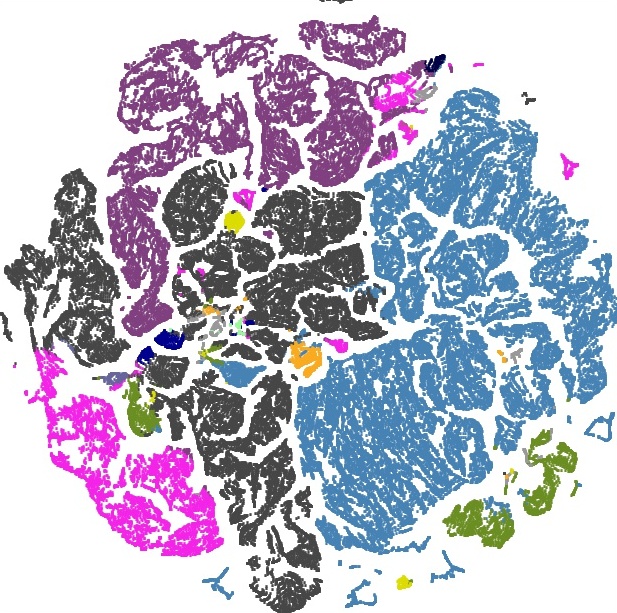}~ &
			\includegraphics[width=0.19\textwidth]{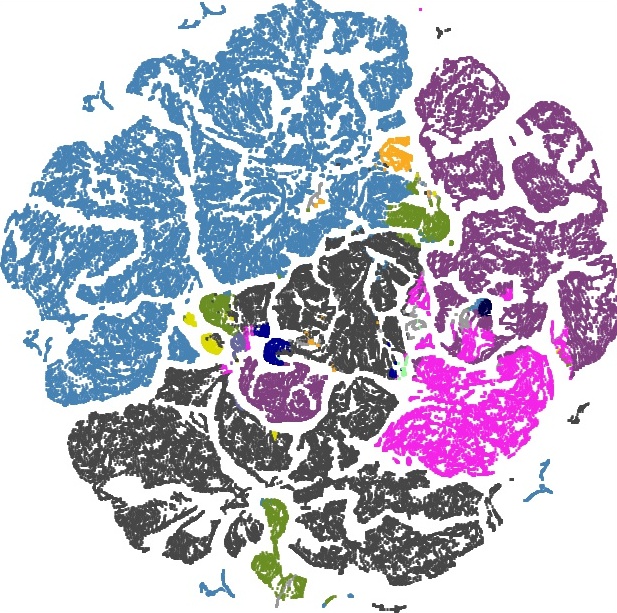} \\
			\multicolumn{5}{c}{\includegraphics[width=0.98\textwidth]{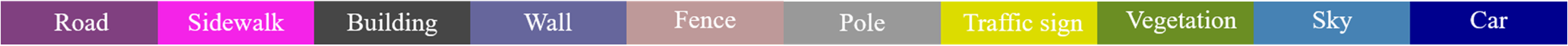} }\\
	\end{tabular}}
   \caption{ {Visualization of embedded features via t-SNE from two image samples of \textbf{Foggy Zurich} test set. The features (from left to right) are extracted from the pre-trained model on source domain and the adaptive models after each self-training round. Features are colored according to class labels.}}
\label{fig:tngfeature}
\end{figure*}

 {We use t-SNE~\cite{tsne} to visualize the feature representation of the domain adaptive process in Fig.~\ref{fig:tngfeature}. Since we re-train the model for four rounds, we show the features in five phrases, \textit{i.e.}, the pre-trained model on the source domain and four adaptive models after each self-training round. It can be observed that the classes containing large number of pixels, such as \textit{Road}, \textit{Sky} and \textit{Building}, are not well clustered by the pre-trained model. 
Since our method can select most reliable pseudo labels, the feature distributions are gradually refined as the number of self-training rounds increases. The feature presentation exhibits much clearer after round 4, revealing that our TDo-Dif method can provide correct supervision labels for target domain data. }

\tabcolsep=0.5pt

\section{Conclusion}
In this paper, we have shown the benefits of using target domain knowledge in an self-training framework to improve the performance of state-of-the-art domain adaptive semantic segmentation models in real foggy scene. To achieve this, we propose a superpixel-based spatial diffusion and an optical flow-based temporal diffusion scheme that exploit the characteristics of the target data to generate reliable and dense pseudo labels. When the former better increases the number of pseudo labels and the latter better maintain the confidence of the diffused labels. In addition, two new losses are proposed to restrain the spatial and temporal consistency of the features in the re-training stage to help the model better adapt to the characteristics of the target domain. Experiments on real foggy images show that the proposed TDo-Dif method significantly densify the pseudo labels and further boost the performance of semantic segmentation beyond the state-of-the-art counterparts.



\ifCLASSOPTIONcaptionsoff
  \newpage
\fi


\end{document}